\pdfoutput=1
\documentclass[11pt]{article}

\usepackage[utf8]{inputenc} % allow utf-8 input
\usepackage[T1]{fontenc}    % use 8-bit T1 fonts
\usepackage{booktabs}       % professional-quality tables
\usepackage{amsfonts}       % blackboard math symbols
\usepackage{nicefrac}       % compact symbols for 1/2, etc.
\usepackage{microtype}      % microtypography
\usepackage[table]{xcolor}  % colors
\PassOptionsToPackage{hyphens}{url} % allow long URLs to wrap while staying clickable

\usepackage{./smile}

\usepackage{tikz}
\usetikzlibrary{matrix, positioning}
\usepackage{pgfplots}
\pgfplotsset{compat=1.18}
\usepackage{wrapfig}
\usepackage{xcolor}
\usepackage{cancel}
\usepackage{acronym}

\definecolor{mypink}{HTML}{FF1493}
\definecolor{darkred}{HTML}{BB253B}
\definecolor{lightgreen}{HTML}{90C0BF}

\definecolor{darkcyan}{HTML}{0081B9}

\newmdenv[
  backgroundcolor=gray!10,
  skipabove=1em,
  skipbelow=1em,
  leftline=false,
  topline=false,
  bottomline=false,
  rightline=false,
  linecolor=gray!88,
  linewidth=4pt
]{githubquote}

\usepackage[most,skins,theorems]{tcolorbox}
\tcbset{
  aibox/.style={
    width=\linewidth,
    top=10pt,
    bottom=4pt,
    colback=blue!6!white,
    colframe=black,
    colbacktitle=black,
    enhanced,
    center,
    attach boxed title to top left={yshift=-0.1in,xshift=0.15in},
    boxed title style={boxrule=0pt,colframe=white,},
  }
}

\definecolor{rliableolive}{HTML}{BBCC33}
\definecolor{rliableblue}{HTML}{77AADD}
\definecolor{rliablered}{HTML}{EE8866}
\definecolor{LightCyan}{rgb}{0.88,1,1}
\definecolor{darkblue}{HTML}{2878D9}
\definecolor{navyblue}{HTML}{0000FF}

\newtcolorbox{AIbox}[2][]{aibox,title=#2,colback=rliableblue!10!white,#1}

\providecolor{yaleblue}{HTML}{00356B}
\colorlet{findingframe}{yaleblue}
\definecolor{findingaccent}{HTML}{D97757}
\colorlet{findingbg}{yaleblue!5!white}

\newtcolorbox[%
  auto counter,
  number within=section,
  crefname={finding}{findings},
  Crefname={Finding}{Findings},
]{finding}[1][]{%
  enhanced,
  breakable,
  before skip=10pt,
  after skip=10pt,
  colback=findingbg,
  colframe=findingframe,
  boxrule=0pt,
  arc=3pt,
  outer arc=3pt,
  left=12pt, right=12pt, top=8pt, bottom=10pt,
  before upper={%
    \noindent\textbf{Finding~\thetcbcounter\ifblank{#1}{}{. #1}.}\enspace
  },
}

\usepackage{subcaption}

\usepackage{titlesec}
\usepackage{xcolor}
\definecolor{sectionblue}{HTML}{00356B}

\titleformat{\section}
  {\Large\bfseries\color{sectionblue}}{\thesection}{1em}{}
\titleformat{\subsection}
  {\large\bfseries\color{sectionblue}}{\thesubsection}{1em}{}
\titleformat{\subsubsection}
  {\normalsize\bfseries\color{sectionblue}}{\thesubsubsection}{1em}{}
\titleformat{\paragraph}[runin]
  {\normalfont\normalsize\bfseries\color{black}}{}{0pt}{}

\newcommand{\compInfMathDelta}{+7.23}

\newcommand{\compFixReasonDelta}{+6.00}
\newcommand{\compFixMathDelta}{+2.55}

\newcommand{\compStrongReasonDelta}{+7.52}
\newcommand{\compStrongMathDelta}{+3.28}

\newcommand{\compDevOnlyReasonDelta}{+5.83}
\newcommand{\compDevOnlyMathDelta}{+6.13}

\newcommand{\compInfStrongReasonGap}{0.3}

\newcommand{\compHoldoutSize}{9{,}038}

\newcommand{\compLeakInfDev}{41.9}
\newcommand{\compLeakDevOnlyDev}{86.5}

\newcommand{\compLeakInfHoldout}{39.2}
\newcommand{\compLeakDevOnlyHoldout}{38.4}

\usepackage{fancyhdr}
\newcommand\blfootnote[1]{%
  \begingroup
  \renewcommand\thefootnote{}\footnote{#1}%
  \addtocounter{footnote}{-1}%
  \endgroup
}

\title{
  \vskip-30pt
  \textbf{INFUSER}: \textbf{Inf}luence-G\textbf{u}ided \textbf{S}elf-\textbf{E}volution Improves \textbf{R}easoning
}

\author{
  Siyu Chen$^{1}$ \quad
  Miao Lu$^{2}$ \quad
  Beining Wu$^{3}$ \quad
  Heejune Sheen$^{1}$ \quad
  Fengzhuo Zhang$^{1}$ \\[2pt]
  Shuangning Li$^{3}$ \quad
  Zhiyuan Li$^{4}$ \quad
  Jose Blanchet$^{2}$ \quad
  Tianhao Wang$^{5}$ \quad
  Zhuoran Yang$^{1}$ \\[6pt]
  {\small $^{1}$Yale University \quad $^{2}$Stanford University \quad $^{3}$University of Chicago} \\[1pt]
  {\small $^{4}$Toyota Technological Institute at Chicago \quad $^{5}$University of California, San Diego}
}
\date{}
\begin{document}

\maketitle

\blfootnote{%
  Author emails:
  \texttt{siyu.chen.sc3226@yale.edu},
  \texttt{miaolu@stanford.edu},
  \texttt{beiningw@uchicago.edu},
  \texttt{heejune.sheen@yale.edu},
  \texttt{fengzhuo.zhang@yale.edu},
  \texttt{shuangning.li@chicagobooth.edu},
  \texttt{zhiyuanli@ttic.edu},
  \texttt{jblanche@stanford.edu},
  \texttt{tianhaowang@ucsd.edu},
  \texttt{zhuoran.yang@yale.edu}.
}

\pagestyle{plain}

  \vspace{-10mm}
\begin{abstract}
Self-evolution offers a scalable path to stronger reasoning: a
pretrained language model improves itself with only minimal external
supervision. Yet existing methods either depend on extensively curated
or teacher-generated training data, or, when the generator runs
unsupervised, reward it by a difficulty heuristic that need not
improve the solver. We
introduce \textbf{INFUSER}, an iterative co-training framework with two
co-evolving roles: a \textit{Generator} that drafts questions and reference
golden answers from a pool of unstructured, automatically collected
documents, and a \textit{Solver} that improves by training on them. The solver is trained
with standard correctness rewards against the generator-provided answers,
while the generator is rewarded by an \textit{optimizer-aware influence
score} that measures whether each proposed question would actually improve
the solver on the target distribution. Because this continuous, noisy
influence score is poorly served by standard GRPO, we propose \textbf{DuGRPO}, a
dual-normalized variant of GRPO, for generator training. Together, these turn
the document pool into an \textit{adaptive curriculum} that favors questions
useful to the current solver, not just hard ones. On Qwen3-8B-Base, INFUSER
outperforms strong self-evolution baselines with over 20\% relative
improvement on Olympiad and SuperGPQA benchmarks, and an 8B INFUSER
co-evolving generator outperforms a frozen 32B thinking generator on math and
coding. Ablations confirm each design choice is necessary, and two
extensions, applying INFUSER to an instruction-finetuned anchor and
augmenting it with rule-verifiable RLVR data, further demonstrate the
flexibility and generalizability of the framework.
Code is available at \url{https://github.com/FFishy-git/INFUSER}.
\end{abstract}

\begin{figure}[!h]
  \centering
  \vspace{-5mm}
  \begin{minipage}[t]{0.45\linewidth}
    \centering
    \includegraphics[width=\linewidth]{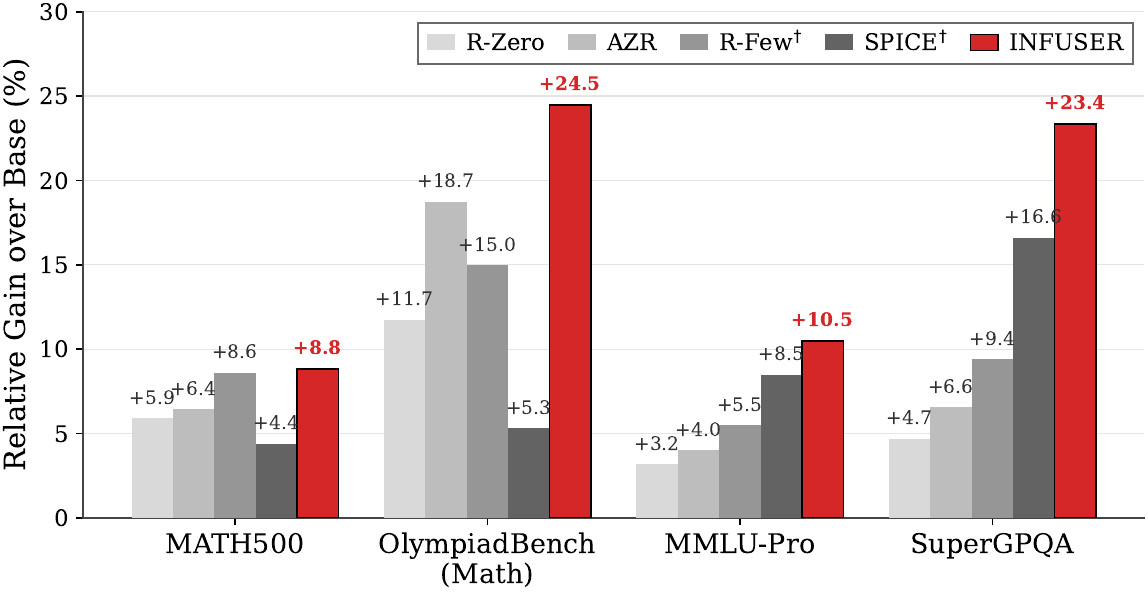}
  \end{minipage}
  \begin{minipage}[t]{0.42\linewidth}
    \centering
    \includegraphics[width=\linewidth]{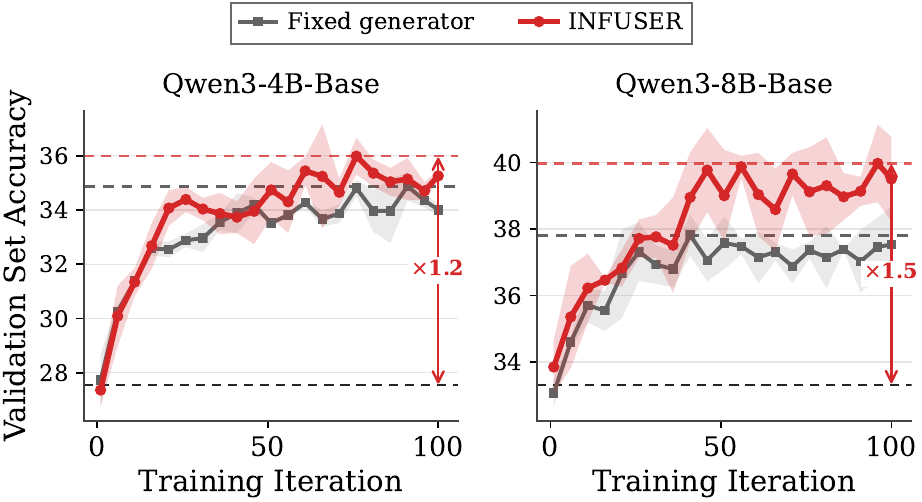}
  \end{minipage}
  \caption{\textbf{INFUSER} on Qwen3 base anchors. \emph{Left:}
    relative accuracy gain over Qwen3-8B-Base on four headline
    benchmarks for each self-evolution method. \emph{Right:} validation-set
    accuracy curves over training iterations for INFUSER versus a
    fixed-generator baseline with matching hyperparameters on
    Qwen3-4B-Base (left subpanel) and Qwen3-8B-Base (right subpanel);
    curves are averaged over 3 random seeds.}
  \label{fig:main_qw8bb}
  \vspace{-5mm}
\end{figure}

\section{Introduction}
\label{sec:intro}
Reinforcement learning with verifiable rewards (RLVR) underlies much of the recent progress in reasoning for large language models~\citep{guo2025deepseek,team2025kimi,shao2024deepseekmath,yu2025dapo,zhang2025rlsurvey}, but its scalability is bottlenecked by the supply of high-quality, verifiable training data, which is costly to produce in both research and industry settings.
\emph{Self-evolution} offers a path beyond this bottleneck: a generator proposes high-quality training data with itself or from unstructured documents, and a solver trains on that data. The whole improvement loop runs without an externally curated training corpus or teacher model~\citep{huang2025rzero,fang2025spice,yu2025rfew}. In principle, this either creates training signal from the model itself or converts abundant unstructured corpora into the structured signals that RLVR consumes.

Existing self-evolution methods, however, share two limitations that constrain their effectiveness.
The first concerns \emph{grounding}: the anchoring of generated training data in external sources rather than only the model's own outputs. Pure self-play methods such as R-Zero~\citep{huang2025rzero} forgo such an anchor and draw supervision entirely from the model's own outputs, which bounds learning by the model's prior knowledge and exposes the solver to hallucinated reference answers; executor-based methods such as AZR~\citep{zhao2025azr} substitute a code or symbolic executor for documents, restoring formal verifiability but restricting the framework to domains in which such an executor exists, e.g., code and mathematics.
The second concerns the \emph{generator's training objective}. Document-grounded approaches such as SPICE~\citep{fang2025spice} draw training questions from an external corpus, yet reward the generator by a difficulty heuristic that is maximized when the solver succeeds on approximately half of its rollouts. Difficulty is a coarse surrogate for utility: a question may register as difficult because it is ambiguously phrased, misaligned with its source document, or paired with an incorrect generated reference answer, and training on such a question carries no guarantee of improving the solver.
These two limitations together leave the following question open:
\begin{center}
\itshape Can we train a generator to produce document-grounded training data that genuinely improves the current solver, while co-evolving with it?
\end{center}

\begin{figure}[!t]
  \vspace{-6mm}
  \centering
  \includegraphics[width=\linewidth]{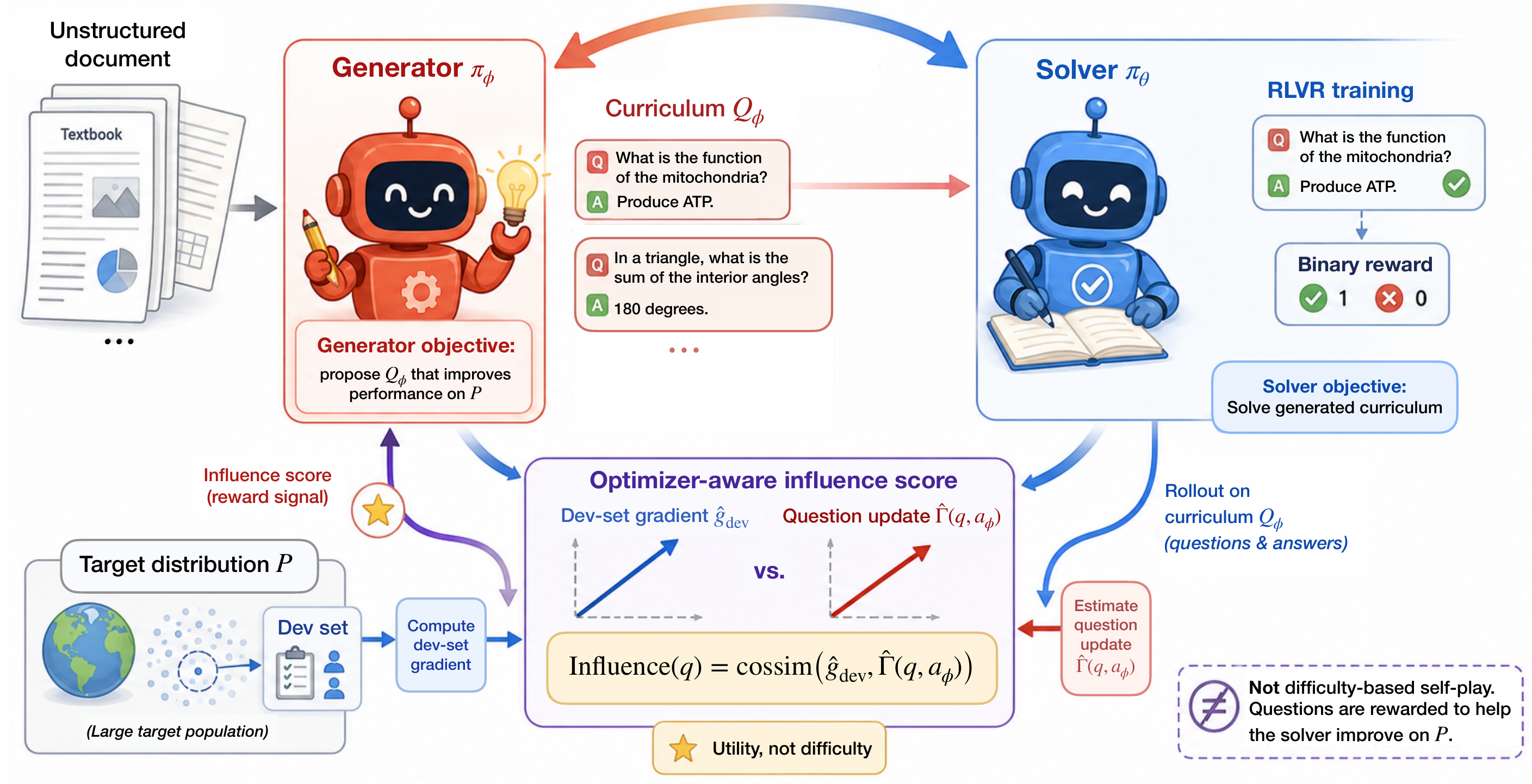}
  \caption{\small \textbf{INFUSER} casts document-grounded self-evolution as
    bilevel co-evolution between a generator and a solver.  The generator
    proposes a curriculum from unstructured documents, the solver improves on
    this curriculum through RLVR training, and the generator is rewarded by an
    optimizer-aware influence score that measures whether each generated
    question induces a solver update aligned with target-distribution
    improvement.}
  \label{fig:system_overview}
  \vspace{-5mm}
\end{figure}

We address this question by formulating self-evolution as a bilevel game between a generator $\pi_\phi$ and a solver $\pi_\theta$, both initialized from the same pretrained model.
As illustrated in \Cref{fig:system_overview}, for each iteration, the generator proposes self-generated question--answer (QA) pairs conditioned on an unstructured corpus (textbook chunks in our experiments), which form the \emph{curriculum}, and the solver is trained via a standard RLVR pipeline on this curriculum.
To make these QA pairs more helpful for improving the solver's capability on the distribution of reasoning tasks we ultimately care about (our \emph{target distribution}), we leverage a small QA dataset sampled from that target, referred to as a development dataset (\textit{dev set}), to anchor the generator's optimization objective.
The document pool supplies candidate curricula; it is not itself the target distribution.
Rather than scoring a question by how difficult it is for the solver, we equip the generator with an \emph{optimizer-aware influence score}, a per-question scalar that quantifies whether training the solver on the candidate question would actually improve its expected reward on the dev set.
This score reduces to the cosine alignment between the solver-side dev-set gradient and the question's solver-side AdamW-induced update direction, and can be efficiently computed from minibatch data.

Leveraging this bilevel game framework with the influence score serving as the generator's reward, we propose \textbf{INFUSER} (\textbf{INF}luence-g\textbf{U}ided
\textbf{S}elf-\textbf{E}volution Improves \textbf{R}easoning), a flexible self-evolution framework where both the generator and solver are trained using policy gradient methods, e.g., variants of GRPO \citep{shao2024deepseekmath}.  
We instantiate INFUSER by optimizing the solving using Dr.GRPO~\citep{liu2025understanding} and propose to train the generator using \textbf{DuGRPO}, a variant of GRPO whose advantage estimator combines group-level and batch-level normalization to accommodate the continuous, noisy nature of the influence reward. \Cref{fig:system_overview} illustrates the resulting data flow.
 
INFUSER delivers strong empirical gains under this design. On Qwen3-8B-Base it attains the top score on every category average (math, general reasoning, medical, and coding) and on $8$ of $14$ individual benchmarks, with relative gains over the base model exceeding $20\%$ on GPQA-Diamond, SuperGPQA, BBEH, AIME, HMMT, and OlympiadBench (Math). 
When comparing Qwen3-4B-Base and Qwen3-8B-Base anchors, we find that INFUSER's gains are much more consistent in model size than other baselines, highlighting the ability to scale self-evolution to larger models.
Notably, an $8$B INFUSER \emph{co-evolving} generator already outperforms a \emph{frozen} $32$B thinking generator significantly on math and coding. A generator-quality analysis further shows that the co-evolving generator produces increasingly well-posed and challenging questions across training, and the solver tracks this rising curriculum, so both players improve under the coupled training loop.

Finally, INFUSER generalizes along two further axes. It continues to improve an already \emph{instruction-finetuned} anchor (OLMo-3-7B-Instruct-SFT), leading on $10$ of $13$ benchmarks versus the fixed generator baseline (\S\ref{sec:instruction_finetune}); and a single INFUSER loop can augment document-grounded self-evolution with rule-verifiable RLVR, eliciting enhanced reasoning depth for better performance on challenging math benchmarks (\S\ref{sec:rlvr_hybrid}).

\paragraph{Related Work.}
INFUSER builds on recent progress in reinforcement learning with verifiable rewards (RLVR) for language-model reasoning. DeepSeekMath introduced GRPO as an efficient RL objective for mathematical reasoning~\citep{shao2024deepseekmath}, and DeepSeek-R1-Zero showed that rule-based RL can elicit long-chain reasoning from a pretrained base model without an SFT cold start~\citep{guo2025deepseek}. Follow-up studies show that this ``zero-style'' RLVR recipe is sensitive to base-model capability, reward design, query difficulty, and training dynamics~\citep{zeng2025simplerl}, while broad-domain systems such as General-Reasoner extend verifiable RL beyond math with large curated problem collections~\citep{ma2025generalreasoner}. This line establishes RLVR as a powerful post-training paradigm, but it still leaves open how to obtain training questions that are both verifiable and useful for the current model.

Self-improvement and self-play methods address this data bottleneck by letting the model generate or select its own training signal. STaR bootstraps reasoning traces through iterative generation and filtering~\citep{zelikman2022star}, while recent self-evolution methods train generators, challengers, or conjecturers to produce tasks near the solver's current capability boundary~\citep{huang2025rzero,zhao2025azr,dong2025stp}. Document-grounded variants such as SPICE further mine corpus environments to produce reasoning tasks from unlabeled text~\citep{fang2025spice}. These approaches make the curriculum adaptive, but the generator is often rewarded by pass-rate, difficulty, or heuristic filtering signals. INFUSER instead asks a more direct question: would training on this generated question improve the solver on the target distribution?

Our answer connects self-evolution with influence-guided data optimization and meta-learning. Classical influence functions measure how training examples affect downstream predictions~\citep{koh2017understanding}, and scalable gradient-alignment methods such as LESS use related signals to select useful instruction-tuning data from an existing pool~\citep{xia2024less, wang2026opus, wang2024greats}. Closely related, GradAlign~\citep{yang2026gradalign} dynamically selects RL training problems by cosine alignment between per-problem policy gradients and an aggregate validation gradient, refreshing these estimates as the policy evolves. Recent synthesis methods train teachers or generators to produce influential data for a target student~\citep{li2024montessori,fan2026optimsyn}. INFUSER differs by jointly co-evolving the generator and solver from the same pretrained model: the solver learns from document-grounded generated QA pairs, while the generator is trained through a bilevel objective approximated by an optimizer-aware influence reward tied to held-out solver performance. This places INFUSER within data-centric meta-learning~\citep{vilalta2002perspective,vanschoren2018meta,hospedales2021meta}, but with an evolving curriculum rather than a fixed synthetic dataset. A detailed discussion is deferred to \S\ref{sec:related_work}.

\section{Method}\label{sec:method}
\noindent \textbf{Notation.}
Throughout this paper, we  write $\theta$ and $\phi$ for the solver and generator parameters. 
Both solver and generator models are initialized from the same pretrained checkpoint, but
maintain separate parameters and optimizer states throughout training.
We let  $q$  denote a question, and let $\{a, a_{\phi}, a^*\}$ denote various answers to the question $q$. 
We write $\mathcal{P}$ for the target distribution over verified QA pairs $(q, a^*)$ that the solver is intended to improve on. In the main experiments, this target is instantiated by a science-reasoning dev set sampled from SuperGPQA Science; broader benchmark suites are used to measure aligned performance and transfer rather than to define the training target. For
nonzero vectors $u$ and $v$, we define
$\mathrm{cossim}(u, v) \coloneqq \langle u, v \rangle /
(\|u\|\,\|v\|)$. See complete notation table in
\S\ref{app:setup_notation}.

\subsection{Game-theoretic Formulation for Self-Evolution}
\label{sec:formulation}
We formulate self-evolution as a bilevel game in which the solver trains on a curriculum of QA pairs proposed by the generator, and the generator is in turn optimized so that the induced solver update improves performance on the target distribution. 
The generator and the solver play the roles of \emph{leader} and \emph{follower}, respectively.
We let $\pi_{\phi}$ and $\pi_{\theta}$ denote generator and solver language models, respectively, where $\phi$ and $\theta$ are parameters. 
Given any question $q$, the solver model $\pi_{\theta}$ outputs an answer $a \sim \pi_{\theta}(\cdot \mid q)$ through the conditional generation of the language model.  
In contrast, the generator $\pi_{\phi}$ takes an unstructured document $d$ as input, and generates a QA pair $(q, a_{\phi})$ based on $d$, i.e., $(q, a_{\phi}) \sim \pi_{\phi}(\cdot \mid d)$. 
Here $q$ is the \emph{generated question} and  $a_\phi$ is the generator's \emph{proposed reference answer}, which may be noisy or even wrong. 
To obtain a curriculum of QA pairs, denoted by $\cQ_\phi$, we sample $(q, a_{\phi})$ from $ \pi_{\phi}$, with the document $d$ chosen from a document pool, denoted by $\mathcal{D}_{\mathrm{doc}}$. 
Here $\mathcal{D}_{\mathrm{doc}}$ contains unstructured texts relevant to the target distribution $\mathcal{P}$, ensuring that the generator is \emph{grounded}.

\begin{wrapfigure}{R}{0.46\textwidth}
  \centering
\captionsetup{font=footnotesize,width=\linewidth}
  % \vspace{-4mm}
  \includegraphics[width=\linewidth]{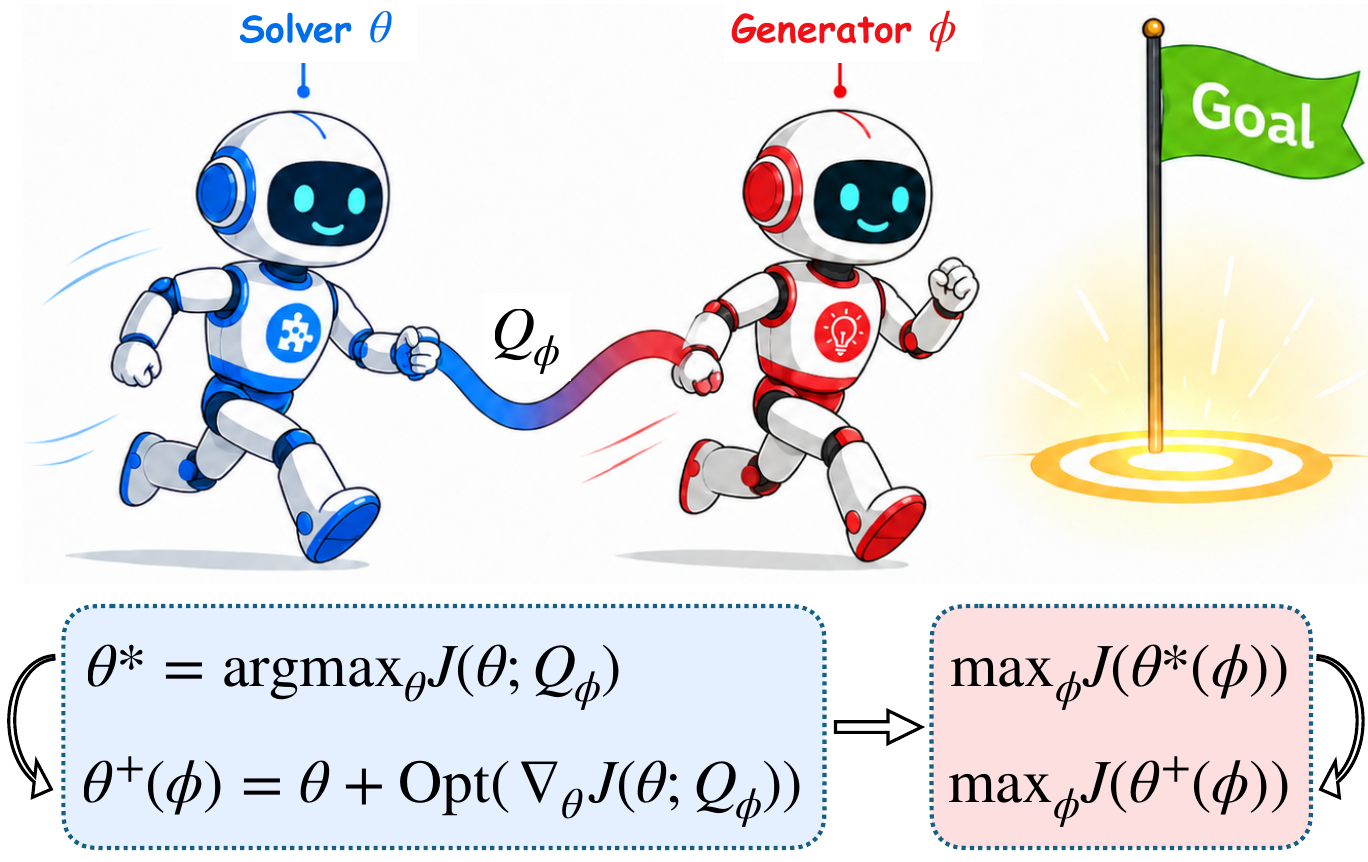}
  % \vspace{-5mm}
  \caption{Bilevel view of the problem: the generator proposes a curriculum, the solver optimizes on it. The generator's goal is to induce a solver optimization that best generalizes to the target distribution. The formula at the bottom illustrates approximations made to \eqref{eq:ideal_bilevel}}
\label{fig:bilevel_illustrate}
  % \vspace{-3mm}
\end{wrapfigure}

In a nutshell, in the bilevel game of self-evolution, the objectives of the solver $\pi_{\theta}$ and generator $\pi_{\phi}$ are as follows:
\begin{itemize}
    \item [(i)] The solver $\pi_{\theta}$ aims to solve the  curriculum of QA data $\cQ_{\phi}$ generated by the generator $\pi_{\phi}$; 
    \item [(ii)] The generator $\pi_{\phi}$ aims to generate $\cQ_{\phi}$ that is beneficial for learning $\mathcal{P}$, in the sense that, after training on $\cQ_{\phi}$, the solver $\pi_{\theta}$ achieves a higher accuracy for solving questions from $\mathcal{P}$. 
\end{itemize}
Moreover,  the solver $\pi_{\theta} $ and generator $\pi_{\phi}$ are initialized from the same language model, trained iteratively at the same time, while interacting with each other. 
In its idealized population form, illustrated in \Cref{fig:bilevel_illustrate}, this bilevel game is mathematically formulated as  

\begin{equation}
  \begin{aligned}
    &\max_{\phi}\quad
    J(\theta^*(\phi))
    \coloneqq
    \mathbb{E}_{{(q, a^*) \sim \cP, \,
                a \sim \pi_{\theta^*(\phi)}(\cdot \mid q)}}
    \bigl[ r(a, a^*; q) \bigr] \\
    &\text{s.t.}\quad
    \theta^*(\phi)
    = \argmax_{\theta}\; J(\theta;\cQ_\phi), 
    \phantom{\text{s.t.}\quad}
    J(\theta;\cQ_\phi)
    \coloneqq
    \mathbb{E}_{{(q, a_\phi) \sim \cQ_\phi,\, 
                a \sim \pi_\theta(\cdot \mid q)}}
    \bigl[ r(a, a_\phi; q) \bigr].
  \end{aligned}
\label{eq:ideal_bilevel}
\end{equation}
Here, $r(a,b;q)\in\{0,1\}$ denotes a binary  \emph{verifiable reward} function, which quantifies whether answer $a$ is correct for question $q$, using $b$ as the \emph{reference}. 
We omit the dependency of unstructured document $d$ in \eqref{eq:ideal_bilevel} to simplify the notation, which is used to generate $\cQ_{\phi}$.

\paragraph{Interpretation of \eqref{eq:ideal_bilevel}.}
In the \textbf{lower level} problem of \eqref{eq:ideal_bilevel}, we generate a QA dataset $\cQ_{\phi}$ using generator $\pi_{\phi}$, and train the solver $\pi_\theta$ by assuming $a_{\phi}$ is the ground truth answer. 
With the binary reward $r$, $J(\theta;\cQ_\phi)$ corresponds to the accuracy of the solver $\pi_{\theta}$ on the curriculum $\cQ_{\phi}$. 
For a fixed generator, the best solver (best-response) is denoted by $\theta^*(\phi)=\argmax_\theta J(\theta;\cQ_\phi)$, which corresponds to the ideal solver fully trained on data generated from $\pi_{\phi}$. 
Fixing the generator, the solver's problem is the same as the standard RLVR problem with data $\cQ_{\phi}$, and thus can be solved using policy-gradient type algorithms \citep{shao2024deepseekmath, guo2025deepseek}.
Furthermore, in the \textbf{upper level} problem of \eqref{eq:ideal_bilevel},
the objective $J(\theta^*(\phi))$ corresponds to the accuracy of the \emph{best-response} solver model on the \emph{target distribution} $\mathcal{P}$. The generator aims to maximize this objective indirectly by designing better $\cQ_{\phi}$. Ideally, if $\cQ_{\phi}$ is close to the $\cP$, then $\theta^*(\phi)$ is close to the best model for $\cP$.

\paragraph{Cooperative by design: rewarding the generator for helping, not hindering.}
In \eqref{eq:ideal_bilevel}, the  generator $\pi_{\phi}$ is  optimized not merely to produce answerable questions, but  more importantly, to induce a solver response that improves target-distribution performance.
The game is ``cooperative'' only in the operational sense that the generator is rewarded for improving the solver rather than defeating it.
We note that the bilevel game in \eqref{eq:ideal_bilevel} is not a cooperative game in the strict sense of game theory, because the generator and solver do not share the same optimization objective --- it is a non-cooperative game where each player has its own objective \citep{bacsar1998dynamic}. 
In particular, the solver only optimizes the generated-reference objective $J(\theta;\cQ_\phi)$, which is a \emph{proxy} for the true target $J(\theta)$, while the generator's job is precisely to keep that proxy faithful, shaping $\cQ_\phi$ so that progress on $J(\theta;\cQ_\phi)$ translates into progress on $J(\theta)$. 
The cooperative nature is achieved by reward design --- the generator is rewarded by improving the solver's performance on the target distribution. 
To achieve such a goal, 
intuitively, we want to ensure (i) the solver learns to solve the curriculum $\cQ_{\phi}$ and (ii) the curriculum $\cQ_{\phi}$ is close to the target distribution $\mathcal{P}$. 
For a perfect generator such that $\cQ_{\phi}$ has the same distribution as $\mathcal{P}$, the two optimization objectives in \eqref{eq:ideal_bilevel} coincide, hence improving the solver also benefits the generator.

\paragraph{Benefits of \eqref{eq:ideal_bilevel} compared with RLVR.} When we have access to the target distribution $\cP$, a direct approach is to train the solver using samples from $\cP$ via RLVR. 
It seems that self-evolution in \eqref{eq:ideal_bilevel} is a detour. We argue that this approach offers two advantages:
\begin{itemize}
    \item [(i)] The solver is never directly trained on $\cP$. Rather, $\cP$ is used as a reference for the generator and the solver is trained on synthetic data based on unstructured texts. 
    Thus, the self-evolution approach requires less golden data than RLVR, which is more appealing when the golden data is costly to obtain. 
   
    \item [(ii)] More importantly, when $\mathcal{P}$ is too challenging for the language model, direct RLVR is challenging. This is because the training signals of policy gradient algorithms such as GRPO~\citep{shao2024deepseekmath} are computed by the \emph{relative advantage} of repeated rollouts. When $\mathcal{P} $ is challenging, most of the generated answers are incorrect, and thus the  training signals are weak, which makes GRPO struggle. 
    In contrast, 
    by bringing a generator into the scope and training the solver using generator's synthetic data, we are able to obtain more meaningful training signals for the solver. This is because the generator can generate easier QA pairs to guide the solver, and gradually increase the difficulty level during self-evolution. 
\end{itemize}

\paragraph{From bilevel formulation to practical training.}
This ideal formulation clarifies the target, but it is not yet a practical training objective.
It overlooks two key aspects of online self-evolution.
First, every generator update would require recomputing the lower-level best response $\theta^*(\phi)$ by training the solver to convergence on the current curriculum $\cQ_\phi$, which is prohibitively expensive at LLM scale.
Second, the best-response view is static: it evaluates a curriculum only after full solver adaptation, overlooking the fact that solver at different stages of training may have different needs, and a curriculum that is good for the final adapted solver may not be good for the solver during the course of optimization.
This means the generator also needs to  \textit{co-evolve} with the solver. 
We therefore replace the ideal population game with a turn-based, myopic one-step objective that encompasses the above-mentioned considerations.

\vspace{1mm}
\noindent\textcolor{sectionblue}{\(\triangleright\)\,\textbf{\textit{Per-iteration lower level (solver).}}}
We consider the question generation process to be document-conditioned. 
At each iteration, the generator samples documents $d \sim \mathcal{D}_{\mathrm{doc}}$ and produces self-generated QA pairs
$(q, a_\phi) \sim \pi_\phi(\cdot \given d)$, forming a minibatch from $\cQ_\phi$ (see \S\ref{app:document_pipeline} for the construction of $\mathcal{D}_{\mathrm{doc}}$).
The generated pair is checked for parseable QA format, but the document is not a formal verifier for the factual correctness of $a_\phi$.
The solver then takes a single RL update on this batch, defining the \emph{solver-update map} $\theta^+(\phi)$:
\begin{equation}
  \theta^+(\phi)
  =
  \theta + \Delta\theta(\phi),
  \qquad
	  \Delta\theta(\phi)
	  = \mathrm{Opt}\bigl(
	      \nabla_\theta J(\theta;\, \mathcal{Q}_\phi)
	    \bigr),
	  \label{eq:solver_step}
\end{equation}
where $\mathrm{Opt}(\cdot)$ is the optimizer update rule (e.g., AdamW) and $\nabla_\theta J(\theta;\, \mathcal{Q}_\phi)$ is the solver's policy gradient computed using minibatch $\mathcal{Q}_{\phi}$. 
Under \eqref{eq:solver_step}, the generator's curriculum $\cQ_\phi$ induces a \emph{solver step} $\theta^+(\phi)$.
This approximation avoids the cost of full convergence to $\theta^*(\phi)$, while maintaining a useful coupling between the generator's curriculum and the solver's optimization trajectory.
In the next part we will see how $\theta^+(\phi)$ can be used to derive a practical training objective for the generator.

\vspace{1mm}
\noindent\textcolor{sectionblue}{\(\triangleright\)\,\textbf{\textit{Per-iteration upper level (generator).}}}
The ideal upper level depends on the exact best response $\theta^*(\phi)$, but INFUSER only has the one-step adapted solver $\theta^+(\phi)$ from \eqref{eq:solver_step}.
Replacing $\theta^*(\phi)$ with this practical update map gives the generator objective:
\begin{equation}
  \max_{\phi}\;
  J(\theta^+(\phi))
  \quad
  \text{s.t.}\quad
  \theta^+(\phi)\ \text{is given by \eqref{eq:solver_step}}.
  \label{eq:lower}
\end{equation}
In this reduced problem, the solver updates its parameters $\theta$ using the generator's curriculum to obtain $\theta^+(\phi)$, while the generator updates $\phi$ to shape a curriculum whose induced solver step best improves target-distribution performance, as measured by $J(\theta^+(\phi))$.
This formulation precisely captures the nested learning nature for the generator, where the influence of the generator's curriculum on the performance is mediated through the solver-update map $\theta^+(\phi)$.
Since the population objective $J(\theta^+(\phi))$ is not directly computable, \S\ref{sec:loop} instantiates it with a held-out development-set surrogate. That is, we replace the expectation with respect to $\mathcal{P}$ in  \eqref{eq:ideal_bilevel} by the empirical mean over a fixed dev set sampled from $\mathcal{P}$.  See \S\ref{sec:loop} for details. 
 
\paragraph{Challenges in solving \eqref{eq:lower}.}
Even if we simplify the inner optimization to a single RL step, directly solving \eqref{eq:lower} via first-order methods is still challenging. 
The main challenge lies in the fact that $\nabla_{\phi} J(\theta^+(\phi))$ requires $\nabla_{\phi} \theta^+(\phi)$, which is hard to compute.
Two typical ways to solve \eqref{eq:lower} are black-box outer-loop search and exact meta-gradient optimization.
Both are impractical for LLM-scale online curriculum learning: the former reruns the full inner update for each generator proposal, while the latter backpropagates through the solver update as in MAML-style bilevel optimization~\citep{finn2017model}.
We defer details on these algorithms to \S\ref{app:alternatives}.
In the next subsection, we adopt a first-order approximation that turns the outer objective \eqref{eq:lower} into an influence-guided learning signal for the generator.

\subsection{Self-evolution through Influence-Guided Optimization}
\label{sec:influence}

To avoid the cost of exact bilevel differentiation in solving \eqref{eq:lower}, we adopt a
first-order approximation inspired by influence
functions~\citep{v_Mises_1947,Hampel_1974,koh2017understanding}.
Consider the first-order Taylor expansion of the outer
objective $J(\theta^+(\phi))$ around the current solver parameters $\theta$:
\begin{equation}
  J(\theta^+(\phi))
  \approx
  J(\theta)
  +
  \bigl\langle
    \nabla_\theta J(\theta),\;
    \Delta\theta(\phi)
  \bigr\rangle.
  \label{eq:taylor}
\end{equation}
We use this as a local first-order approximation: RL fine-tuning
typically uses very small learning rates (order of $10^{-6}$), making
the single-step update $\|\Delta\theta(\phi)\|$ small, while the neglected
term is second order in the update norm.
See the bound in \S\ref{app:influence_derivation}.
Since $J(\theta)$ does not depend on $\phi$, the generator can optimize the inner product term as a proxy for improving $J(\theta^+(\phi))$.

\paragraph{SGD example: influence scores as reward.}
Temporarily supposing $\mathrm{Opt}(\cdot)$ is the vanilla SGD update, we can decompose
$\langle \nabla_\theta J(\theta), \Delta\theta(\phi) \rangle$ over
the individual questions in the curriculum $\mathcal{Q}_\phi$ to obtain
a per-question score that the generator can optimize via policy gradients.
With a little abuse of notation, writing
 $g(q, a_\phi) = \nabla_\theta J(\theta;\, q, a_\phi)$ as the policy gradient induced by question $(q, a_\phi)\in \cQ_\phi$, SGD with learning rate $\eta_s$ on the mean batch loss gives
\begin{equation}
  \bigl\langle
    \nabla_\theta J(\theta),\;
    \Delta\theta(\phi)
  \bigr\rangle
  =
  {
    % \textstyle
    \frac{\eta_s}{|\mathcal{Q}_\phi|}
    \sum_{(q, a_\phi) \in \mathcal{Q}_\phi}
    \bigl\langle
      \nabla_\theta J(\theta),\;
      g(q, a_\phi)
    \bigr\rangle
  }.
  \label{eq:sgd_decomp}
\end{equation}
Each question's contribution is captured by the per-sample inner
product $\langle \nabla_\theta J(\theta),\, g(q, a_\phi) \rangle$, the
classical influence function, which gives a clean per-question score.
Crucially, this score depends only on the static gradient $\nabla_\theta J(\theta)$ and the generated question--answer pair $(q, a_\phi)$, so it can be treated as standard \textbf{per-sample reward} for the generator and optimized with policy gradients, sidestepping the need to differentiate through the solver update map $\Delta\theta(\phi)$.

\paragraph{Optimizer-aware influence score.}
For the AdamW solver optimizer, we summarize the update induced by each
generated pair with an optimizer-preconditioned per-question direction
$\Gamma(q,a_\phi)$. This gives the following score.

\begin{definition}[Optimizer-aware influence score]\label{def:influence_score}
For solver parameter $\theta$, let $J(\theta)$ denote the target
performance. For a generated pair $(q, a_\phi)$, define the population
single-question solver objective and its gradient as
\begin{equation}
  J(\theta; q, a_\phi)
  \coloneqq
  \mathbb{E}_{a \sim \pi_\theta(\cdot \mid q)}[r(a, a_\phi; q)],
  \qquad
  g(q, a_\phi)
  \coloneqq
  \nabla_\theta J(\theta; q, a_\phi).
  \label{eq:single_question_solver_gradient}
\end{equation}
Let $\Gamma(q, a_\phi)$ denote the AdamW-preconditioned per-question
solver update direction induced by $g(q,a_\phi)$, holding the current
solver optimizer state fixed. The exact preconditioning rule and its
connection to AdamW are given in \S\ref{app:adamw_preconditioning}. The
optimizer-aware influence score for generated $(q, a_\phi)$ at solver
state $\theta$ is then
\begin{equation}
  s(q, a_\phi)
  \;\coloneqq\;
  \mathrm{cossim} \bigl(
    \nabla_\theta J(\theta),\;
    \Gamma(q, a_\phi)
  \bigr).
  \label{eq:influence_score}
\end{equation}
\end{definition}
Here, $\Gamma(q, a_\phi)$ plays the role of $g(q, a_\phi)$ in the SGD decomposition~\eqref{eq:sgd_decomp}, so a positive cosine in~\eqref{eq:influence_score} means training on $q$ is expected to improve $J(\theta)$ (see \S\ref{app:influence_derivation} for the derivation of $\Gamma$).
We use cosine similarity rather than a raw inner product to avoid a spurious correlation between sequence length and gradient norm~\citep{xia2024less}.

\paragraph{Influence score as generator's RL reward.}
As in the SGD case, $s(q, a_\phi)$ then serves directly as the generator's RL reward, with the generator facing the following optimization problem:
\begin{equation}
  \max_{\pi_\phi} \mathbb{E}_{d \sim \mathcal{D}_{\mathrm{doc}},\, (q, a_\phi) \sim \pi_\phi(\cdot \mid d)}[s(q, a_\phi)].
  \label{eq:optimizer_aware_influence}
\end{equation}
This optimizer-aware influence score therefore serves as a dense signal that judges whether the generated questions are useful for the solver at its current state, without requiring explicit optimizer differentiation or expensive black-box search.
In the main algorithm, we only use $s(q, a_\phi)$ as the generator's scalar reward: each generated question is rated by its usefulness to the current solver, and the generator is updated to produce questions with higher influence scores.
For completeness, \S\ref{app:influence_reinforce} gives the corresponding REINFORCE estimator for \eqref{eq:optimizer_aware_influence}; the actual generator update used by INFUSER is the DuGRPO update in \S\ref{sec:loop}, which normalizes these continuous influence rewards.

\subsection{INFUSER: Co-evolving Generator and Solver with Influence-Guided RL}
\label{sec:loop}

\begin{figure}[!t]
  \vspace{-2mm}
  \centering
  \includegraphics[width=0.8\linewidth]{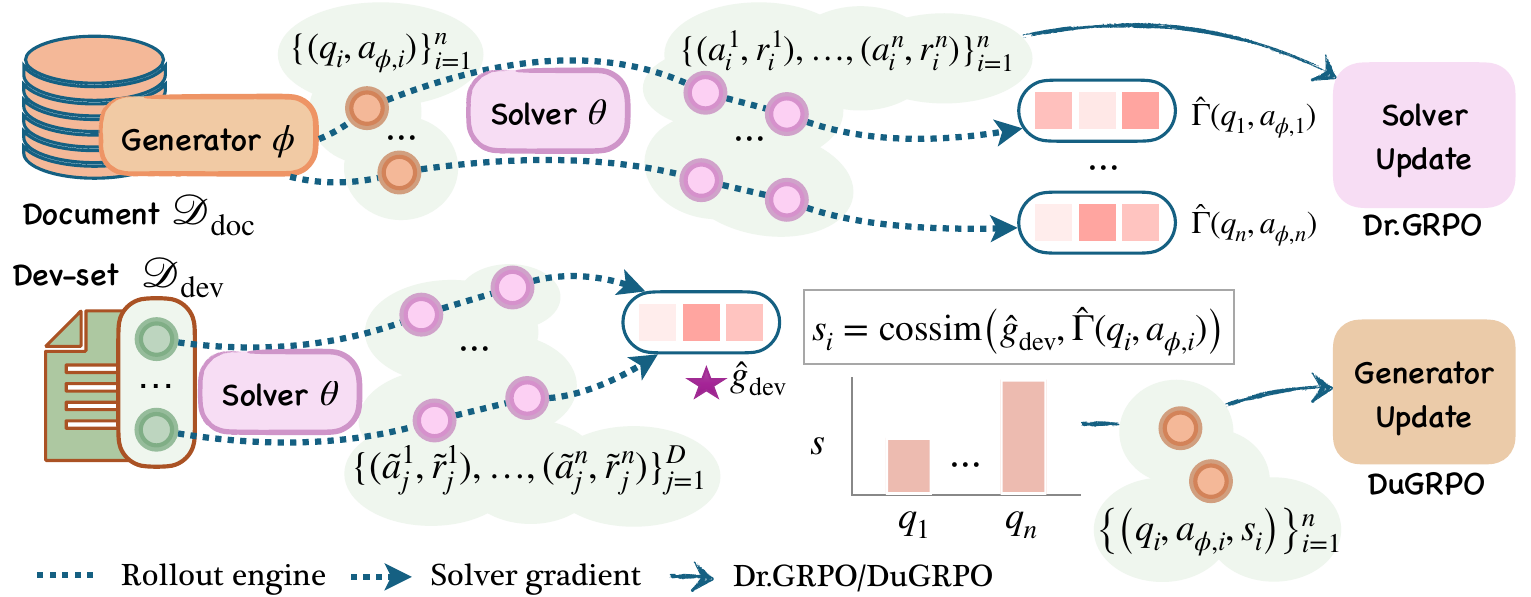}
  \caption{\small Detailed data flow for INFUSER. \textbf{Top:} In
    Phase 2, the generator produces self-generated QA pairs from
    documents, and in phase 3, the solver produces answers to
    these questions and receives binary rewards. Each question also
    receives a solver-side AdamW update direction $\hat\Gamma(q_i, a_{\phi, i})$.
    \textbf{Bottom Left:} In Phase 1, the solver
   produces answers to dev set questions, which are used to
    compute the solver-side reference gradient $\hat g_{\mathrm{dev}}$. \textbf{Bottom
    Right:} In Phase 4, the influence scores are computed as the cosine similarity
    $\mathrm{cossim}(\hat g_{\mathrm{dev}}, \hat\Gamma(q_i, a_{\phi, i}))$ and used
    as rewards for the generator update. The solver's answers and rewards from Phase 3 are used for the solver update.}
  \label{fig:system_overview_detailed}
  \vspace{-3mm}
\end{figure}

We now propose INFUSER (see \Cref{alg:infuser}), a practical online algorithm that co-evolves the generator and solver using the one-step influence approximation above.
At each iteration, INFUSER estimates a solver-side dev-set target direction, scores generated questions by their alignment with that direction, and alternates generator and solver RL updates on the resulting curriculum.
The main components are therefore a dev-set-based empirical influence estimate, alternating solver--generator optimization, and a Dual-normalized Group Relative Policy Optimization (DuGRPO) update that stabilizes generator learning from continuous influence rewards.
We detail these ingredients below.

\begin{algorithm}[!ht]
\small
\caption{INFUSER (Detailed version in \Cref{alg:infuser_full})}
\label{alg:infuser}
\DontPrintSemicolon
\KwIn{Pretrained LLM (for initializing both $\pi_\theta$ and $\pi_\phi$), a document pool $\mathcal{D}_{\mathrm{doc}}$, a small fixed dev set $\mathcal{D}_{\mathrm{dev}}$, batch size $B$, group size $n$, maximum training iterations $T$, and invalid question penalty $\rho_{\mathrm{inv}}$.}
\KwOut{A solver improved by training on a generator-adapted curriculum.}
\For{training loop $1, \dots, T$}{
\BlankLine
  \textbf{Phase 1: Ask what is the improvement direction.}
  Run the current solver on $\mathcal{D}_{\mathrm{dev}}$ with $n$ rollouts and compute a solver-side dev reference gradient $\hat g_{\mathrm{dev}}$ by \eqref{eq:hat_g_dev}.\;
\BlankLine

  \textbf{Phase 2: Ask the generator for candidate question-answer pairs.}
  Sample $B$ documents from $\mathcal{D}_{\mathrm{doc}}$. For each sampled document $d$, have the generator write $n$ question-answer pairs $(q, a_\phi) \sim \pi_\phi(\cdot\given d)$. Filter out invalid questions with format issue to obtain curriculum $\cQ_\phi$.\;
\BlankLine

  \textbf{Phase 3: Test how each candidate would train the solver.}
  For each $(q, a_\phi)\in\cQ_\phi$, run the solver on each generated question $q$ for $n$ times, score its answers against the reference answer $a_\phi$, and compute the per-question solver update direction $\hat g(q, a_\phi)$ by \eqref{eq:hat_g_q_a_phi}.\;
\BlankLine

  \textbf{Phase 4: Reward questions by optimizer-aware influence score.}
  For each $(q, a_\phi)\in\cQ_\phi$, compute its AdamW update direction $\hat\Gamma(q, a_\phi)$ from $\hat g(q, a_\phi)$, and assign reward $\hat s(q, a_\phi)= \mathrm{cossim}(\hat g_{\mathrm{dev}}, \hat\Gamma(q, a_\phi))$. For each invalid question, assign penalty $\rho_{\mathrm{inv}}$.\;
\BlankLine

  \textbf{Phase 5: Improve the generator.}
  Treat $n$ sampled questions from the same document as a group, update the generator with per question rewards computed in Phase 4 and the DuGRPO advantage in \eqref{eq:gen_advantage}.\;
\BlankLine

  \textbf{Phase 6: Improve the solver.}
  Treat $n$ sampled answers from the same question as a group, update the solver on curriculum $\cQ_\phi$ using solver rollouts and scores obtained in Phase 3 and apply Dr.GRPO.\;
}
\Return{the trained solver $\pi_\theta$.}
\end{algorithm}

\FloatBarrier

\paragraph{Dev-set-based influence score estimate.}
The ideal outer objective $J$ averages over the full target distribution $\mathcal{P}$, but INFUSER only needs a local direction that tells the current solver what ``improving on the target task'' means.
We estimate this direction from a \emph{small fixed} development set $\mathcal{D}_{\mathrm{dev}}=\{(q_i, a_i^*)\}_i$ sampled from $\mathcal{P}$, replacing $J$ with the empirical surrogate
\begin{equation}
  \hat{J}(\theta)
  \;=\;
  {
    % \textstyle
    \frac{1}{|\mathcal{D}_{\mathrm{dev}}|}
    \sum_{(q, a^*) \in \mathcal{D}_{\mathrm{dev}}}
    \mathbb{E}_{a \sim \pi_\theta(\cdot \mid q)}
    [r(a, a^*; q)]
  },
  \label{eq:dev_surrogate}
\end{equation}
which is made fully computable by the rollout-based estimator in \eqref{eq:unified_obj} below.
At the beginning of each iteration, we roll out the current solver on $\mathcal{D}_{\mathrm{dev}}$ and compute the solver-side gradient $\nabla_\theta \hat{J}(\theta)$.
This solver-side gradient is then held fixed as the reference direction in the influence score \eqref{eq:influence_score}: a generated question is useful when its induced solver-update direction aligns with this dev-improving direction.

The dev set therefore acts as an anchor for credit assignment, not as solver training data or generator prompt context.
The solver update is performed on generated curriculum questions, and the generator is not prompted with dev questions.
Since this anchor defines what counts as useful for the solver, $\mathcal{D}_{\mathrm{dev}}$ should reflect the target question style; we analyze its effect on the resulting curriculum in \S\ref{sec:curriculum_analysis} and \S\ref{sec:comparison}.

\paragraph{Training objectives for solver--generator co-evolution.} 
So far, we have described the population quantities that define INFUSER.
In implementation, four quantities are estimated from rollouts: the solver-side dev reference gradient $\hat g_{\mathrm{dev}}$, the per-question solver direction used to form $\hat{\Gamma}(q,a_\phi)$, the actual solver update gradient, and the generator update gradient.
The first three are solver-side estimates, all differentiated with respect to $\theta$; the last is a generator-side estimate differentiated with respect to $\phi$.
They all use the same clipped rollout objective, differing only in the policy, input, sampled outputs, and advantage.

Let us take $\psi$ to represent $\theta$ or $\phi$, let $\pi_\psi$ be a policy with input $z$, let $\{x_1,\dots,x_n\}$ be a group of outputs sampled from the rollout policy $\pi_{\psi_{\mathrm{old}}}(\cdot\mid z)$.
We estimate the corresponding gradient by plugging the row-specific advantage $\hat A_i$ into
\begin{equation}
  \mathcal{J}_\psi(z)
  =
  {
    % \textstyle
    % \mathbb{E}_{(x_1, \dots, x_n)\sim \pi_{\psi_{\mathrm{old}}}(\cdot\mid z)}\biggl[
      \frac{1}{n}\sum_{i=1}^{n}
      \frac{1}{C}\sum_{t=1}^{|x_i|}
      \min\,\bigl\{
        \rho_{i,t}\cdot\hat{A}_i,\;
        \mathrm{clip}(\rho_{i,t},\,1{-}\epsilon,\,1{+}\epsilon)\cdot\hat{A}_i
      \bigr\}
    % \biggr]
  },
  \label{eq:unified_obj}
\end{equation}
where $\rho_{i,t} = \pi_\psi(x_{i,t} \mid z,x_{i,<t})
/ \pi_{\psi_{\mathrm{old}}}(x_{i,t} \mid z,x_{i,<t})$ is the
token-level importance sampling ratio, $n$ is the group size, $\hat A_i$ is the advantage, $\epsilon$ is the clipping hyperparameter, and $C$ is a fixed maximum generation length. The factor $1/C$ replaces GRPO's per-response normalization $1/|x_i|$, removing the length bias identified by~\citet{liu2025understanding} in Dr.GRPO.
Throughout training, we use the same group size $n=8$ for both the solver and generator.
We note that if $\hat A_i$ are the unnormalized advantage, this target coincides with the Dr.GRPO target.

\begin{table}[t]
\centering
\caption{Sample-based gradient estimators in INFUSER. Each row is obtained by instantiating \eqref{eq:unified_obj} with the listed policy, input, sampled output group, and advantage.}
\label{tab:infuser_estimators}
\small
\setlength{\tabcolsep}{3pt}
\renewcommand{\arraystretch}{1.15}
\begin{tabularx}{\linewidth}{@{}>{\raggedright\arraybackslash}p{0.18\linewidth}>{\raggedright\arraybackslash}p{0.12\linewidth}>{\raggedright\arraybackslash}p{0.20\linewidth}>{\raggedright\arraybackslash}p{0.20\linewidth}>{\raggedright\arraybackslash}X@{}}
\toprule
Quantity & Policy & Input $z$ & Output group $x_i$ & Advantage $\hat A_i$ \\
\midrule
Dev reference $\hat g_{\mathrm{dev}}$
& Solver $\pi_\theta$
& Dev pair $(\tilde q,\tilde a^*)$
& Solver answers $\tilde a^i \sim \pi_\theta(\cdot\mid\tilde q)$
& $\tilde r_i-\frac{1}{n}\sum_j\tilde r_j$, where $\tilde r_i=r(\tilde a^i,\tilde a^*;\tilde q)$ \\
Per-question direction $\hat\Gamma(q,a_\phi)$
& Solver $\pi_\theta$
& Generated pair $(q,a_\phi)$
& Solver answers $a^i \sim \pi_\theta(\cdot\mid q)$
& $r_i-\frac{1}{n}\sum_j r_j$, where $r_i=r(a^i,a_\phi;q)$ \\
Solver update
& Solver $\pi_\theta$
& Retained generated pair $(q,a_\phi)$
& Same solver answers $a^i$
& Same mean-centred solver advantage as above \\
Generator update
& Generator $\pi_\phi$
& Document $d$
& QA pairs $(q_i,a_{\phi,i})\sim\pi_\phi(\cdot\mid d)$
& DuGRPO-normalized influence reward, defined in \eqref{eq:gen_advantage} \\
\bottomrule
\end{tabularx}
\end{table}

For any reference pair $(q,b)$, let $\cJ_\theta(q,b)$ denote \eqref{eq:unified_obj} instantiated with the solver policy, input question $q$, solver answers sampled from $\pi_\theta(\cdot\mid q)$, binary rewards $r(a^i,b;q)$, and the mean-centred solver advantage in \Cref{tab:infuser_estimators}.
The first row of the table gives the dev reference direction
\begin{equation}
  \hat{g}_{\mathrm{dev}}
  =
  \frac{1}{|\mathcal{D}_{\mathrm{dev}}|}
  \sum_{(\tilde q, \tilde a^*) \in \mathcal{D}_{\mathrm{dev}}}
  \nabla_\theta \cJ_\theta(\tilde q, \tilde a^*)
  \approx \nabla_\theta \hat J(\theta),
  \label{eq:hat_g_dev}
\end{equation}
computed once per iteration.
For each generated pair, the second row of \Cref{tab:infuser_estimators} gives the finite-rollout gradient 
\begin{equation} 
  \hat g(q,a_\phi)=\nabla_\theta\cJ_\theta(q,a_\phi),
  \label{eq:hat_g_q_a_phi}
\end{equation} 
which estimates $g(q,a_\phi)$ in \eqref{eq:single_question_solver_gradient}; applying the AdamW preconditioning from \S\ref{app:adamw_preconditioning} yields the solver-side optimizer-aware update direction $\hat{\Gamma}(q,a_\phi)$.
The empirical influence reward for the generator is then
\begin{equation}
  \hat{s}(q, a_\phi) = \mathrm{cossim}\bigl(\hat{g}_{\mathrm{dev}},\, \hat{\Gamma}(q, a_\phi)\bigr).
  \label{eq:estimated_influence_score}
\end{equation}
The third row uses the same solver-side objective to update $\theta$ on retained generated questions, while the fourth row plugs the generator policy into \eqref{eq:unified_obj} and uses $\hat{s}(q,a_\phi)$ as the reward.

\begin{wrapfigure}{R}{0.3\textwidth}
  \centering
  \captionsetup{font=footnotesize}
  % \vspace{-4mm}
  \includegraphics[width=\linewidth]{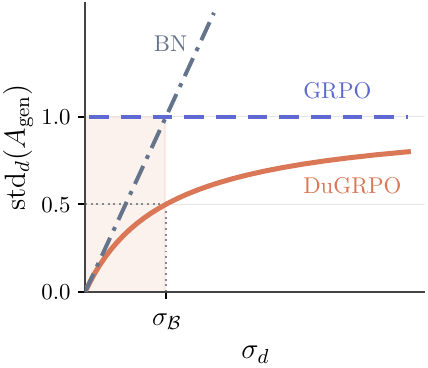}
  % \vspace{-5mm}
  \caption{Within-group advantage std w.r.t. \ $\sigma_d$
    (fix $\sigma_{\mathcal{B}}$).}
  \label{fig:dugrpo_std}
  % \vspace{-5mm}
\end{wrapfigure}

\paragraph{Dual-normalized generator advantage (DuGRPO).}
Using influence scores as generator rewards poses a distinct normalization challenge: unlike binary correctness rewards, these rewards are continuous, noisy, and estimated from finite solver rollouts.
Directly reusing existing RLVR advantage normalizers creates two issues:
\begin{enumerate}[leftmargin=*,topsep=2pt,itemsep=2pt]
  \item \emph{GRPO-style noise amplification.} Normalizing each document group by its own $\sigma_d$ forces even low-variance groups to have unit-scale advantages, amplifying rollout noise when the generated questions have nearly indistinguishable influence scores.
  \item \emph{Dr.GRPO-style high-variance domination.} Using the raw mean-centered advantage without normalization avoids the previous amplification, but lets high-variance document groups dominate the generator gradient.
\end{enumerate}
For a document $d$, let $\mu_d$ and $\sigma_d$ denote the mean and
standard deviation of the influence scores
$\{s(q^k,a_{\phi}^k)\}_{k=1}^n$ generated from $\pi_{\phi}(\cdot\mid d)$. For a document
batch $\mathcal{B}$, define
$\sigma_{\mathcal{B}}\coloneqq
\mathrm{mean}\{\sigma_{d'}:d'\in\mathcal{B}\}$. 
DuGRPO addresses both issues by
keeping the within-group normalizer $\sigma_d$ but adding a cross-group normalizer $\sigma_{\mathcal{B}}$ that can adapt to the overall advantage spread in the batch:
\begin{equation}
  \hat A_{\mathrm{gen}}(q^k, a_{\phi}^k)
  :=
  {
    %  \textstyle
    \frac{s(q^k, a_{\phi}^k) - \mu_d}{\sigma_d + \sigma_{\mathcal{B}} + \epsilon}
  },
  \quad k = 1, \dots, n.
  \label{eq:gen_advantage}
\end{equation}
Its within-group standard deviation is
$\mathrm{std}_d(A_{\mathrm{gen}})=
\sigma_d/(\sigma_d+\sigma_{\mathcal{B}}+\epsilon)$.  
As we show in \Cref{fig:dugrpo_std}, DuGRPO elegantly damps the low-variance
groups ($\sigma_d<\sigma_{\mathcal{B}}$ with no significant advantage) and roughly maintains the unit-spread benefit of GRPO when $\sigma_d>\sigma_{\mathcal{B}}$. 
Two ablation normalizers plotted in \Cref{fig:dugrpo_std} and will be compared in \Cref{sec:comparison} are GRPO-style $\hat{A}_{\mathrm{GRPO}}$ and batch-normalized $\hat{A}_{\mathrm{BN}}$:
\begin{equation}
  \begin{aligned}
  {
    % \textstyle
    \hat A_{\mathrm{GRPO}}(q^k, a_{\phi}^k)
    :=
    \frac{s(q^k, a_{\phi}^k) - \mu_d}{\sigma_d + \epsilon},\qquad
    \hat A_{\mathrm{BN}}(q^k, a_{\phi}^k)
    :=
    \frac{s(q^k, a_{\phi}^k) - \mu_d}{\sigma_{\mathcal{B}} + \epsilon}
  }.
  \end{aligned}
  \label{eq:group_batch_std_advantages}
\end{equation}

\paragraph{Other algorithm components.}
Two implementation details determine which rollouts contribute gradients.
First, we remove only zero-variance groups: document groups with identical influence rewards yield no generator advantage, and question groups with identical correctness rewards yield no solver advantage.
Invalid generated questions are excluded from solver rollouts and influence scoring; in the reported INFUSER runs, their generator-side penalty is set to $\rho_{\mathrm{inv}}=0$.
This choice treats invalid questions as having zero influence because they neither help nor hurt the solver update, and it still discourages invalid generations: valid questions that are useful for training can receive positive influence rewards and therefore win in the generator update.
After variance filtering, the solver trains on the retained generated questions using its standard correctness reward.
Second, the generator update is applied before the solver update on the same rollout batch, so generator credit assignment is based on the current solver state and all dev directions and influence scores are recomputed at the next iteration.
Both models use minibatch size $M=32$, so a retained rollout batch can yield multiple optimizer steps.
To control the resulting off-policy drift from the rollout policy $\pi_{\psi_{\mathrm{old}}}$, we apply token-level Truncated Importance Sampling~\citep{yao2025offpolicyrl}, clipping the current-policy-to-rollout-policy ratio at $\rho_{\max}=2.0$ when evaluating \eqref{eq:unified_obj}.

\section{Experiments}\label{sec:experiments}
\subsection{Training Setup and Benchmark Evaluation}
\label{sec:exp:main_benchmarks}

{\captionsetup{font=small}
\begin{table}[!t]
  \centering
  \caption{%
    Solver accuracy (\%) on held-out benchmarks for Qwen3-4B-Base
    and Qwen3-8B-Base.
    Bolded entries mark the best score among the five training methods
    per row.  Dashes indicate the benchmark was not reported by that method.
    The Base, INFUSER, R-Zero, and AZR columns are trained (where
    applicable) and evaluated by us under the unified harness in
    \S\ref{app:eval_protocol}; daggered columns
    (R-Few\textsuperscript{$\dagger$}, SPICE\textsuperscript{$\dagger$})
    are self-reported numbers taken from the original papers.
    INFUSER entries report the unified-report aggregate mean $\pm$ sample
    standard deviation recomputed from the same three seeded training runs
    (using denominator $n-1$); all other entries are point estimates.
  }
  \label{tab:main_benchmarks}
  \resizebox{\textwidth}{!}{%
  \setlength{\tabcolsep}{4.5pt}%
  \begin{tabular}{l !{\hspace{0.55em}\vrule\hspace{0.55em}} cccccc !{\hspace{0.55em}\vrule\hspace{0.55em}} cccccc}
    \toprule
     & \multicolumn{6}{c}{\textbf{Qwen3-4B-Base}}
     & \multicolumn{6}{c}{\textbf{Qwen3-8B-Base}} \\
    \cmidrule(lr){2-7} \cmidrule(lr){8-13}
    Benchmark
      & \rotatebox{45}{Base} & \rotatebox{45}{INFUSER} & \rotatebox{45}{R-Zero} & \rotatebox{45}{AZR} & \rotatebox{45}{R-Few\textsuperscript{$\dagger$}} & \rotatebox{45}{SPICE\textsuperscript{$\dagger$}}
      & \rotatebox{45}{Base} & \rotatebox{45}{INFUSER} & \rotatebox{45}{R-Zero} & \rotatebox{45}{AZR} & \rotatebox{45}{R-Few\textsuperscript{$\dagger$}} & \rotatebox{45}{SPICE\textsuperscript{$\dagger$}} \\
    \midrule
    \rowcolor{gray!15}\multicolumn{13}{l}{\textit{General reasoning}} \\
    MMLU-Pro & 52.98 & $\mathbf{60.20 \pm 0.48}$ & 55.80 & 57.53 & 56.20 & 58.10 & 59.91 & $\mathbf{66.20 \pm 1.64}$ & 61.82 & 62.32 & 63.20 & 65.00 \\
    GPQA-Diamond & 31.41 & $36.80 \pm 2.25$ & 34.44 & 37.17 & \textbf{39.90} & 39.40 & 36.87 & $45.48 \pm 2.13$ & 42.73 & 44.14 & \textbf{46.50} & 39.40 \\
    SuperGPQA & 25.88 & $\mathbf{33.48 \pm 1.17}$ & 28.40 & 28.31 & 29.40 & 30.20 & 30.62 & $\mathbf{37.77 \pm 1.30}$ & 32.06 & 32.63 & 33.50 & 35.70 \\
    BBEH & 7.57 & $11.22 \pm 1.19$ & 10.06 & 8.70 & 11.80 & \textbf{12.30} & 10.30 & $13.04 \pm 1.06$ & 11.93 & 11.33 & 12.30 & \textbf{14.90} \\
    \rowcolor{findingbg}\hspace{1em}\emph{Category average} & 29.46 & $\mathbf{35.43 \pm 0.26}$ & 32.18 & 32.93 & 34.33 & 35.00 & 34.43 & $\mathbf{40.62 \pm 0.97}$ & 37.14 & 37.61 & 38.88 & 38.75 \\
    \rowcolor{findingbg}\hspace{1em}\emph{Rel.\ improv.\ over Base (\%)} & --- & $\mathbf{+20.26}$ & $+9.23$ & $+11.78$ & $+16.53$ & $+18.81$ & --- & $\mathbf{+17.98}$ & $+7.87$ & $+9.24$ & $+12.92$ & $+12.55$ \\
    \midrule
    \rowcolor{gray!15}\multicolumn{13}{l}{\textit{Math \& physics reasoning}} \\
    MATH500 & 61.20 & $76.65 \pm 1.10$ & 76.85 & 73.90 & \textbf{78.00} & \textbf{78.00} & 76.05 & $\mathbf{82.77 \pm 3.47}$ & 80.55 & 80.95 & 82.60 & 79.40 \\
    AIME2024 & 10.42 & $11.35 \pm 0.47$ & 9.38 & \textbf{13.54} & --- & 12.20 & 12.92 & $18.58 \pm 2.94$ & 13.96 & \textbf{19.48} & --- & 18.40 \\
    AIME2025 & 8.44 & $10.73 \pm 0.89$ & 7.19 & 13.75 & --- & \textbf{19.10} & 11.87 & $15.87 \pm 3.03$ & 13.33 & 14.17 & --- & \textbf{18.20} \\
    HMMT & 2.49 & $2.94 \pm 0.24$ & 2.65 & \textbf{4.50} & --- & --- & 2.96 & $\mathbf{7.04 \pm 3.32}$ & 3.86 & 6.01 & --- & --- \\
    OlympiadBench (Math) & 35.31 & $42.38 \pm 0.31$ & \textbf{43.18} & \textbf{43.18} & 42.80 & 42.70 & 40.36 & $\mathbf{50.24 \pm 4.94}$ & 45.10 & 47.92 & 46.40 & 42.50 \\
    OlympiadBench (Phys) & 10.17 & $10.31 \pm 2.09$ & \textbf{11.44} & 10.17 & --- & --- & 12.29 & $\mathbf{14.41 \pm 0.43}$ & 13.98 & 13.14 & --- & --- \\
    \rowcolor{findingbg}\hspace{1em}\emph{Category average} & 21.34 & $25.73 \pm 0.27$ & 25.12 & \textbf{26.51} & --- & --- & 26.08 & $\mathbf{31.49 \pm 2.80}$ & 28.46 & 30.28 & --- & --- \\
    \rowcolor{findingbg}\hspace{1em}\emph{Rel.\ improv.\ over Base (\%)} & --- & $+20.57$ & $+17.71$ & $\mathbf{+24.23}$ & --- & --- & --- & $\mathbf{+20.74}$ & $+9.13$ & $+16.10$ & --- & --- \\
    \midrule
    \rowcolor{gray!15}\multicolumn{13}{l}{\textit{Medical}} \\
    MedQA & 55.46 & $58.86 \pm 0.68$ & 58.92 & \textbf{59.62} & --- & --- & 64.18 & $\mathbf{65.78 \pm 0.67}$ & 65.12 & 65.28 & --- & --- \\
    MedXpertQA & 13.02 & $13.78 \pm 0.31$ & \textbf{14.57} & 12.65 & --- & --- & 14.49 & $\mathbf{15.25 \pm 0.72}$ & 15.22 & 14.49 & --- & --- \\
    \rowcolor{findingbg}\hspace{1em}\emph{Category average} & 34.24 & $36.32 \pm 0.46$ & \textbf{36.75} & 36.14 & --- & --- & 39.34 & $\mathbf{40.52 \pm 0.36}$ & 40.17 & 39.89 & --- & --- \\
    \rowcolor{findingbg}\hspace{1em}\emph{Rel.\ improv.\ over Base (\%)} & --- & $+6.07$ & $\mathbf{+7.33}$ & $+5.55$ & --- & --- & --- & $\mathbf{+3.00}$ & $+2.11$ & $+1.40$ & --- & --- \\
    \midrule
    \rowcolor{gray!15}\multicolumn{13}{l}{\textit{Coding}} \\
    HumanEval+ & 70.27 & $\mathbf{74.90 \pm 0.92}$ & 73.48 & 72.64 & --- & --- & 75.94 & $78.57 \pm 0.52$ & \textbf{79.19} & 78.05 & --- & --- \\
    LiveCodeBench v1-5 & 20.68 & $\mathbf{22.35 \pm 0.58}$ & 21.82 & 22.33 & --- & --- & 25.23 & $28.01 \pm 0.52$ & 25.91 & \textbf{28.30} & --- & --- \\
    \rowcolor{findingbg}\hspace{1em}\emph{Category average} & 45.47 & $\mathbf{48.63 \pm 0.58}$ & 47.65 & 47.49 & --- & --- & 50.59 & $\mathbf{53.29 \pm 0.51}$ & 52.55 & 53.18 & --- & --- \\
    \rowcolor{findingbg}\hspace{1em}\emph{Rel.\ improv.\ over Base (\%)} & --- & $\mathbf{+6.95}$ & $+4.79$ & $+4.44$ & --- & --- & --- & $\mathbf{+5.34}$ & $+3.87$ & $+5.12$ & --- & --- \\
    \bottomrule
  \end{tabular}%
  }
\end{table}
}

\paragraph{Training configuration.}
For INFUSER, we use a document pool of size
$|\mathcal{D}_{\mathrm{doc}}|=12{,}260$ chunks collected from textbooks
in Astronomy, Biochemistry, Geography, and Physics.
The dev set
$\mathcal{D}_{\mathrm{dev}}$ contains 800 randomly sampled questions from the
SuperGPQA~\citep{superGPQA2025} science subset, comprising 3\% of the
full SuperGPQA set.
We chose SuperGPQA science because it spans diverse scientific subfields and has been carefully curated for question quality.  We use 800 questions by default; \Cref{fig:dev_size_sensitivity,tab:dev_size_sensitivity} report sensitivity over $|\mathcal{D}_{\mathrm{dev}}|\in\{100,200,400,800\}$.
We show in \S\ref{sec:comparison}
that little
leakage occurs through this dev set for INFUSER training,
so we still treat SuperGPQA as a valid evaluation benchmark.  We train INFUSER for
$T=100$ iterations on $8\times$ H100 GPUs with document batch size
$B=128$, group size $n=8$ for both generator and solver rollouts, AdamW
with weight decay $0.01$ and mini-batch size $32$.
We use solver learning rate $2\times 10^{-6}$ and generator learning rates $6\times 10^{-6}$ and $4\times 10^{-6}$ for Qwen3-4B-Base and Qwen3-8B-Base anchors, respectively.
More details are in \Cref{app:training_config}.

\vspace{1mm}
\paragraph{Evaluation protocol.}
We evaluate on general reasoning, math \& physics, and two out-of-domain transfer suites, medical and coding; benchmarks, prompts, and sampling scheme are in \S\ref{app:eval_protocol}.
Throughout, $\Delta$j denotes the performance gap between a method and the base model on the same benchmark.

\paragraph{INFUSER outperforms self-evolution baselines across domains and scales.}
We compare \textbf{INFUSER} with the base model and four contemporaneous self-evolution methods:
R-Zero~\citep{huang2025rzero}, AZR~\citep{zhao2025azr}, R-Few~\citep{yu2025rfew}, and SPICE~\citep{fang2025spice} on Qwen3-4B-Base and Qwen3-8B-Base as anchors. \Cref{tab:main_benchmarks} reports the per-benchmark and category-average accuracies for all methods. 
We observe three key trends:
(i) {\textbf{\textcolor{sectionblue}{INFUSER yields the largest gains on aligned domains.}}} INFUSER's gains are strongest on the general reasoning and Math \& physics benchmarks, with nearly \textbf{$20\%$} improvements for both anchors. These two fields are the most aligned with our document pool and dev set.
(ii) \textbf{\textcolor{sectionblue}{INFUSER unlocks cross-domain transfer.}} Despite the domain gap, INFUSER still improves over the base model on the medical and coding benchmarks, with gains comparable to or exceeding the best baselines. The out-of-domain transfer is consistent with previous findings on RLVR~\citep{wang2025oneshot,wen2025rlvr,liu2025prorl}. 
(iii) \textbf{\textcolor{sectionblue}{INFUSER's gains scale to larger models.}} INFUSER uniformly outperforms other methods on all four category averages at 8B. Its 4B-to-8B gain decay is minimal: math-and-physics holds ($+20.57\% \to +20.74\%$) and general-reasoning drops only $\sim 2.3$ points ($+20.26\% \to +17.98\%$). Baselines lose substantially more, e.g., R-Zero's math gain nearly halves ($+17.71\% \to +9.13\%$) and SPICE's general-reasoning gain drops $\sim 6.3$ points ($+18.81\% \to +12.55\%$).

\paragraph{Across-seed uncertainty.}
The INFUSER entries in \Cref{tab:main_benchmarks} report the mean and
sample standard deviation across three independently trained seeds.  At
the category level, the mean gain over Base exceeds one observed
cross-seed standard deviation for both anchors in every category.  The
largest dispersion is Qwen3-8B-Base math and physics ($2.80$ points),
which we analyze further in \S\ref{sec:rlvr_hybrid}; the corresponding
general, medical, and coding standard deviations are $0.97$, $0.36$, and
$0.51$ points.  These values describe run-to-run dispersion and are not
confidence intervals or significance tests.

\begin{figure}[!t]
  % \vspace{-2mm}
  \centering
  \captionsetup{font=small}
  \captionsetup[subfigure]{font=small}
  \hspace{-7mm}
  \begin{subfigure}[t]{0.73\linewidth}
    \centering
    \includegraphics[width=\linewidth]{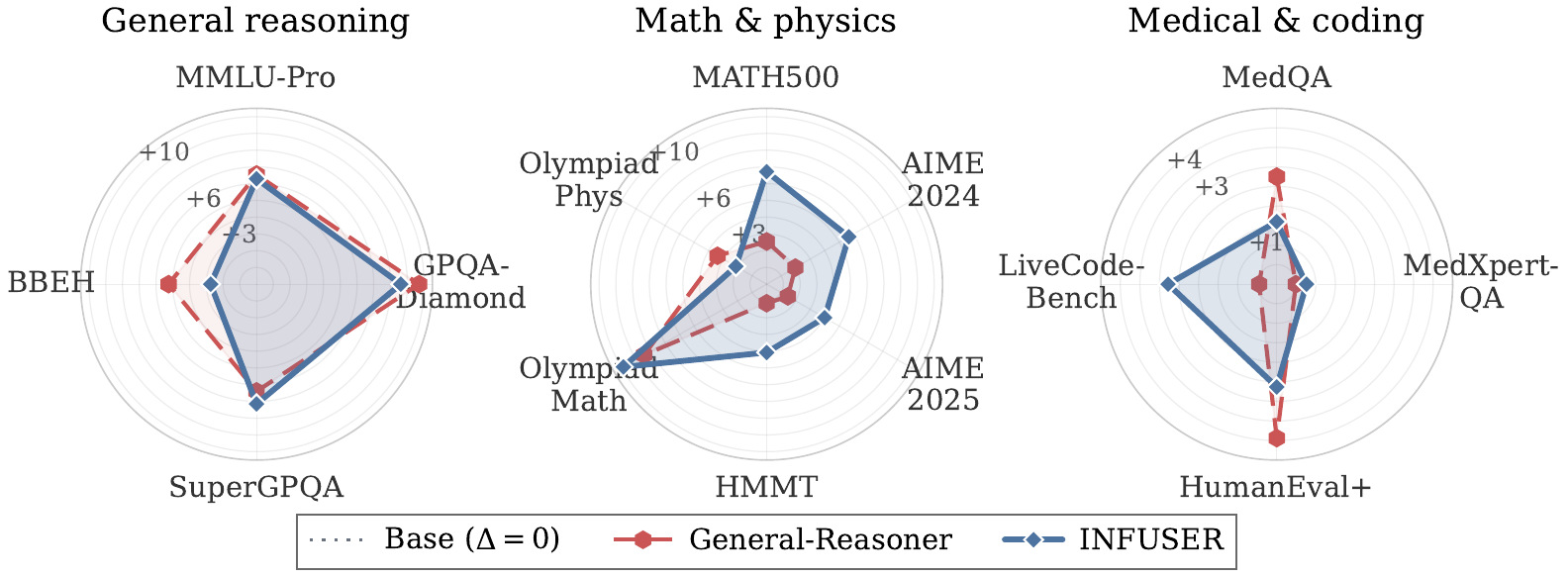}
    \caption{General-Reasoner comparison.}
    \label{fig:qw8bb_general_reasoner_radars}
  \end{subfigure}
  \begin{subfigure}[t]{0.30\linewidth}
    \centering
    \includegraphics[width=\linewidth]{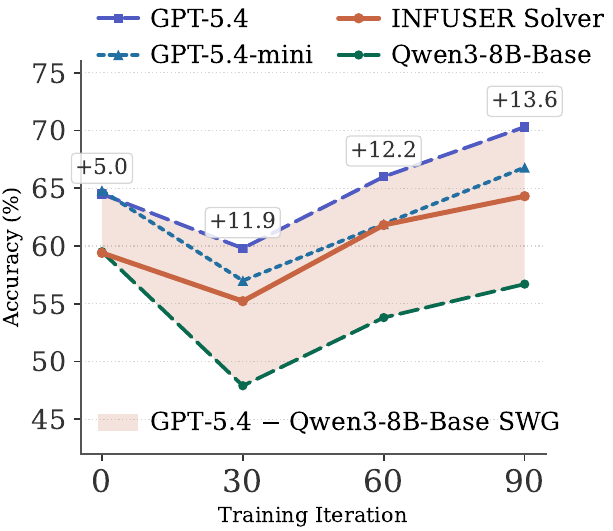}
    \caption{Generator quality.}
    \label{fig:gen_quality_qw8bb}
  \end{subfigure}
  \caption{\small
    \textbf{Left:} per-benchmark $\Delta$ over Qwen3-8B-Base for
    INFUSER and General-Reasoner.
    \textbf{Right:} solver accuracy on the questions produced by
    INFUSER's co-evolving generator in training; the annotation is the
    \emph{strong-against-weak gap} (SWG) between GPT-5.4 and
    Qwen3-8B-Base.}
  \label{fig:compact_benchmark_diagnostics}
  % \vspace{-5mm}
\end{figure}
\paragraph{Comparison with General-Reasoner.}
\Cref{fig:qw8bb_general_reasoner_radars} compares INFUSER with
General-Reasoner~\citep{ma2025generalreasoner} on Qwen3-8B-Base.
Note that General-Reasoner is \emph{not} a self-evolution method: it is standard RLVR over a fixed dataset of $230$K closed-model-curated questions (details in \S\ref{app:training_config}).
INFUSER instead co-evolves the generator with the solver from a $12$K-chunk science-textbook pool, with no closed-source teacher, verifier, or judge inside the iterative training loop. The document-pool construction stage uses an external browsing assistant only to locate open-access textbook sources, as detailed in \S\ref{app:document_pipeline}. Despite this much weaker data-side supervision,
INFUSER is stronger on $5$ of $6$ math benchmarks, remains competitive on the general-reasoning benchmarks, and is more balanced across the out-of-domain
benchmarks.

\begin{figure*}[!t]
  \centering
  \includegraphics[width=\textwidth]{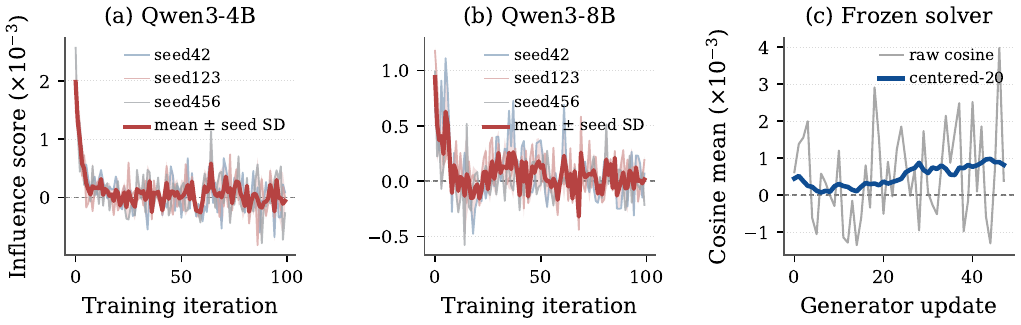}
  \caption{Influence-score dynamics.  \textbf{Left and center:} across
  three seeds, the score starts positive and remains bounded near zero as
  solver updates continually move the dev-set improvement direction.
  Lines show individual seeds, their mean, and the sample-standard-
  deviation band.  \textbf{Right:} with the solver and document batch
  fixed, the centered 20-update alignment increases overall, confirming
  that generator updates improve alignment against a stationary target.}
  \label{fig:influence_reward_diagnostics}
\end{figure*}

\paragraph{Influence-score dynamics.}
Perhaps surprisingly, the influence score does not continue to increase
under co-evolution.  In \Cref{fig:influence_reward_diagnostics}(a--b),
it starts positive, drops within roughly 20 iterations, and then remains
bounded near zero without growing oscillations.  Over iterations
20--99, the mean across per-seed temporal averages remains positive:
$(0.03\pm0.01)\times10^{-3}$ for Qwen3-4B-Base and
$(0.04\pm0.03)\times10^{-3}$ for Qwen3-8B-Base.  This positive mean is
important: generated questions continue to induce solver updates that
improve the dev objective on average.

Why does the score settle near zero despite remaining positive on
average?  We identify the moving solver as the cause.  In the control in
\Cref{fig:influence_reward_diagnostics}(c), we freeze the solver and one
document batch so that every generated question is evaluated against the
same dev-set improvement direction.  The centered alignment then rises
overall from $0.45\times10^{-3}$ to $0.82\times10^{-3}$, confirming that the generator learns to improve
influence when its target is stationary.  During full co-evolution, by
contrast, the solver immediately trains on the generated questions and
thereby changes what it still needs to learn.  Questions aligned with its
current weaknesses become less useful after those weaknesses are
updated, while the generator must adapt to the new ones.  This continual
chasing of a moving target produces the observed near-zero equilibrium;
it does not indicate failed reward optimization.  Sampled values are in
\Cref{tab:influence_diagnostics}.

\subsection{Generator Quality Analysis}
\label{sec:curriculum_analysis}

To assess whether the co-evolving generator produces a useful
curriculum, we evaluate the questions produced by
INFUSER generator (Qwen3-8B-Base anchor) at checkpoints $\{0,30,60,90\}$ with four solvers:
the Qwen3-8B-Base, the evolving INFUSER solver,
GPT-5.4-mini, and GPT-5.4.  
We track both \emph{per-solver accuracy} and the
\emph{strong-against-weak gap} (SWG), which is the accuracy gap between GPT-5.4 and Qwen3-8B-Base on the generator-produced questions, verified against the generator's own reference answers $a_\phi$.
A rising SWG indicates questions that grow harder for the base model yet remain well-posed (i.e., solvable by a strong solver), ruling out degenerate or ill-posed drift.

\paragraph{INFUSER produces a rising curriculum at the solver's learning frontier.}
\Cref{fig:gen_quality_qw8bb} reveals a two-phase dynamic.
From iteration 0 to 30, both solvers' accuracy drops yet the SWG more than doubles from $+5.0$ to $+11.9$, indicating that the added difficulty reflects genuine reasoning challenge rather than ill-posed questions.
From iteration 30 to 90, the generator transitions to a \emph{hardness-vs-quality trade-off}: both solvers' accuracy rises, but GPT-5.4 grows significantly faster, widening the SWG to $+13.6$.
The INFUSER solver stays $7$--$8$ points above the base model and tracks
GPT-5.4-mini from iteration 30 onward, indicating that INFUSER's co-evolving
solver can indeed learn from such a rising curriculum.

\paragraph{A qualitative example.}
We further provide an example in \Cref{fig:gen_quality_examples} to examine in detail
what improves in the generated questions and highlight two attributes: \emph{self-containedness} and \emph{factual correctness of the ground-truth key}. More details can be found in the related discussion in \Cref{app:gen_quality_examples}.

\begin{figure}[H]
  \centering
  % Fixed-width centered label slot so (A), (B), ... align at both parens.
  \newcommand{\mcqlabel}[1]{\makebox[0.8em][c]{#1}}
  % --- Source document card ---
  \begin{tcolorbox}[
    enhanced,
    colback=findingbg, colframe=findingbg,
    boxrule=0pt, arc=4pt,
    coltitle=findingframe, colbacktitle=findingbg,
    titlerule=0pt,
    title={\small\textbf{Source document}: lecture notes on membrane biophysics, ion permeability of lipid bilayers
    %  (doc id 3875)
     },
    fonttitle=\small,
    left=10pt, right=7pt, top=5pt, bottom=5pt,
  ]
  \small
  \vspace{0.2em}
  \centerline{\footnotesize\color{gray!60!black}%
    $\cdots$\;[${\approx}\,3700$ characters skipped]\;$\cdots$}
  \vspace{0.2em}

  A calculation of the image force gives the following result
  for the work necessary to move a charge from water to the
  middle of a membrane,
  %
  % NOTE: inline math with an `aligned` block is used in place of
  % \[ ... \] here on purpose.  `autonum` patches the `equation`
  % environment (which \[...\] expands to) and uses global \label
  % save/restore around it; that global restore escapes the figure
  % group and clobbers caption's local \label redefinition,
  % breaking every subsequent \label{sec:...} in the document
  % with an \caption@@xlabel / \caption@x@label
  % undefined-control-sequence error.  `aligned` is not in
  % autonum's patched-environment list, so it is safe.
  \begin{center}
  $\displaystyle
    \begin{aligned}
      \Delta G
      &= \frac{q^{2}}{2a}\!\left(\frac{1}{\varepsilon_{\mathrm{h}}}
          - \frac{1}{\varepsilon_{\mathrm{w}}}\right)  - \frac{q^{2}}{\varepsilon_{\mathrm{h}}\, l}
          \ln\!\left(\frac{2\varepsilon_{\mathrm{w}}}
                          {\varepsilon_{\mathrm{w}}+\varepsilon_{\mathrm{h}}}\right)
      \qquad (14.2),
    \end{aligned}
  $
  \end{center}
  where $l$ is the membrane thickness, $a$ the ionic radius,
  and $q$ the charge.  If we envision the flux as a
  barrier-crossing process, the rate is proportional to
  $J \propto e^{-\Delta G/\textit{KT}}$~(14.3).

  \vspace{0.2em}
  \centerline{\footnotesize\color{gray!60!black}%
    $\cdots$\;[${\approx}\,3000$ characters skipped]\;$\cdots$}
  \vspace{0.2em}
  \end{tcolorbox}

  \vspace{0.3em}

  % --- Two side-by-side question cards ---
  % Modern muted rose / teal accents (Tailwind rose-700 / emerald-700
  % for titles, rose-500 / emerald-500 for the left rule) instead of
  % saturated red/green mixed with black.  Cards use a left-accent
  % pattern: no colored title bar, soft tinted background, and a thin
  % colored rule on the left edge carrying the semantic color.
  \definecolor{cardBadAccent}{HTML}{F43F5E}%
  \definecolor{cardBadTitle}{HTML}{9F1239}%
  \definecolor{cardBadTint}{HTML}{FFF5F6}%
  \definecolor{cardGoodAccent}{HTML}{10B981}%
  \definecolor{cardGoodTitle}{HTML}{065F46}%
  \definecolor{cardGoodTint}{HTML}{F2FBF6}%
  \begin{minipage}[t]{0.40\textwidth}
    \begin{tcolorbox}[
      enhanced,
      equal height group=qpair,
      colback=cardBadTint, colframe=cardBadTint,
      boxrule=0pt, arc=4pt,
      borderline west={2.2pt}{0pt}{cardBadAccent},
      coltitle=cardBadTitle, colbacktitle=cardBadTint,
      titlerule=0pt,
      title={\small\textbf{Base model question} (ill-posed)},
      fonttitle=\small,
      left=10pt, right=7pt, top=5pt, bottom=5pt,
    ]
    \small
    \textbf{Q.}\ Given the free energy difference equation (14.2)
    for an ion moving from water to the interior of a membrane,
    which of the following factors would NOT increase the ion's
    flux across the membrane according to the barrier-crossing
    rate equation (14.3)?
    \begin{enumerate}[label=(\mcqlabel{\Alph*}), leftmargin=*, labelsep=0.3em, align=left, nosep]
      \item Increasing the ion's effective radius
      \item Decreasing the dielectric constant of the membrane
        interior
      \item Reducing the thickness of the membrane
      \item Lowering the temperature
    \end{enumerate}
    \textbf{Ground truth:}\ (A)~{\color{cardBadAccent}\ding{55}}.
    \end{tcolorbox}
  \end{minipage}\hfill
  \begin{minipage}[t]{0.59\textwidth}
    \begin{tcolorbox}[
      enhanced,
      equal height group=qpair,
      colback=cardGoodTint, colframe=cardGoodTint,
      boxrule=0pt, arc=4pt,
      borderline west={2.2pt}{0pt}{cardGoodAccent},
      coltitle=cardGoodTitle, colbacktitle=cardGoodTint,
      titlerule=0pt,
      title={\small\textbf{Checkpoint 90 question} (well-posed)},
      fonttitle=\small,
      left=10pt, right=7pt, top=5pt, bottom=5pt,
    ]
    \small
    \textbf{Q.}\ Given the free energy difference equation for
    moving an ion from water to the middle of a membrane,
    $\Delta G = \frac{q^{2}}{2a}\!\left(\frac{1}{\varepsilon_{\mathrm{h}}}
      - \frac{1}{\varepsilon_{\mathrm{w}}}\right)
      - \frac{q^{2}}{\varepsilon_{\mathrm{h}}\, l}
        \ln\!\left(\frac{2\varepsilon_{\mathrm{w}}}
                        {\varepsilon_{\mathrm{w}}+\varepsilon_{\mathrm{h}}}\right)$,
    which of the following correctly describes the impact of
    increasing the membrane thickness $l$ on the flux $J$ of an ion,
    assuming the flux is proportional to $e^{-\Delta G/\textit{KT}}$?
    \begin{enumerate}[label=(\mcqlabel{\Alph*}), leftmargin=*, labelsep=0.3em, align=left, nosep]
      \item Increasing $l$ decreases $\Delta G$ and thus increases $J$
      \item Increasing $l$ increases $\Delta G$ and thus decreases $J$
      \item Increasing $l$ has no effect on $\Delta G$ and thus no
        effect on $J$
      \item Increasing $l$ decreases $\Delta G$ but increases $J$
        only slightly
      \item Increasing $l$ increases $\Delta G$ but decreases $J$
        only slightly
      \item Increasing $l$ makes $\Delta G$ zero and thus $J$
        infinite
    \end{enumerate}
    \textbf{Ground truth:}\ (B)~{\color{cardGoodAccent}\ding{51}}.
    \end{tcolorbox}
  \end{minipage}
  \caption{Qualitative comparison of questions produced by
    INFUSER's co-evolving generator from the same source document at
    checkpoint~$0$ and checkpoint~$90$.}
  \label{fig:gen_quality_examples}
\end{figure}

\FloatBarrier

\subsection{Ablation Study}
\label{sec:comparison}
To better understand the role of each component in INFUSER, we compare
against alternative training strategies that represent natural design
choices.  Unless stated otherwise, all experiments in this section use
\textbf{Qwen3-8B-Base} as the anchor model, with
$\mathcal{D}_{\mathrm{dev}}$ set to an 800-question subset of
SuperGPQA Science.
For each method, we perform a single seeded training run and report the best checkpoint selected by the validation protocol of \S\ref{app:eval_protocol:ckpt_selection} (validation accuracy on a small validation set, checked every $5$ training iterations). The reported scores therefore differ slightly from those in \Cref{tab:main_benchmarks}, which are averaged over three seeds.

\subsubsection{Ablation on generator}
\begin{figure}[!t]
  % \vspace{-6mm}
  \centering
  \captionsetup{font=small}
  \captionsetup[subfigure]{font=small}
  \begin{subfigure}[t]{0.25\linewidth}
    \centering
    \includegraphics[width=\linewidth]{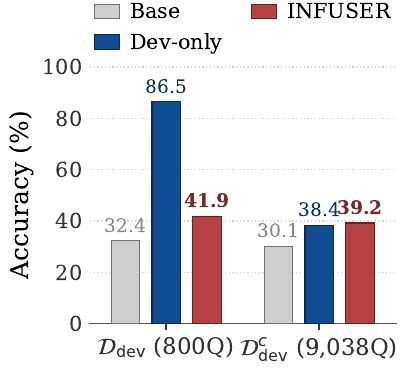}
    \caption{Dev-set leakage.}
    \label{fig:dev_leakage}
  \end{subfigure}\hfill
  \begin{subfigure}[t]{0.24\linewidth}
    \centering
    \includegraphics[width=\linewidth]{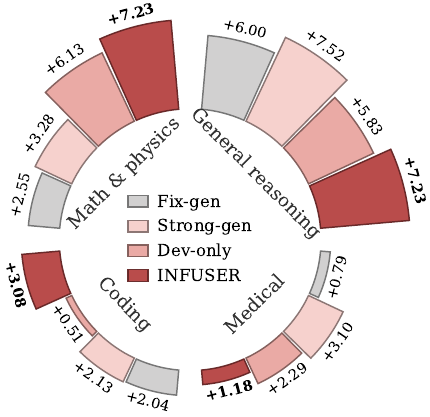}
    \caption{Generator ablation.}
    \label{fig:source_ablation}
  \end{subfigure}\hfill
  \begin{subfigure}[t]{0.24\linewidth}
    \centering
    \includegraphics[width=\linewidth]{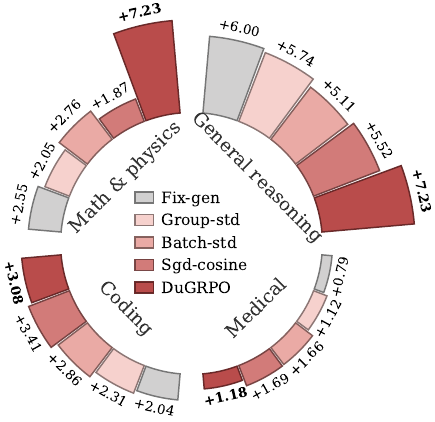}
    \caption{Update-rule ablation.}
    \label{fig:norm_ablation}
  \end{subfigure}\hfill
  \begin{subfigure}[t]{0.24\linewidth}
    \centering
    \includegraphics[width=\linewidth]{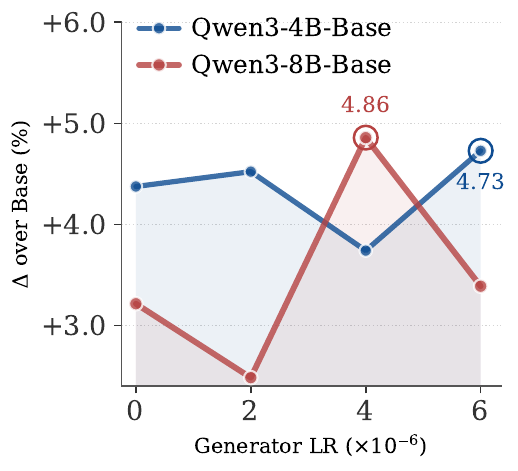}
    \caption{Generator-LR sweep.}
    \label{fig:glr_sweep}
  \end{subfigure}
  \caption{%
    Generator ablations on Qwen3-8B-Base.
    \textbf{(a)} Dev-only memorizes $\mathcal{D}_{\mathrm{dev}}$ while trailing INFUSER on the held-out complement $\mathcal{D}_{\mathrm{dev}}^{\mathrm{c}}$.
    \textbf{(b)} $\Delta$ over Base by benchmark category for four generator-source variants (Fix-gen, Strong-gen, Dev-only, INFUSER).
    \textbf{(c)} Two DuGRPO normalization variants and one influence-score variant without optimizer-awareness.
    \textbf{(d)} Mean accuracy over 14 benchmarks vs.\ generator learning rate on Qwen3-4B-Base and Qwen3-8B-Base.
    For each method, we report the best checkpoint chosen by the validation protocol of \S\ref{app:eval_protocol:ckpt_selection}.}
  \label{fig:compact_ablation_summary}
  % \vspace{-5mm}
\end{figure}

The generator is the component that turns $\mathcal{D}_{\mathrm{doc}}$
into a usable training curriculum. We ablate it along three axes: whether the generator is needed (\emph{Dev-only}), whether it must co-evolve with the solver (\emph{Fix-gen}), and whether it must be trained at all (\emph{Strong-gen}). The comparison is summarized in \Cref{fig:compact_ablation_summary}.  All four runs share
the same solver configuration (learning rate, batch size, number of
iterations) as the INFUSER anchor on Qwen3-8B-Base.

\paragraph{Direct dev-set training still generalizes, but trails INFUSER.}
To test whether the generator is needed at all, we introduce a
\textbf{Dev-only} baseline that drops the generator and trains the
solver directly on the 800-question $\mathcal{D}_{\mathrm{dev}}$.
This baseline is useful as a diagnostic, and it is not a pure failure:
\Cref{fig:source_ablation} shows that it improves over the base model
on all four category averages.  However, it is not a viable training
recipe: high-quality evaluation questions are scarce and expensive to
produce~\citep{rein2023gpqa,superGPQA2025}, and using them as solver
training data contaminates the signal used to measure progress.  To
diagnose whether such direct training generalises, we evaluate every
method on two splits drawn from the same
source distribution: $\mathcal{D}_{\mathrm{dev}}$ itself, and the
\emph{held-out complement} $\mathcal{D}_{\mathrm{dev}}^{\mathrm{c}}$,
defined as the $\compHoldoutSize$ remaining SuperGPQA Science questions after
$\mathcal{D}_{\mathrm{dev}}$ is removed.
\Cref{fig:dev_leakage} illustrates this.  On $\mathcal{D}_{\mathrm{dev}}$, Dev-only scores
$\compLeakDevOnlyDev\%$, far above INFUSER's $\compLeakInfDev\%$.  But on
$\mathcal{D}_{\mathrm{dev}}^{\mathrm{c}}$, Dev-only collapses to
$\compLeakDevOnlyHoldout\%$ while INFUSER holds at $\compLeakInfHoldout\%$.  In other words, Dev-only
mostly memorises the $800$-question training sample rather than
learning $\mathcal{P}$, whereas INFUSER's nearly identical scores on
the two splits indicate that the dev signal has been turned into a
generalising curriculum.  INFUSER also leads on the math, general
reasoning and coding category averages in \Cref{fig:source_ablation},
demonstrating the value of using a generator to turn the dev signal into a
renewable curriculum $\mathcal{Q}_\phi$.

\paragraph{INFUSER beats training with a larger frozen generator.}
A natural question is whether the generator needs to co-evolve with the
solver at all.  An initial document-conditioned generator
might already provide enough useful questions, or a much stronger frozen
generator might compensate for the lack of adaptation by producing higher-quality questions.  We therefore test two fixed-generator
baselines under the same solver updates as INFUSER: \textbf{Fix-gen}
keeps the same 8B generator frozen at its initial checkpoint, while
\textbf{Strong-gen} replaces it with a frozen Qwen3-32B thinking model.

At the same-size level, Fix-gen underperforms INFUSER on \emph{all} four categories, and the gap is especially large on math ($\compInfMathDelta$ vs. $\compFixMathDelta$).
Thus, a static same-size generator is not enough: the curriculum
must track the solver's changing learning frontier.  Scaling the frozen
generator helps. 
Strong-gen wins on
general reasoning and medical.  However, INFUSER, with only an 8B generator, still wins on math and coding, and is within $\compInfStrongReasonGap$ points of Strong-gen on general
reasoning.

\begin{finding}[Adaptation beats scale on reasoning, scale wins on knowledge]\label{find:influence_vs_scale}
  An 8B INFUSER generator outperforms a
  frozen 32B strong generator when used to train the solver on math and coding, and trails by only
  $\compInfStrongReasonGap$ points on general reasoning.  The
  frozen 32B generator's broader prior helps mainly on the more
  knowledge-heavy domains, namely general reasoning and medical.
\end{finding}

\paragraph{INFUSER benefits from both document knowledge and dev-set influence.}
The generator-source ablation in
\Cref{fig:source_ablation} further reveals that the dev set and the document
pool appear to play different roles.  On math, the single-source
\textbf{Dev-only} variant already delivers a large gain among the
baselines ($\compDevOnlyMathDelta$), substantially ahead of the two document-conditioned
runs (Fix-gen $\compFixMathDelta$, Strong-gen $\compStrongMathDelta$).  This pattern reverses on
general reasoning: both document-conditioned runs (Fix-gen $\compFixReasonDelta$,
Strong-gen $\compStrongReasonDelta$) outperform Dev-only ($\compDevOnlyReasonDelta$).  This suggests
that the dev-set signal directs the curriculum toward math-style
logical reasoning, while the document pool supplies broader source
material that transfers better to general reasoning.

\begin{finding}[INFUSER excels at combining knowledge and reasoning]\label{find:devset_doc_pool_complementary}
  The document-conditioned generator supplies diverse training
  content, while the influence signal from
  $\mathcal{D}_{\mathrm{dev}}$ directs that content toward the
  desired reasoning patterns.  The resulting joint gains are the
  largest we observe on math and remain competitive on general
  reasoning.
\end{finding}
Together, \Cref{find:devset_doc_pool_complementary} confirms the central INFUSER design: pair a document-conditioned generator with influence-guided supervision from $\mathcal{D}_{\mathrm{dev}}$, so that the curriculum becomes increasingly targeted and high-quality as training progresses.

\subsubsection{Optimizer-aware influence score and DuGRPO are essential}
The generator update rule combines two ingredients: the DuGRPO
advantage in \eqref{eq:gen_advantage}, which normalizes the raw
influence signal at both the within-group and batch levels, and the
optimizer-aware influence score itself
(defined in \S\ref{sec:method}, \eqref{eq:influence_score}).  We
ablate each ingredient in isolation while holding the
\textbf{Qwen3-8B-Base INFUSER configuration} and generator learning
rate ($4\times 10^{-6}$) fixed.  \texttt{group\_std} (the standard GRPO advantage
$A_{\mathrm{GRPO}}$) and \texttt{batch\_std} ($A_{\mathrm{BN}}$),
both defined in \eqref{eq:group_batch_std_advantages}, keep only the
within-group or only the batch term of the DuGRPO advantage,
respectively.  \texttt{sgd\_cosine} keeps
the full DuGRPO advantage but swaps the optimizer-aware influence
score for a plain SGD-style cosine, dropping the AdamW preconditioner
$\Gamma(q, a_\phi)$ in favor of the raw per-question gradient $g(q, a_\phi)$ from the
SGD decomposition in \eqref{eq:sgd_decomp}.  For
reference we also overlay the \emph{Fix-gen} baseline from the
previous section, which freezes the generator entirely and therefore
provides a lower bound for any generator-update rule in this
configuration.  The resulting comparison is shown in
\Cref{fig:norm_ablation}.

Replacing DuGRPO with either \texttt{group\_std} or
\texttt{batch\_std} reduces performance on math and general reasoning,
yielding performance close to the Fix-gen reference
(\Cref{fig:norm_ablation}).  Swapping the
optimizer-aware influence score for the plain \texttt{sgd\_cosine}
variant has the same effect: general reasoning and math regress toward Fix-gen, while coding and medical change by less than a point.  In effect, neither alternative normalization nor the
SGD-style influence score produces effective generator training,
so the solver trains against a curriculum that is indistinguishable
from a frozen generator.  This matches the motivation in
\eqref{eq:gen_advantage}: DuGRPO's combined group and batch scaling,
together with the optimizer-aware influence score, turns
noisy influence signals into generator updates that move beyond the
frozen-generator baseline.

\subsubsection{Generator learning rate requires per-anchor tuning}
To justify the anchor-specific generator learning rates used in
\S\ref{sec:exp:main_benchmarks}, we sweep the generator learning rate
over $\{0, 2\times 10^{-6}, 4\times 10^{-6}, 6\times 10^{-6}\}$ on
both Qwen3-4B-Base and Qwen3-8B-Base, where $0$ corresponds to the
Fix-gen baseline and nonzero points report the best checkpoint per
run selected by the validation protocol of
\S\ref{app:eval_protocol:ckpt_selection}.
As shown in \Cref{fig:glr_sweep}, the sweep peaks at
$6\times 10^{-6}$ on Qwen3-4B-Base and at $4\times 10^{-6}$ on
Qwen3-8B-Base, which are exactly the generator learning rates used
for each anchor in \S\ref{sec:exp:main_benchmarks}.  More broadly,
the anchor-dependent and non-monotone shape of the sweep underscores
that INFUSER is a dynamical two-player game: the generator
learning rate sets the tempo at which the generator adapts to the
solver, and the best operating point is a joint property of the two
players rather than a universal step-size choice.

\subsubsection{Sensitivity to dev-set size}
\label{sec:dev_size_sensitivity}
\begin{wrapfigure}{r}{0.45\linewidth}
  \centering
  \vspace{-8pt}
  \includegraphics[width=\linewidth]{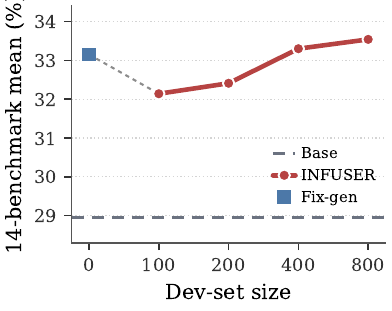}
  \caption{Qwen3-4B-Base dev-set-size sensitivity.  The point at zero is
  Fix-gen, which disables influence-guided generator updates; the dashed
  horizontal line is the untrained Base mean.}
  \label{fig:dev_size_sensitivity}
  \vspace{-8pt}
\end{wrapfigure}
We additionally vary the fixed SuperGPQA-Science dev set over
$\{100,200,400,800\}$ examples on Qwen3-4B-Base while holding the
remaining training and evaluation setup fixed.  For reference, the
zero point in \Cref{fig:dev_size_sensitivity} is Fix-gen, which keeps the
generator frozen.  Although all four INFUSER settings improve over the
base model, the 100- and 200-example settings score $32.14$ and $32.41$,
below Fix-gen at $33.15$.  Thus, limited dev data can become a bottleneck:
a noisy or unrepresentative estimate of the target gradient can bias the
generator update enough to underperform leaving the generator frozen.
Performance recovers with 400 examples ($33.30$) and is highest with the
default 800 ($33.54$), suggesting that influence training requires a
sufficiently representative target anchored to the target distribution.  Full results and the
per-domain breakdown are in
\Cref{tab:dev_size_sensitivity,fig:dev_size_sensitivity_domains}.

\subsection{Pass@\texorpdfstring{$k$}{k} study}
\begin{figure}[!t]
  \centering
  \includegraphics[width=\linewidth]{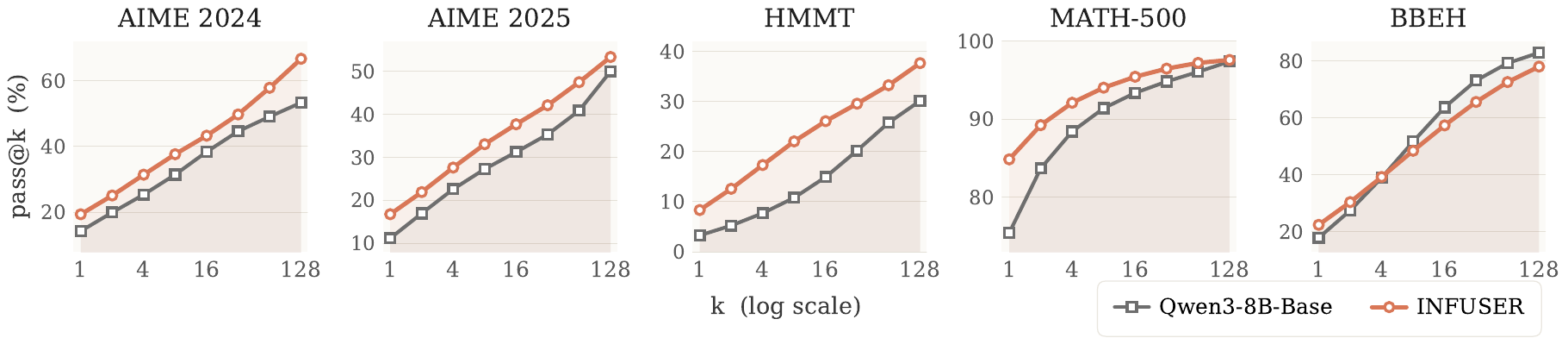}
  \caption{%
    pass@$k$ curves on four open-form math benchmarks plus BBEH for the
    Qwen3-8B-Base anchor, comparing the base model to
    \textbf{INFUSER}.  We exclude pure multiple-choice
    tasks because pass@$k$ on a small answer space is heavily inflated
    by random guessing.}
  \label{fig:qw8bb_passk_curves}
\end{figure}

Top-1 accuracy alone cannot distinguish whether a method merely sharpens
its best sample or improves the support of its sampled
reasoning distribution.  To make pass@$k$ meaningful, we therefore
restrict attention to hard math and general-reasoning benchmarks that
are not pure multiple choice.  On pure MCQ tasks, a model can
artificially improve pass@$k$ by sampling many guesses over a small
answer space, so the resulting curve reflects random-choice
coverage rather than reasoning diversity.  \Cref{fig:qw8bb_passk_curves}
therefore plots the Qwen3-8B-Base anchor on four open-form math
benchmarks plus BBEH, a general-reasoning benchmark whose answer space
is broad enough to suppress random-guessing effects.
INFUSER's curve stays above the base model for all
$k \le 128$ on AIME~2024, AIME~2025, HMMT, and MATH-500. By contrast, on the general-reasoning
benchmark BBEH, the two curves cross at $k=4$.  This reflects a
distinct reasoning pattern: on general reasoning, the model produces more consistent but less diverse outputs than on math.

\FloatBarrier

\section{Extension to Instruction-Finetuned Models}\label{sec:instruction_finetune}
Our main experiments all start from pretrained base models
(Qwen3-4B-Base and Qwen3-8B-Base).  A natural question is whether
INFUSER still yields gains when the anchor is already an
\emph{instruction-finetuned} (IF) model whose next-token distribution
has been reshaped by supervised finetuning.  The IF setting is also a
stricter stress test: INFUSER must adapt the model to a document-grounded
curriculum without destroying the formatting and instruction-following
habits learned during SFT.

We test this with \textbf{OLMo-3-7B-Instruct-SFT}~\citep{olmo3},
chosen because OLMo-3 releases its training recipe and instruction-tuning
mixture, which makes attribution and contamination auditing possible for
an IF anchor.  By contrast, instruction-tuned Qwen3 checkpoints have
undergone private post-training with large-scale SFT, RL, and/or teacher
distillation, making additional gains or regressions much harder to
attribute.  We compare the untrained IF \textbf{Base}, \textbf{Fix-gen}
(solver-only DrGRPO with a frozen generator), and \textbf{INFUSER} under
the same document pool, 800-question SuperGPQA Science development set,
and evaluation protocol as the main experiments.  Both trained runs use
solver learning rate $2\times 10^{-6}$; INFUSER uses generator learning
rate $4\times 10^{-6}$.  Full setup, checkpoint-selection, anchor-choice,
and contamination-audit details are deferred to
\Cref{app:instruction_finetune_details,app:dedup_check}.  The audit finds
no near-duplicate overlap between either $\mathcal{D}_{\mathrm{doc}}$ or
$\mathcal{D}_{\mathrm{dev}}$ and the released OLMo-3 SFT mixture under
the protocol of \S\ref{app:dedup_check}.

\begin{figure}[t]
  \centering
  % \vspace{-5mm}
  \captionsetup{font=small}
  \includegraphics[width=0.8\linewidth]{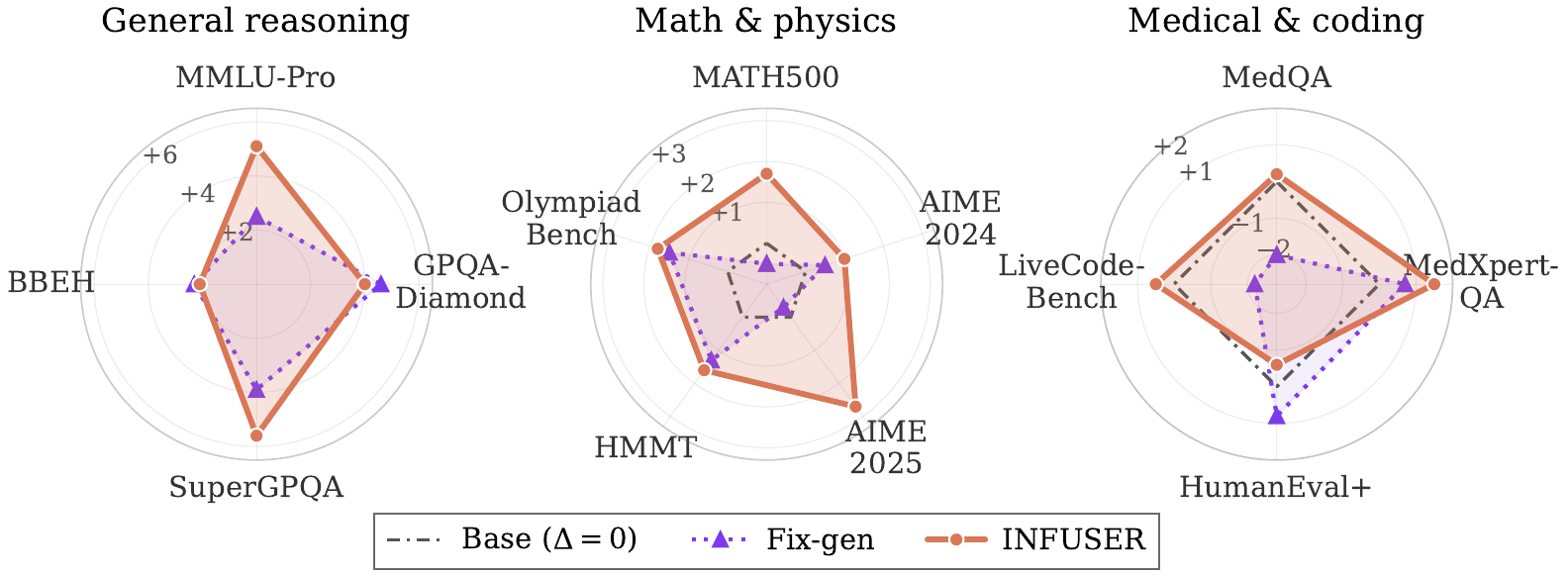}
  \caption{%
    Per-benchmark $\Delta$ over the OLMo-3-7B-Instruct-SFT base for
    Fix-gen and INFUSER, grouped into general reasoning (left),
    math \& physics (center), and medical \& coding (right).
    The dash-dot ring at $\Delta = 0$ marks the base IF checkpoint.
    The full per-benchmark accuracy table is in
    \Cref{tab:olmo_instruct_sft}.
  }
  \label{fig:olmo_sft_radars}
  % \vspace{-5mm}
\end{figure}

\paragraph{Results.}
\Cref{fig:olmo_sft_radars} shows the per-benchmark gain of Fix-gen and
INFUSER over the IF base.  INFUSER leads on $10$ of the $13$ benchmarks
and attains the highest overall average.  We see substantial gains on
MMLU-Pro ($\Delta=+5.1$) and SuperGPQA ($\Delta=+5.6$), and the same
alignment pattern from the pretrained anchors reappears: gains are
largest on general reasoning, the INFUSER polygon encloses Fix-gen on
all five math axes, and out-of-domain medical/coding transfer is
smaller.  Fix-gen also improves general reasoning, but it sits inside
INFUSER on math and dips below the base IF checkpoint on MedQA and
LiveCodeBench.  Thus, influence-guided generator updates continue to add
value beyond solver-only DrGRPO even after the anchor has already been
instruction-finetuned.

\FloatBarrier

\section{Extension: Augmenting Self-Evolution with RLVR}\label{sec:rlvr_hybrid}
Our main experiments drive self-evolution entirely from unlabeled
documents: every solver update is supervised by questions the generator
synthesizes from $\mathcal{D}_{\mathrm{doc}}$.  A complementary source of
solver signal is rule-verifiable RLVR, where the solver trains directly
on externally answered problems scored by a programmatic checker.  The two
signals are usually studied in isolation.  We ask whether a single
\textbf{INFUSER} loop can combine them, training one solver jointly on
document-grounded science self-evolution and verifiable math RLVR.  The
question is motivated by a concrete failure mode of the science-only
setting as follows.

\begin{figure}[!ht]
  \centering
  \begin{minipage}[t]{0.56\textwidth}
    \vspace{0pt}
    \centering
      \resizebox{\linewidth}{!}{%
  \begin{tabular}{lrrrr}
    \toprule
     & Math \& phys. & General & Medical & Coding \\
    \midrule
    \rowcolor{gray!15}\multicolumn{5}{l}{\textit{Science-only INFUSER}} \\
    seed 456 & 33.31 & 41.66 & 40.52 & 53.67 \\
    seed 123 & 28.26 & 40.47 & 40.88 & 53.51 \\
    seed 42 & 32.89 & 39.75 & 40.15 & 52.71 \\
    \rowcolor{findingbg}\hspace{1em}\emph{Average} & \textcolor[HTML]{1A6FAF}{$\mathbf{31.49 \pm 2.80}$} & $40.62 \pm 0.96$ & $40.52 \pm 0.37$ & $53.29 \pm 0.51$ \\
    \midrule
    \rowcolor{gray!15}\multicolumn{5}{l}{\textit{Math-RLVR \& INFUSER}} \\
    seed 456 & 33.07 & 39.82 & 38.89 & 52.00 \\
    seed 123 & 32.31 & 38.94 & 39.14 & 52.18 \\
    seed 42 & 32.18 & 39.36 & 40.15 & 53.28 \\
    \rowcolor{findingbg}\hspace{1em}\emph{Average} & \textcolor[HTML]{B64342}{$\mathbf{32.52 \pm 0.48}$} & $39.37 \pm 0.44$ & $39.39 \pm 0.67$ & $52.49 \pm 0.69$ \\
    \bottomrule
  \end{tabular}%
  }
  \end{minipage}\hfill%
  \begin{minipage}[t]{0.4\textwidth}
    \vspace{0pt}
    \centering
    \includegraphics[width=\linewidth]{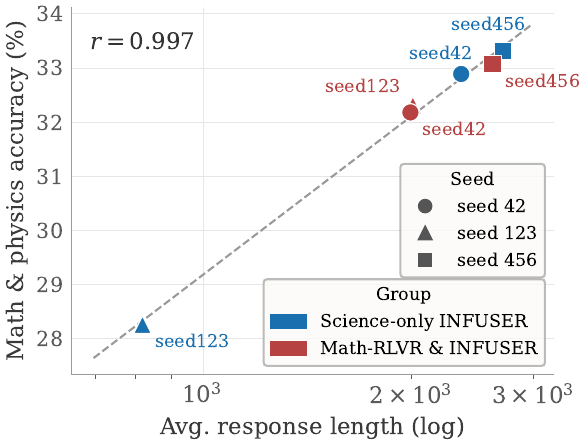}%
  \end{minipage}
  \caption{%
    Hybrid science+RLVR on Qwen3-8B-Base, three seeds per setting.
    \textbf{Left:} per-seed category accuracy (averaged over the same
    benchmark grouping as \Cref{tab:main_benchmarks}); \emph{Average} rows
    give mean\,$\pm$\,sample std, with \textcolor[HTML]{1a6faf}{\textbf{blue}} marking
    the unstable science-only scores on
    math.  \textbf{Right:} evaluation-time response length versus
    math-and-physics accuracy across all six seeds, with a log-linear fit
    ($r = 0.997$).  Together, adding verifiable math RLVR
    (\textcolor{red!70!black}{red}) reduces the Science-only
    (\textcolor[HTML]{1a6faf}{blue}) math variance by anchoring every seed
    to sufficient test-time compute.  Full setup in
    \S\ref{app:self_evolve_rlvr_hybrid}.
  }
  \label{fig:hybrid_overview}
\end{figure}

\paragraph{A seed instability in math coupled with response length collapse.}
In \Cref{fig:hybrid_overview}, left, we show both the average category accuracy and the cross-seed sample standard deviation of that accuracy across three seeds for the INFUSER Qwen3-8B-Base anchor (under Science-only INFUSER group). 
Across three seeds, science-only INFUSER is stable on
general reasoning, medical, and coding (cross-seed sample standard
deviations of $0.96$, $0.37$, and $0.51$ percentage points), but unstable
on math \& physics, where the cross-seed standard deviation is
$2.80$ percentage points.  Plotting each
checkpoint's evaluation-time response length against its math-and-physics
accuracy (\Cref{fig:hybrid_overview}, right), we find a strong log-linear
correlation ($r = 0.997$).  The solver's
learned evaluation-time response length, i.e., how much reasoning it
allocates per problem, almost entirely determines its math accuracy:
seeds that yield in longer responses score higher,
while the seed whose length collapses to ${\approx}800$ tokens scores lowest.  
Notably, the science-only INFUSER seed-123 running fails to incentivize the solver to sustain the thinking length that math problems require, and instead collapses to a suboptimal equilibrium with short responses and low math accuracy.
We attribute this to that science-only setting providing
little signal that anchors reasoning depth required by hard math problems.

\paragraph{RLVR-augmented training}
It is observed by previous RLVR work \citep{guo2025deepseek,team2025kimi,hu2025open} that RLVR involving math reasoning elicits a strong pattern on long CoT reasoning. 
Therefore, we naturally hypothesize that adding a verifiable math RLVR signal to the training loop will anchor reasoning depth across seeds, resolving the seed-dependent equilibrium ambiguity and stabilizing math performance.
We test it by adding a verified math component to both the
dev set that acts as influence anchor and the document pool that provides training curriculum, leaving the Qwen3-8B-Base
INFUSER recipe otherwise unchanged.  The $800$-question dev set
$\mathcal{D}_{\mathrm{dev}}$ is split evenly between SuperGPQA Science MCQs and AIME free-form problems (we use AIME data before 2024 to avoid benchmark leakage), so the influence direction carries both
science-MCQ and math signal.  The document pool augments the $12{,}260$
science document chunks with $10{,}000$ externally
Putnam/AIME-history rows where ground truth answers are already attached. These math rows supply directly verifiable
RLVR targets.  Curriculum construction, verifier routing, and the full
per-seed table are deferred to \S\ref{app:self_evolve_rlvr_hybrid}.

\begin{figure}[!ht]
  \centering
  \includegraphics[width=\textwidth]{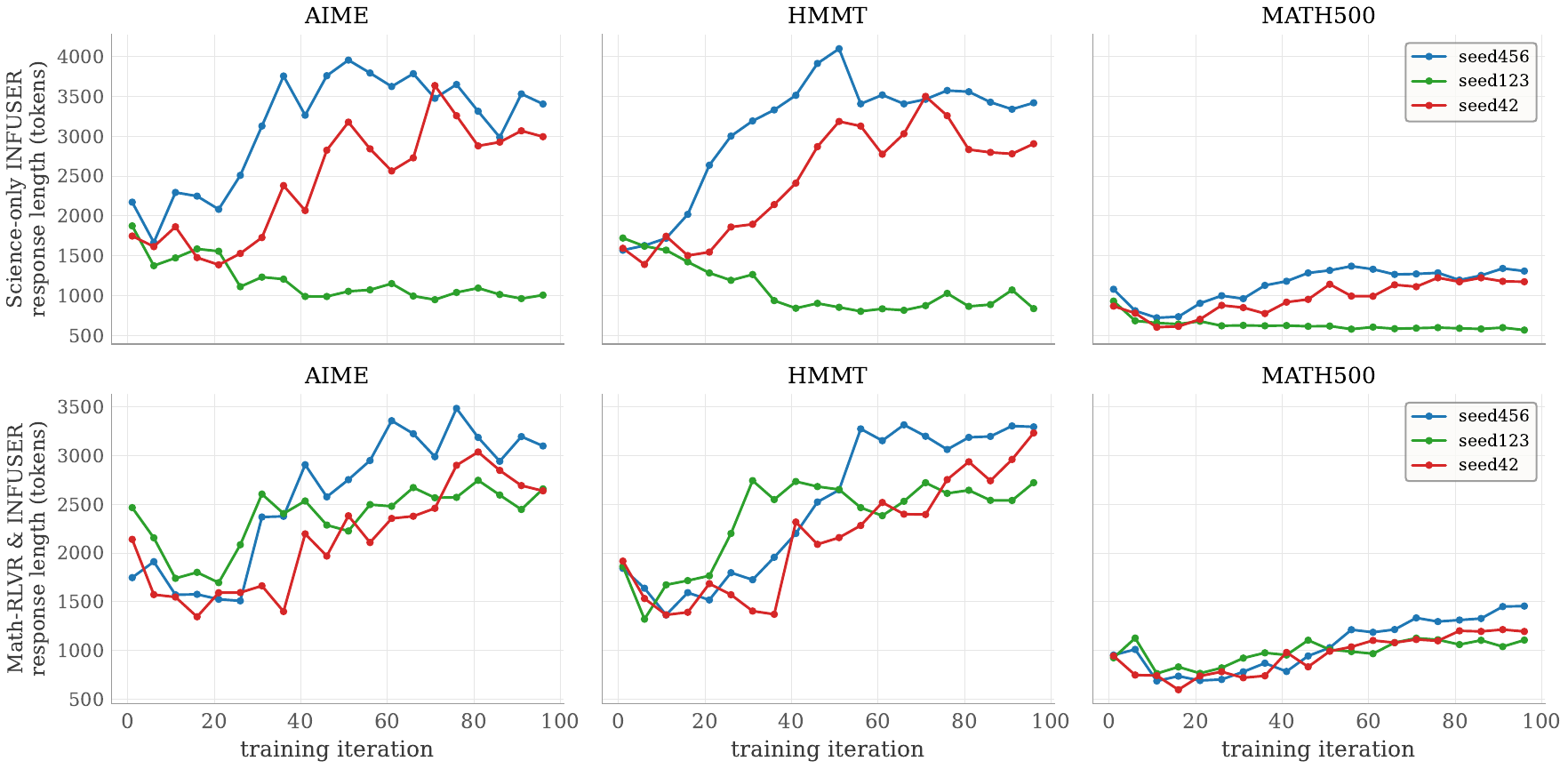}
  \caption{%
    Response length on AIME, HMMT, and MATH500 over the training course.
    Top row: three Science-only INFUSER seeds, which diverge onto different
    length regimes, the source of the cross-seed math accuracy variance.
    Bottom row: three Math-RLVR \& INFUSER seeds, which collapse onto a
    tightly clustered trajectory, confirming that verifiable math RLVR
    anchors reasoning depth across seeds.
  }
  \label{fig:hybrid_response_length_qw8bb}
\end{figure}

\paragraph{RLVR-augmented INFUSER improves math via long CoT.}
Adding verifiable math RLVR significantly reduces the seed variance it was
designed to target (\Cref{fig:hybrid_overview}, left).  The cross-seed
standard deviation of the math-and-physics average falls from $2.80$ to
$0.48$ percentage points, while the
math-and-physics mean rises from $31.49$ to $32.52$.  \Cref{fig:hybrid_response_length_qw8bb} exposes the
mechanism: the three science-only seeds (top row) diverge onto different
response-length regimes on AIME, HMMT, and MATH500, whereas the three
hybrid seeds (bottom row) collapse onto a single tightly clustered length
trajectory in training, and we see the reasoning length steadily increase over training, a good sign of the emergence of a long-CoT regime.
The alignment of length stabilization
with accuracy stabilization confirms that well-designed verifiable RLVR augmentation can work well with the rising curriculum.

The stabilization is not free (\Cref{fig:hybrid_overview}, left).
General reasoning, medical, and coding each dip slightly below the
science-only baseline ($40.63\to39.37$, $40.52\to39.39$, and
$53.30\to52.49$).  This is a budget-allocation effect: the science pool is $12{,}260$ of
$22{,}260$ training rows (${\approx}55\%$), so under a fixed solver-step
budget ($T=100$) the model sees only half as many science documents as the original INFUSER training,
weakening precisely the science-related fields uniformly.

\begin{finding}[Hybrid INFUSER improves math reasoning under a fixed budget]\label{find:hybrid_rlvr}
A single INFUSER loop can jointly run document-grounded science
self-evolution with math RLVR.  The RLVR component strengthens the long CoT behavior that is crucial for math reasoning, while we only see a small decline on other fields under the fixed training budget.
\end{finding}

\FloatBarrier

\section{Conclusion}\label{sec:conclusion}
We introduced \textbf{INFUSER}, a self-evolution framework that casts
generator--solver co-training as a cooperative bilevel game and rewards
the generator not by how \emph{hard} its questions are, but by how
\emph{useful} they are to the current solver. The key ingredient is an
optimizer-aware influence score \eqref{eq:influence_score} that, through
a first-order approximation of the bilevel objective, measures whether
training on a generated question moves the solver along a dev-anchored
target direction. Optimizing this score with \textbf{DuGRPO}, a
dual-normalized policy-gradient update tailored to the continuous and
noisy influence reward, turns an unstructured document pool into an
adaptive curriculum that tracks the solver's learning frontier rather
than a fixed notion of difficulty.

Empirically, INFUSER outperforms strong self-evolution baselines on
Qwen3-4B-Base and Qwen3-8B-Base, with the largest gains on the domains
most aligned with its document pool and dev set, positive transfer to
out-of-domain medical and coding benchmarks, and gains that persist as
the anchor scales from 4B to 8B where baselines decay. Our analyses
support the central design: an 8B co-evolving generator outperforms a
frozen 32B one on math and coding (\Cref{find:influence_vs_scale}), the
document pool and dev set play complementary
roles (\Cref{find:devset_doc_pool_complementary}), and both the
optimizer-aware influence score and DuGRPO are necessary for the
generator to move beyond a frozen-generator baseline. The framework
further extends to instruction-finetuned anchors (\S\ref{sec:instruction_finetune}) and
composes with verifiable RLVR to stabilize math reasoning depth (\S\ref{sec:rlvr_hybrid}).

\paragraph{Limitations and future work.}
First, INFUSER relies on a small dev set to define the target direction, so its
gains are strongest where this anchor and the document pool are well
aligned; extending the influence signal to steer co-evolution toward
out-of-domain targets remains open.
Second, although INFUSER produces a rising curriculum (\S\ref{sec:curriculum_analysis}), the
strong-solver evaluation and qualitative inspection (\Cref{fig:gen_quality_examples}) show that
the correctness and quality of the generated questions are still not close to perfect, which
leaves room for further improvement, e.g., by equipping the generator with a more sophisticated
agent loop with external tools for auditing the question quality and fixing errors.
Third, the hybrid results
expose a fixed-budget trade-off in which strengthening one domain
can slightly weaken others, suggesting adaptive allocation across signal
sources as a natural next step.

More broadly, influence-guided
self-evolution offers a path to convert abundant unstructured corpora
into structured training signal without a curated training set or teacher
model, and we believe coupling utility-based curriculum generation with
larger document pools and longer training horizons is a promising
direction for scaling reasoning.

\section*{Acknowledgement}
We want to thank Stanford Marlowe Cluster \citep{kapfer2025marlowe} for providing GPU resources. 
J.B. gratefully acknowledges supports from ONR under award N00014-24-1-2655, and the National Science Foundation (NSF) under grants 2312204 and 2403007.
T.W. acknowledges support from Google.

\newpage
\bibliographystyle{ims}
\bibliography{reference}

\newpage
\appendix

{\color{sectionblue} \tableofcontents}
\clearpage
\section{Setup and Notation}
\label{app:setup_notation}

\Cref{tab:notation_setup,tab:notation_algo} collect the notation used in
\S\ref{sec:method}. \Cref{tab:notation_setup} covers the problem setup:
models and data, the curriculum and rewards, and the population
objectives together with the one-step influence score.
\Cref{tab:notation_algo} covers the algorithmic side: the rollout-based
estimators, the RL update, and the training hyperparameters.

\begin{table}[H]
\centering
\caption{Notation for the problem setup and influence score
  (\S\ref{sec:method}).}
\label{tab:notation_setup}
\small
\renewcommand{\arraystretch}{1.25}
\begin{tabular}{@{}cp{\dimexpr\linewidth-5cm}@{}}
\toprule
\textbf{Symbol} & \textbf{Meaning} \\
\midrule
\multicolumn{2}{@{}l}{\textit{Models, distributions, and data}}\\
$\theta,\ \phi$   & Solver and generator model parameters \\
$\pi_\theta,\ \pi_\phi$
                  & Solver and generator policies parameterized by
                    $\theta$ and $\phi$ \\
$\mathcal{P}$     & Target distribution over verified question--answer
                    pairs $(q, a^*)$ \\
$\mathcal{D}_{\mathrm{dev}}$
                  & Development set: a finite sample
                    $\{(q_i, a_i^*)\}_{i=1}^m$ from $\mathcal{P}$, used
                    only to anchor the influence direction \\
$\mathcal{D}_{\mathrm{doc}}$
                  & Source document pool for question generation; not
                    itself the target distribution \\
$d$               & A document sampled from
                    $\mathcal{D}_{\mathrm{doc}}$ \\
\addlinespace[2pt]
\multicolumn{2}{@{}l}{\textit{Questions, answers, curriculum, and reward}}\\
$q,\ a^*$         & A question and its true gold answer (on
                    $\mathcal{P}$ or $\mathcal{D}_{\mathrm{dev}}$) \\
$(q, a_\phi)\sim\pi_\phi(\cdot\mid d)$
                  & Generated question $q$ with its
                    \emph{generated golden answer} $a_\phi$ from
                    document $d$ \\
$a_\phi$          & Generated golden answer; a proxy for $a^*$ inside
                    $\mathcal{Q}_\phi$, possibly noisy or wrong \\
$a\sim\pi_\theta(\cdot\mid q)$
                  & Solver answer sampled given $q$ \\
$r(a,b;q)\in\{0,1\}$
                  & Correctness reward for solver answer $a$ on $q$,
                    scored against reference $b$ ($b=a^*$ on
                    $\mathcal{P}/\mathcal{D}_{\mathrm{dev}}$, $b=a_\phi$
                    on $\mathcal{Q}_\phi$) \\
$\mathcal{Q}_\phi=\{(q_j,a_{\phi,j})\}$
                  & Curriculum of question--answer pairs induced by
                    $\pi_\phi$ \\
\addlinespace[2pt]
\multicolumn{2}{@{}l}{\textit{Objectives, solver update, and influence score}}\\
$J(\theta)$       & Solver's target performance,
                    $\mathbb{E}_{(q,a^*)\sim\mathcal{P}}
                    \mathbb{E}_{a\sim\pi_\theta(\cdot\mid q)}
                    [r(a,a^*;q)]$ \\
$J(\theta;\mathcal{Q}_\phi)$
                  & Solver objective on the curriculum (reference
                    $a_\phi$); $J(\theta;q,a_\phi)$ is its
                    single-question version, see
                    \eqref{eq:single_question_solver_gradient} \\
$\hat{J}(\theta)$ & Empirical estimate of $J(\theta)$ on
                    $\mathcal{D}_{\mathrm{dev}}$, \eqref{eq:dev_surrogate} \\
$\theta^*(\phi)$  & Ideal best-response solver,
                    $\argmax_\theta J(\theta;\mathcal{Q}_\phi)$ \\
$\theta^+(\phi)$  & One-step solver-update map,
                    $\theta+\Delta\theta(\phi)$, \eqref{eq:solver_step} \\
$\Delta\theta(\phi),\ \mathrm{Opt}(\cdot)$
                  & One-step parameter update
                    $\mathrm{Opt}(\nabla_\theta J(\theta;\mathcal{Q}_\phi))$
                    and the optimizer update rule (AdamW) \\
$g(q,a_\phi)$     & Per-question solver policy gradient
                    $\nabla_\theta J(\theta;q,a_\phi)$ \\
$\Gamma(q,a_\phi)$
                  & AdamW-preconditioned per-question update direction
                    induced by $g(q,a_\phi)$
                    (\S\ref{app:adamw_preconditioning}) \\
$\mathrm{cossim}(u,v)$
                  & Cosine similarity
                    $\langle u,v\rangle/(\|u\|\,\|v\|)$ \\
$s(q,a_\phi)$     & Optimizer-aware influence score
                    $\mathrm{cossim}(\nabla_\theta J(\theta),
                    \Gamma(q,a_\phi))$, \eqref{eq:influence_score} \\
\bottomrule
\end{tabular}
\end{table}

\begin{table}[H]
\centering
\caption{Notation for INFUSER's rollout estimators, RL update, and
  hyperparameters (\S\ref{sec:method}).}
\label{tab:notation_algo}
\small
\renewcommand{\arraystretch}{1.25}
\begin{tabular}{@{}cp{\dimexpr\linewidth-5cm}@{}}
\toprule
\textbf{Symbol} & \textbf{Meaning} \\
\midrule
\multicolumn{2}{@{}l}{\textit{Rollout estimators and RL update}}\\
$\mathcal{J}_\psi(z)$
                  & Clipped rollout objective for policy $\pi_\psi$ on
                    input $z$, \eqref{eq:unified_obj}; $\psi\in\{\theta,\phi\}$ \\
$\pi_{\psi_{\mathrm{old}}}$
                  & Rollout (behavior) policy that generated a batch \\
$\rho_{i,t}$      & Token-level importance ratio
                    $\pi_\psi/\pi_{\psi_{\mathrm{old}}}$ in
                    \eqref{eq:unified_obj} \\
$\hat A_i$        & Per-row advantage in \eqref{eq:unified_obj};
                    $\hat A_{\mathrm{sol}}$ is the mean-centered solver
                    advantage \\
$\hat g_{\mathrm{dev}}$
                  & Rollout estimate of $\nabla_\theta\hat J(\theta)$,
                    the dev reference direction, \eqref{eq:hat_g_dev} \\
$\hat g(q,a_\phi),\ \hat\Gamma(q,a_\phi)$
                  & Finite-rollout estimates of $g$ and $\Gamma$ \\
$\hat s(q,a_\phi)$
                  & Empirical influence reward
                    $\mathrm{cossim}(\hat g_{\mathrm{dev}},
                    \hat\Gamma(q,a_\phi))$, \eqref{eq:estimated_influence_score} \\
$\mu_d,\ \sigma_d$
                  & Mean and std of $\{s(q^k,a_{\phi}^k)\}_{k=1}^n$ for
                    document $d$ \\
$\mathcal{B},\ \sigma_{\mathcal{B}}$
                  & Document batch and its cross-group normalizer
                    $\mathrm{mean}\{\sigma_{d'}:d'\in\mathcal{B}\}$ \\
$\hat A_{\mathrm{gen}}$
                  & DuGRPO generator advantage, \eqref{eq:gen_advantage} \\
$\hat A_{\mathrm{GRPO}},\ \hat A_{\mathrm{BN}}$
                  & Group-only and batch-only ablation normalizers,
                    \eqref{eq:group_batch_std_advantages} \\
\addlinespace[2pt]
\multicolumn{2}{@{}l}{\textit{Hyperparameters and constants}}\\
$n$               & Group size for solver and generator rollouts
                    ($n=8$) \\
$B$               & Document batch size ($B=128$) \\
$M$               & Minibatch size for optimizer steps ($M=32$) \\
$T$               & Number of training iterations (answer loops) \\
$\eta_{\mathrm{sol}},\ \eta_{\mathrm{gen}}$
                  & Solver and generator learning rates \\
$\epsilon$        & PPO-style clipping coefficient in
                    \eqref{eq:unified_obj} \\
$C$               & Fixed maximum generation length (Dr.GRPO length
                    normalizer) \\
$\rho_{\max}$     & Truncated importance-sampling clip ($\rho_{\max}=2.0$) \\
\bottomrule
\end{tabular}
\end{table}

\paragraph{Remark (shared initialization).}
The generator and solver maintain \emph{separate} model weights
$\phi$ and $\theta$, each with its own optimizer state.  Both are
initialized from the same pretrained checkpoint but diverge during
training.

\section{Related Works}
\label{sec:related_work}

\textbf{Reinforcement Learning for LLM Reasoning.} Reinforcement learning has surged as a main-stream post-training method. DeepSeek-R1-Zero first showed that rule-based RL can elicit long-chain reasoning, self-reflection, and verification from a base model, while also exposing readability and language-mixing issues that motivated the later cold-start pipeline of DeepSeek-R1~\citep{guo2025deepseek}. This line builds on DeepSeekMath, which introduced Group Relative Policy Optimization (GRPO) for efficient mathematical RL~\citep{shao2024deepseekmath}, and is related to Kimi~k1.5, which scales long-context RL for strong long-CoT reasoning~\citep{team2025kimi}. Subsequent works study the robustness and scalability of zero-style RLVR: SimpleRL-Zoo shows that its success depends on base-model capability, reward design, query difficulty, and training dynamics~\citep{zeng2025simplerl}; Open-Reasoner-Zero reproduces R1-Zero-like length and performance scaling with an open PPO/GAE recipe~\citep{hu2025open}; Logic-RL validates rule-based RL on logical reasoning~\citep{xie2025logic}; and small-model or data-limited studies show that RLVR can still yield reasoning gains under constrained model size, data, or compute~\citep{dang2025reinforcement,wang2025oneshot}. Beyond math, General-Reasoner extends RLVR to broad domains using large-scale verifiable data and generative verification~\citep{ma2025generalreasoner}, Search-R1 incorporates retrieval-augmented reasoning~\citep{jin2025search}, and RAGEN studies multi-turn agentic RL with new stability and reward-shaping challenges~\citep{wang2025ragen}. In parallel, algorithmic refinements improve long-CoT RL training: DAPO introduces decoupled clipping and dynamic sampling~\citep{yu2025dapo}; Dr.~GRPO identifies and corrects length-related GRPO bias~\citep{liu2025understanding}; VAPO develops value-based augmented PPO~\citep{yue2025vapo}; PRIME uses implicit process rewards~\citep{cui2025process}; and open systems such as DeepScaleR and Skywork-OR1 further study RL recipes, entropy control, and length scaling for compact reasoning models~\citep{luo2025deepscaler,he2025skywork}. Finally, mechanism studies debate whether RLVR genuinely expands reasoning capacity or mainly reallocates probability mass over reasoning paths already present in the base model~\citep{yue2025does,wen2025rlvr}. Together, these works establish zero-style RLVR as a promising paradigm for eliciting reasoning from base models, while leaving open questions about base-model prerequisites, verifier design, exploration, reward sparsity, length bias, and the source of reasoning improvement. In contrast, INFUSER does not rely on a curated pool of human-authored or frontier-model-filtered verifiable training problems; instead, it produces document-grounded training signals from unlabeled documents using the model itself, with only a small held-out target sample $\mathcal{D}_{\mathrm{dev}}$ as external supervision to anchor the generator's reward.

\textbf{Self-play and adaptive curriculum generation} improve LLMs with RL by generating, filtering, or scheduling training problems according to the model's evolving capability rather than using a fixed human-curated dataset. Early self-improvement methods such as STaR iteratively generate and filter rationales for fine-tuning~\citep{zelikman2022star}, while Quiet-STaR extends latent rationale generation to arbitrary text~\citep{zelikman2024quiet}. For alignment, SPIN improves a model by contrasting its own responses with human demonstrations~\citep{chen2024self}, SPPO casts preference optimization as a self-play game~\citep{wu2024self}, and self-rewarding models use the model itself as a judge to generate rewards for iterative improvement~\citep{yuan2024self}. More recent reasoning-oriented methods apply self-play directly to RL: R-Zero co-evolves a Challenger and Solver to generate tasks near the Solver's capability boundary~\citep{huang2025rzero}; Absolute Zero removes external data by proposing and solving verifiable code-reasoning tasks with executor-based validation~\citep{zhao2025azr}; R-Few uses a few human examples to guide self-evolution and stabilize the curriculum~\citep{yu2025rfew}; SPICE mines corpus environments to construct document-grounded reasoning tasks~\citep{fang2025spice}; and SGS scales conjecturer--prover self-play in Lean4 by adding a model-as-guide role that scores generated problems for target-relevance and naturalness to mitigate conjecturer reward hacking over long training horizons~\citep{bailey2026scaling}. Related self-play frameworks study transferable reasoning through zero-sum games~\citep{liu2025spiral} and online attacker--defender training for safety~\citep{liu2025chasing}. In parallel, adaptive curriculum methods make RL post-training more sample-efficient by constructing reverse curricula from correct demonstrations~\citep{xi2024training}, automatically adjusting expert-iteration rewards~\citep{zhao2024automatic}, scheduling problem distributions with learnability and exploration criteria~\citep{wang2025dump}, formulating curriculum selection as a non-stationary bandit~\citep{chen2025self}, progressing from easy to hard tasks~\citep{parashar2025curriculum}, or adapting difficulty and hints to model capability~\citep{wu2025progressive}. Recent analyses further compare self-play with standard RLVR and SFT through update sparsity, entropy dynamics, and proposer reward design~\citep{chae2025towards}. The works most closely related to our setting are STP~\citep{dong2025stp} and SPICE~\citep{fang2025spice}, both of which generate questions adversarially. For example, in STP, a conjecturer is trained to generate conjectures that are barely provable by the current prover, thereby inducing an adaptive, self-generated curriculum. In contrast, we formulate generator learning as a cooperative bilevel curriculum game, and approximate the generator's outer-loop update using an influence-based first-order signal derived from held-out performance.

\textbf{Influence-guided training data selection and synthesis.} A separate line of work measures the utility of each training example by how much it improves a downstream objective. Influence functions~\citep{koh2017understanding} formalize this leave-one-out perturbation analysis, and gradient-alignment surrogates make it tractable at scale: LESS~\citep{xia2024less} approximates the AdamW-induced influence with low-rank gradient features and uses the resulting score as an offline filter over an existing instruction-tuning pool, and CROPI extends this selection paradigm to RLVR with an off-policy influence estimator built from pre-collected trajectories and sparse random projections, used to drive a multi-stage curriculum~\citep{zhu2025cropi}. GradAlign~\citep{yang2026gradalign} dynamically selects RL training problems using cosine similarity between each problem's policy gradient and the average policy gradient on a small, trusted validation set, periodically resampling on-policy rollouts and recomputing the gradients as the policy evolves. This shares INFUSER's use of held-out gradient alignment to estimate training utility: GradAlign uses this signal to select existing problems, while INFUSER uses an optimizer-aware score as a reward for a question generator that co-evolves with the solver. Recent works extend this signal from selection to generation. Montessori-Instruct~\citep{li2024montessori} measures the local data influence of synthesized instructions on a student model and trains a frozen teacher via DPO to favor high-influence outputs for instruction tuning. Concurrent work OptimSyn~\citep{fan2026optimsyn} couples an optimizer-aware influence score with a GRPO-trained rubric generator that synthesizes QA pairs conditioned on a seed document, closing the synthesis--training loop on a frozen target model. INFUSER shares the optimizer-aware per-question influence score with this line and shares with OptimSyn in particular the use of an RL-trained generator. We differ from these synthesis methods in that we (i) jointly co-evolve the solver and the generator from the same pretrained model, rather than improving training data for a frozen target student; (ii) operate directly on unlabeled documents with binary correctness rewards against the generator's reference answer, without rubric mediation or instruction-tuning supervision; and (iii) introduce DuGRPO to handle the variance of the continuous influence reward in a multi-document, multi-question batch.

\textbf{Meta-Learning} studies how knowledge accumulated across a distribution of tasks can improve adaptation to new tasks~\citep{vilalta2002perspective,vanschoren2018meta,wang2018dataset,hospedales2021meta,huisman2021survey}. Existing methods are typically grouped into four families: model-based, optimization-based, metric-based, and data-based approaches~\citep{vanschoren2018meta,hospedales2021meta,huisman2021survey}. Model-based methods encode adaptation directly into the architecture through recurrent dynamics, external memory, or fast weights. Representative examples include memory-augmented neural networks~\citep{santoro2016meta}, Meta Networks~\citep{munkhdalai2017meta}, recurrent learned optimizers~\citep{andrychowicz2016learning,ravi2017optimization}, and attention-based architectures such as SNAIL~\citep{mishra2017simple}. Optimization-based methods instead learn parameters that can be adapted to a new task with only a few gradient steps~\citep{finn2017model}. A canonical example is MAML, which learns an initialization optimized for rapid post-adaptation generalization~\citep{finn2017model}. Later variants improve this paradigm through learned update directions~\citep{li2017meta}, low-dimensional adaptation~\citep{zintgraf2019fast}, and latent-space adaptation~\citep{rusu2018meta}. Metric-based methods learn an embedding space or similarity rule for few-shot prediction by comparing query examples with a small support set~\citep{vinyals2016matching,snell2017prototypical,sung2018learning}. Representative methods include Matching Networks~\citep{vinyals2016matching}, Prototypical Networks~\citep{snell2017prototypical}, and Relation Networks~\citep{sung2018learning}. Later extensions improve metric flexibility through task conditioning~\citep{oreshkin2018tadam} or by combining transfer learning with episodic adaptation~\citep{sun2019meta}. Data-based meta-learning meta-learns a small synthetic training set, rather than an initialization, optimizer, or metric, such that training on the synthetic data approximates training on the full dataset~\citep{wang2018dataset,zhao2020dataset,yu2023dataset}. This line of work, often known as dataset distillation or dataset condensation, was initiated by \citet{wang2018dataset} and later improved through gradient matching~\citep{zhao2020dataset}. Different from the previous three families, these methods meta-learn the training data itself and thus form a separate data-based paradigm~\citep{yu2023dataset}. In contrast, our focus is on enabling the generator to adapt the curriculum to the solver's needs as the solver continuously improves, which can be viewed as a form of meta-learning with an evolving target task.

\textbf{Multi-Agent RL for Language Models.} Our work is also related to multi-agent reinforcement learning (MARL) for language models.
Full-scale LLM training in MARL environments faces nontrivial challenges~\citep{wan2025rema,liu2025chasing}.
Existing language-model and multi-agent works address these challenges by using lighter models~\citep{sarkar2025training}, simplifying communication-game environments~\citep{graesser2019emergent}, or studying self-play in text-game negotiation settings~\citep{liao2024efficacy}.
In contrast, INFUSER jointly updates the solver and generator with carefully designed learning paces, which stabilize their interaction and improve performance.

\section{Omitted Details}
\label{app:omitted_details}

\subsection{Full Algorithm}
\label{app:full_algorithm}

\SetKwBlock{PhaseDev}{Phase 1.\ Solver-side Dev Reference}{}
\SetKwBlock{PhaseBatch}{Phase 2.\ Batch rollout and parsing}{}
\SetKwBlock{PhaseInfluence}{Phase 3.\ Influence Estimation}{}
\SetKwBlock{PhaseGenUpdate}{Phase 4.\ Generator Update}{}
\SetKwBlock{PhaseSolUpdate}{Phase 5.\ Solver Update}{}

\begin{algorithm}[!htbp]
\small
\caption{Full INFUSER: Influence-Guided Self-Evolution}
\label{alg:infuser_full}
\DontPrintSemicolon
\KwIn{Pretrained LLM (init.\ for $\pi_\theta$, $\pi_\phi$); doc pool $\mathcal{D}_{\mathrm{doc}}$; dev set $\mathcal{D}_{\mathrm{dev}}$; doc batch size $B$; group size $n$; answer-loop count $T$; minibatch size $M$; learning rates $\eta_{\mathrm{sol}}, \eta_{\mathrm{gen}}$; invalid question penalty $\rho_{\mathrm{inv}}$; AdamW hyperparams.}
\KwOut{Trained solver $\pi_\theta$ and trained generator $\pi_\phi$.}
\For{$t \leftarrow 1$ \KwTo $T$}{
  \PhaseDev{
    \textcolor{findingaccent!85!black}{$\triangleright$\,\emph{Solver rollout on $\mathcal{D}_{\mathrm{dev}}$}}\;
    \ForEach{$(\tilde q, \tilde a^*)\in\cD_{\mathrm{dev}}$}{
      Sample $\{\tilde a^i\}_{i=1}^n \sim \pi_\theta(\cdot \mid \tilde q)$ and compute rewards $\{r(\tilde a^i, \tilde a^*; \tilde q)\}_{i=1}^n$\;
      $\hat{A}_{\mathrm{sol}}(\tilde q, \tilde a^i) \leftarrow r(\tilde a^i, \tilde a^*; \tilde q) - \mathrm{mean}\{r(\tilde a^j, \tilde a^*; \tilde q)\}_{j=1}^n$ for $i = 1, \dots, n$\;
    }
    \textcolor{findingaccent!85!black}{$\triangleright$\,\emph{Solver-side reference gradient on $\mathcal{D}_{\mathrm{dev}}$}}\;
    Compute $\hat{g}_{\mathrm{dev}} \leftarrow |\cD_{\mathrm{dev}}|^{-1}\sum_{(\tilde q,\tilde a^*)\in\cD_{\mathrm{dev}}}\nabla_\theta \cJ(\theta;\,\tilde q,\tilde a^*)$ using $\hat{A}_{\mathrm{sol}}$\;
  }
  \PhaseBatch{
    Sample document batch $\cB_{\mathrm{gen}}=\{d_b\}_{b=1}^B \sim \mathcal{D}_{\mathrm{doc}}$\;
    \textcolor{findingaccent!85!black}{$\triangleright$\,\emph{Generator rollout on document batch}}\;
    \ForEach{$d\in\cB_{\mathrm{gen}}$}{
      Generate $\{(q^i, a_{\phi}^i)\}_{i=1}^n \sim \pi_\phi(\cdot \mid d)$\;
    }
    Parse generated outputs into valid question--answer pairs $\cQ_\phi$ and malformed generations $\cI_\phi$\;
    \textcolor{findingaccent!85!black}{$\triangleright$\,\emph{Solver rollout on generated questions}}\;
    \ForEach{$(q, a_\phi)\in \cQ_\phi$}{
      Sample $\{a^i\}_{i=1}^n \sim \pi_\theta(\cdot \mid q)$ and compute rewards $\{r(a^i, a_\phi; q)\}_{i=1}^n$\;
      $\hat{A}_{\mathrm{sol}}(q, a^i) \leftarrow r(a^i, a_\phi; q) - \mathrm{mean}\{r(a^j, a_\phi; q)\}_{j=1}^n$ for $i = 1, \dots, n$\;
      \textcolor{findingaccent!85!black}{$\triangleright$\,\emph{Per question solver gradient}}\;
      Compute $\hat g(q, a_\phi) \leftarrow \nabla_\theta \cJ_\theta(q, a_\phi)$ using $\hat{A}_{\mathrm{sol}}$\;
    }
  }
  \PhaseInfluence{
    \ForEach{$(q, a_\phi)\in\cQ_\phi$}{
      $\hat{\Gamma}(q, a_\phi) \leftarrow$ solver-side AdamW-induced update direction from $\hat g(q, a_\phi)$\;
      $\hat s(q, a_\phi) \leftarrow \mathrm{cossim}\bigl(\hat{g}_{\mathrm{dev}},\, \hat{\Gamma}(q, a_\phi)\bigr)$\;
    }
  }
  \PhaseGenUpdate{
    Assign $\hat s(x)\leftarrow\rho_{\mathrm{inv}}$ for each invalid generation $x\in\cI_\phi$\;
    \ForEach{$d\in\cB_{\mathrm{gen}}$}{
      Compute DuGRPO advantage $\hat{A}_{\mathrm{gen}}(d, x_i) \leftarrow (\hat s(x_i)-\mathrm{mean}\{\hat s(x_j)\}_{j=1}^n)/(\sigma_d+\sigma_{\mathcal{B}}+\epsilon)$ for each generated output $x_i$ from $d$, where valid $x_i=(q^i,a_{\phi}^i)$ use influence rewards and invalid $x_i$ use $\rho_{\mathrm{inv}}$\;
    }
    For each document in $\cB_{\mathrm{gen}}$, treat the $n$ generated questions as a group; Filter out zero-variance documents in $\mathcal{B}_{\mathrm{gen}}$; update $\phi$ by AdamW steps on \eqref{eq:unified_obj} with $\hat{A}_{\mathrm{gen}}$, learning rate $\eta_{\mathrm{gen}}$ and minibatch size $M$\;
  }
  \PhaseSolUpdate{
    For each question in $\cQ_\phi$, treat the $n$ sampled answers as a group; 
    Filter out zero-variance questions in $\cQ_\phi$; update $\theta$ by AdamW steps on \eqref{eq:unified_obj} with $\hat{A}_{\mathrm{sol}}$ (computed in Phase~2), learning rate $\eta_{\mathrm{sol}}$ and minibatch size $M$\;
  }
}
\Return Trained solver $\pi_\theta$ and generator $\pi_\phi$\;
\end{algorithm}

\FloatBarrier

\paragraph{Implementing Phase~3 under FSDP.}
The per-question direction $\hat\Gamma(q, a_\phi)$ in Line~21 of
\Cref{alg:infuser_full} is obtained by reusing the standard FSDP
forward/backward path rather than per-sample autograd: each generated
question is processed as its own mini-batch, and an in-place optimizer
hook combines the resulting sharded gradient with the actor's live
AdamW second-moment state to form $\hat\Gamma(q, a_\phi)$ on the fly
without materializing parameter-sized per-question tensors. The
FSDP, microbatching, and memory-budget details are given in
\S\ref{app:per_question_gradient}.

\subsection{Policy-gradient view of the influence reward}
\label{app:influence_reinforce}

Equation~\eqref{eq:optimizer_aware_influence} has the standard
policy-gradient form once the optimizer-aware influence score is treated
as a sampled scalar reward. Treating $d$ as the state,
$(q, a_\phi)$ as the action, and $s(q, a_\phi)$ as the return, the
REINFORCE policy-gradient estimator~\citep{williams1992simple} is
\begin{equation}
  \nabla_\phi
  \mathbb{E}_{d \sim \mathcal{D}_{\mathrm{doc}},\,
              (q, a_\phi) \sim \pi_\phi(\cdot \mid d)}
  \big[ s(q, a_\phi) \big]
  =
  \mathbb{E}_{d \sim \mathcal{D}_{\mathrm{doc}},\,
              (q, a_\phi) \sim \pi_\phi(\cdot \mid d)}
  \bigl[s(q, a_\phi) \nabla_\phi \log \pi_\phi(q, a_\phi \mid d)\bigr].
  \label{eq:influence_reinforce}
\end{equation}
This expression is only a policy-gradient view of why the influence
score can serve as a generator reward. The implemented generator update
uses the DuGRPO objective in \S\ref{sec:loop}.

\subsection{Generator Question Quality Example}
\label{app:gen_quality_examples}

\paragraph{A qualitative example}
Beyond the aggregate gap, we examine \emph{what} about the
generator's questions improves over training.
\Cref{fig:gen_quality_examples} shows two questions produced by
INFUSER's generator from the same source document, one at the
beginning of training and one after $90$ iterations. The comparison
illustrates the quality improvement along two axes that the aggregate
strong-minus-base gap does not distinguish.
First, \emph{self-containedness}: the base-model question refers to
``equation~(14.2)'' and ``equation~(14.3)'' without stating the
self-energy expression or the barrier-crossing rate law, whereas the
checkpoint~$90$ question restates the relevant equations inside the
stem.  Second, \emph{factual correctness of the ground-truth key}: the
base-model question marks increasing the ion radius as the answer even
though it lowers $\Delta G$ and raises the flux, while the
checkpoint~$90$ answer is consistent with the
$-q^{2}/(\varepsilon_{\mathrm{h}}\, l)$ term.  Together, these changes
convert the generator's output from ``ill-posed hard'' into
``well-posed with high quality,'' the regime that the aggregate
strong-minus-base gap in \Cref{fig:gen_quality_qw8bb} is designed to
detect.

\subsection{Instruction-Finetuned Anchor Extension}
\label{app:instruction_finetune_details}

This appendix gives the setup details omitted from the compact
instruction-finetuned (IF) anchor experiment in
\S\ref{sec:instruction_finetune}.

\paragraph{Goal}
The main experiments start from pretrained base anchors
(Qwen3-4B-Base and Qwen3-8B-Base).  The IF extension asks whether
INFUSER still improves a model whose next-token distribution has
already been reshaped by supervised instruction tuning, and whether
the learned generator continues to help beyond solver-only DrGRPO with
a frozen generator.

\paragraph{Anchor, data, and variants}
We use OLMo-3-7B-Instruct-SFT~\citep{olmo3} as the IF anchor.  The
document pool $\mathcal{D}_{\mathrm{doc}}$ and the
$800$-question SuperGPQA Science development set
$\mathcal{D}_{\mathrm{dev}}$ are identical to those used in
\S\ref{sec:exp:main_benchmarks}.  We compare three variants:
\textbf{Base}, the OLMo-3-7B-Instruct-SFT checkpoint with no
additional RL training; \textbf{Fix-gen}, a solver-only DrGRPO run
with the generator frozen at its initial checkpoint; and
\textbf{INFUSER}, which updates the generator using preconditioned
cosine influence.  Both training runs use solver learning rate
$2\times 10^{-6}$; INFUSER uses generator learning rate
$4\times 10^{-6}$.  Unless noted otherwise, the runs inherit the main
training configuration: $T=100$ iterations, document batch size
$B=128$, group size $n=8$ for both generator and solver rollouts,
AdamW with weight decay $0.01$, and mini-batch size $32$.

\paragraph{Evaluation and checkpoint selection}
All three columns are evaluated with the same benchmark suite,
prompting, sampling, and answer-extraction pipeline as
\Cref{tab:main_benchmarks}.  The suite contains the same
general-reasoning, math/physics, medical, and coding benchmarks used
for the base-anchor experiments.  The Fix-gen and INFUSER columns in
\Cref{tab:olmo_instruct_sft} use the same best-checkpoint selection
protocol as \Cref{tab:main_benchmarks}: every $5$ training iterations
we score the run on the held-fixed validation set described in
\S\ref{app:eval_protocol:ckpt_selection} and report the iteration with
the highest validation accuracy, which lands at checkpoint~$85$ for
both runs.

\paragraph{Choice of IF anchor}
We choose OLMo-3-7B-Instruct-SFT instead of an instruction-finetuned
Qwen3 checkpoint for attribution.  Qwen3 instruct checkpoints have
already undergone large-scale supervised finetuning followed by RL or
teacher distillation, and their post-training corpus is not public.
Consequently, additional gains or regressions from INFUSER would be
hard to separate from unknown post-training data and objectives.
OLMo-3 releases its training recipe and instruction-tuning mixture,
which lets us audit whether $\mathcal{D}_{\mathrm{doc}}$ or
$\mathcal{D}_{\mathrm{dev}}$ overlaps with the anchor's SFT data.

\paragraph{Contamination audit}
We audit the released OLMo-3 SFT mixture,
\path{allenai/Dolci-Instruct-SFT}~\citep{allenai2025dolci}, using the
near-duplicate protocol in \S\ref{app:dedup_check}.  The audit builds word-$13$-gram
MinHashLSH indexes at Jaccard threshold $0.8$ and
MinHashLSH-Ensemble containment indexes at threshold $0.8$ over
$\mathcal{D}_{\mathrm{doc}}$ and $\mathcal{D}_{\mathrm{dev}}$, then
queries the released Dolci samples against both sets.  The scan finds
zero matches for either $\mathcal{D}_{\mathrm{doc}}$ or
$\mathcal{D}_{\mathrm{dev}}$, so the IF-anchor gains in
\S\ref{sec:instruction_finetune} cannot be explained by direct
near-duplicate leakage into the OLMo-3 SFT mixture under this
protocol.

\paragraph{Full results}
\Cref{tab:olmo_instruct_sft} reports the per-benchmark accuracies
behind the radar plots in \Cref{fig:olmo_sft_radars}.  INFUSER leads
on $10$ of the $13$ benchmarks and has the highest overall average
($29.8$ vs.\ $28.7$ for Fix-gen and $27.7$ for Base).  Its largest
category lift is on general reasoning ($+4.2$ over Base), followed by
math/physics reasoning ($+1.8$).  Medical gains are small but
positive, while coding remains essentially tied with Base, consistent
with neither the development set nor the document pool covering code.

\FloatBarrier

\section{Derivation of the Influence Score}
\label{app:influence_derivation}

This appendix gives the detailed derivation of the per-question
influence score $s(q, a_\phi)$ used by the generator in~\S\ref{sec:influence}.
We first formalise the first-order approximation of the outer
objective and justify dropping the second-order remainder, then
specialise to the AdamW optimiser and introduce the
per-question decoupling surrogate.

\subsection{Second-order remainder of the first-order approximation}
\label{app:second_order_remainder}

Recall that the generator's objective is
$J(\theta^+(\phi))$, where
$\theta^+(\phi) = \theta + \Delta\theta(\phi)$ is the solver's
parameters after one inner-loop update~\eqref{eq:solver_step}.
By Taylor's theorem with the mean-value form of the remainder,
\begin{equation}
  J(\theta^+)
  \;=\;
  J(\theta)
  \;+\;
  \bigl\langle
    \nabla_\theta J(\theta),\;
    \Delta\theta(\phi)
  \bigr\rangle
  \;+\;
  R_2,
  \qquad
  R_2
  \;=\;
  \tfrac{1}{2}\,
  \Delta\theta(\phi)^{\!\top}
  H_J(\tilde{\theta})\,
  \Delta\theta(\phi),
  \label{eq:taylor_full}
\end{equation}
for some $\tilde{\theta}$ on the line segment between $\theta$ and
$\theta^+$, where $H_J(\tilde{\theta}) = \nabla^2_\theta
J(\tilde{\theta})$ is the Hessian of $J$ at $\tilde{\theta}$. The expression for $R_2$ is exact at $\tilde{\theta}$ rather than a higher-order series remainder.  Whenever
$H_J$ is locally bounded in operator norm on this segment,
\begin{equation}
  |R_2|
  \;\leq\;
  \tfrac{1}{2}\,
  \bigl\|H_J(\tilde{\theta})\bigr\|_{\mathrm{op}}\,
  \|\Delta\theta(\phi)\|^2
  \;=\;
  O\!\bigl(\|\Delta\theta(\phi)\|^2\bigr),
\end{equation}
so $|R_2|$ is quadratic in the step size while the first-order term in
\eqref{eq:taylor_full} is linear.

In our setting, the single-step update $\|\Delta\theta(\phi)\|$ is controlled by AdamW's adaptive normalization combined with small learning rates. By construction, the bias-corrected ratio $\hat{m}_t / (\sqrt{\hat{v}_t} + \epsilon)$ is bounded coordinate-wise: in the standard regime $(1-\beta_1) > \sqrt{1-\beta_2}$, each coordinate satisfies $|\hat{m}_t / \sqrt{\hat{v}_t}| \leq (1-\beta_1)/\sqrt{1-\beta_2}$ \citep[\S2.1]{kingma2015adam}, so the gradient term of the AdamW step~\eqref{eq:adamw_step} obeys $\|\eta_s \, \hat{m}_t / (\sqrt{\hat{v}_t} + \epsilon)\|_\infty \lesssim \eta_s \, (1-\beta_1)/\sqrt{1-\beta_2}$. With $\eta_s$ on the order of $10^{-6}$ (see~\S\ref{app:training_config}), several orders of magnitude smaller than typical pre-training learning rates, and a constant prefactor $(1-\beta_1)/\sqrt{1-\beta_2} \approx 3.16$ at our standard $(\beta_1, \beta_2) = (0.9, 0.999)$, the first-order term is a local surrogate whose accuracy improves as $\|H_J\|_{\mathrm{op}}\|\Delta\theta(\phi)\|^2$ becomes small.

\subsection{AdamW preconditioning and per-question decoupling}
\label{app:adamw_preconditioning}

Starting from the first-order approximation~\eqref{eq:taylor}, the
generator's objective reduces to shaping the inner product
$\langle \nabla_\theta J(\theta),\, \Delta\theta(\phi) \rangle$
through its choice of curriculum $\mathcal{Q}_\phi$.  The structure of
this inner product depends on how the optimiser maps the raw gradient
to a parameter step.  We specialise to AdamW here to keep the
derivation concrete and directly aligned with our implementation.

\paragraph{AdamW update decomposition.}
Under our ascent convention for the solver objective $J$, the AdamW
parameter update decomposes into a gradient-dependent term and a
weight-decay term:
\begin{equation}
  \Delta\theta(\phi)
  \;=\;
  \underbrace{
    \eta_s \, \frac{\hat{m}_t}{\sqrt{\hat{v}_t} + \epsilon}
  }_{\text{adaptive gradient step}}
  \;-\;
  \underbrace{
    \eta_s \, \lambda_t \, \theta
  \vphantom{\frac{\hat{m}_t}{\sqrt{\hat{v}_t}}}
  }_{\text{weight decay}},
  \label{eq:adamw_step}
\end{equation}
where $\hat{m}_t$ and $\hat{v}_t$ are the bias-corrected first- and
second-moment estimates that depend on
$\nabla_\theta J(\theta;\,
\mathcal{Q}_\phi)$, and $\lambda_t$ is the weight-decay coefficient at
step $t$.
The weight-decay term $-\eta_s \lambda_t \theta$ is independent of
$\mathcal{Q}_\phi$ and therefore contributes a constant to
$\langle \nabla_\theta J(\theta),\, \Delta\theta(\phi) \rangle$
that does not affect the generator's optimisation over $\phi$.
Dropping it, the generator's objective reduces to maximising
\begin{equation}
  \bigl\langle
    \nabla_\theta J(\theta),\;
    \frac{\hat{m}_t}{\sqrt{\hat{v}_t} + \epsilon}
  \bigr\rangle,
  \label{eq:adamw_inner}
\end{equation}
where the positive scalar $\eta_s$ is absorbed into the argmax over
$\phi$.  This
parallels the SGD case in~\eqref{eq:sgd_decomp}, where the
per-question contribution takes the form
$\langle \nabla_\theta J(\theta),\, g(q, a_\phi) \rangle$.

\paragraph{Batch coupling under AdamW.}
Let $g(q, a_\phi) = \nabla_\theta J(\theta;\, q, a_\phi)$
denote the solver-loss gradient induced by question $q$ (and its generated golden answer $a_\phi$).  Let
$\bar{g}_{\cQ} = \frac{1}{|\cQ|}\sum_{(q, a_\phi) \in \cQ} g(q, a_\phi)$ denote the mean
gradient over $\cQ$, and let $\bar{g}_{\cQ}^{\,2}$ denote its
elementwise square, following the standard AdamW second-moment
update.  AdamW then uses
\begin{equation}
  \hat{m}_t
  =
  \frac{\beta_1 m_{t-1} + (1-\beta_1)\bar{g}_{\cQ}}{1-\beta_1^t},
  \qquad
  \hat{v}_t
  =
  \frac{\beta_2 v_{t-1} + (1-\beta_2)\bar{g}_{\cQ}^{\,2}}{1-\beta_2^t},
  \label{eq:adamw_batch_stats}
\end{equation}
where $m_{t-1}, v_{t-1}$ are the first- and second-moment states from
previous solver update steps and $\beta_1, \beta_2$ are the corresponding decay
rates.  Under these exact AdamW semantics, a question's gradient
$g(q, a_\phi)$ does not contribute to~\eqref{eq:adamw_inner} in a purely
additive way: it changes both the momentum term in the numerator and
the adaptive normaliser in the denominator through $\bar{g}_{\cQ}$,
so its effective contribution depends on the other gradients in
$\cQ$.  This coupling prevents any clean per-sample attribution of the
batch update to individual questions.

\paragraph{Per-question decoupling surrogate.}
To recover a per-sample score, we borrow inspiration
from~\citet{xia2024less} and consider a surrogate objective in which
each question $q$ is evaluated as if it were the only question in the
curriculum.  We use the AdamW second-moment preconditioned direction
associated with $q$ alone (modulo the positive step size $\eta_s$):
\begin{equation}
  \Gamma(q, a_\phi)
  \;=\;
  \frac{
    g(q, a_\phi)
  }{
    \sqrt{
      \bigl(\beta_2 v_{t-1} + (1-\beta_2)\, g(q, a_\phi)^2\bigr) / (1 - \beta_2^t)
    }
    + \epsilon
  }.
  \label{eq:adamw_moments}
\end{equation}
We intentionally omit the first-moment numerator from this per-question
score. The exact AdamW numerator would be
\begin{equation}
  \hat m_t(q, a_\phi)
  \coloneqq
  \frac{\beta_1 m_{t-1} + (1-\beta_1)g(q, a_\phi)}
       {1-\beta_1^t}.
  \label{eq:omitted_adamw_momentum_appendix}
\end{equation}
The carried-over momentum state $m_{t-1}$ is shared across all questions
scored in the same iteration. Including this shared vector would mix
optimizer history into the relative comparison among generated questions,
whereas the goal of the influence score is to measure how each individual
question's gradient aligns with the target direction. The second-moment
term is retained because it provides the AdamW adaptive scaling used by
the implemented similarity optimizer.
In principle we could use
$\langle \nabla_\theta J(\theta),\, \Gamma(q, a_\phi) \rangle$ directly as the
influence score for question $q$.  However, as noted
by~\citet{xia2024less}, this raw inner product introduces a spurious
correlation between sequence length and gradient norm: longer
responses accumulate more tokens in the sum defining $g(q, a_\phi)$ and so
systematically produce larger $\|\Gamma(q, a_\phi)\|$, biasing the score
toward length rather than directional alignment with
$\nabla_\theta J(\theta)$.  We therefore replace the inner product
in~\eqref{eq:adamw_inner} with cosine similarity, recovering the
main-text influence score~\eqref{eq:influence_score}.

\subsection{Per-question gradient computation under FSDP}
\label{app:per_question_gradient}

We now describe how the per-question direction $\Gamma(q, a_\phi)$ in
\eqref{eq:adamw_moments} is computed at LLM scale. The implementation
reuses the standard FSDP forward/backward path, so each iteration's
influence-scoring phase costs roughly one solver-update epoch's worth
of compute on the generated batch, with no extra resident memory
beyond the actor's own gradient and optimizer shards.

\paragraph{Sequential per-question backward.}
For each generated pair
$(q, a_\phi) \in \mathcal{B}_{\mathrm{gen}}$, the solver processes the
question's $n_{\mathrm{sol}} = 8$ answer rollouts as a self-contained
mini-batch and runs one Dr.GRPO forward/backward pass with the loss
in~\eqref{eq:unified_obj} restricted to that question. After the
backward pass, the parameter-gradient buffer of the FSDP-sharded actor
holds exactly the per-question gradient $g(q, a_\phi)$ defined in
\S\ref{app:adamw_preconditioning}, with each rank holding only its
parameter shard. Questions are processed sequentially, and the same
gradient buffer is zeroed and reused between questions so that no
per-sample gradient tensor is ever materialized.

\paragraph{Microbatched gradient accumulation per question.}
Each per-question mini-batch is further split into micro-batches of
size $b_\mu = 4$ rollouts (matching the solver-update micro-batch
in~\S\ref{app:training_config}), with the loss scaled by
$1/(\text{\# micro-batches})$ on each backward pass so that the
accumulated buffer at the end of the mini-batch equals
$g(q, a_\phi)$. A question's $n_{\mathrm{sol}}$ rollouts are first
dispatched evenly across the $\mathrm{dp}$ data-parallel ranks, so each
rank backpropagates through $n_{\mathrm{sol}} / \mathrm{dp}$ rollouts
before the FSDP all-reduce assembles $g(q, a_\phi)$ across shards.
Activation memory in this phase is therefore bounded by the same
envelope as ordinary policy training.

\paragraph{AdamW preconditioning via an in-place optimizer hook.}
Once the buffer holds $g(q, a_\phi)$, the per-question direction
$\Gamma(q, a_\phi)$ is built on the fly without ever materializing a
parameter-sized $\Gamma$ tensor. A lightweight ``similarity optimizer''
takes the place of the real AdamW step for this phase: for every
sharded parameter it reads the live second-moment state $v_{t-1}$ and
decay rate $\beta_2$ from the actor's existing AdamW optimizer, forms
$v_t(q, a_\phi) = \beta_2 v_{t-1} + (1{-}\beta_2)\, g(q, a_\phi)^2$ and
the bias-corrected denominator
$\sqrt{v_t(q, a_\phi) / (1 - \beta_2^t)} + \epsilon$ in place, and
accumulates the local scalars
$\langle \Gamma(q, a_\phi),\, g_{\mathrm{dev}} \rangle$ and
$\|\Gamma(q, a_\phi)\|^2$ on each shard. A single 3-element
\texttt{all\_reduce} per mini-batch then yields the global numerator
and denominator of the cosine similarity in
\eqref{eq:influence_score}; no parameter-sized tensor crosses ranks.

\paragraph{Memory and compute footprint.}
Beyond the resident actor weights, the additional FSDP-shard memory
held during Phase~3 is (i) one shard of the parameter-gradient buffer,
already sized for ordinary training, and (ii) one shard of the
dev-gradient reference $g_{\mathrm{dev}}$ produced once per iteration
when the solver computes $\nabla_\theta \hat{J}(\theta)$
(\S\ref{sec:loop}). The AdamW second-moment shard $v_{t-1}$ is the
actor's existing optimizer state and is loaded onto GPU at the start
of the phase when CPU-offload is enabled. Per-question compute is one
forward and one backward pass on the question's $n_{\mathrm{sol}}$
rollouts, so the wall-clock cost of Phase~3 with
$|\mathcal{B}_{\mathrm{gen}}| = B \cdot n_{\mathrm{gen}} = 128 \cdot 8 = 1024$
generated questions matches one solver-update epoch over the same
$1024 \cdot n_{\mathrm{sol}} = 8192$ rollouts. In our 8B-Base runs on a
single 8$\times$H100 node this places Phase~3 at the same order of
magnitude as a single solver-update sweep, never the dominant cost in
the iteration.

\paragraph{Measured wall-clock breakdown.}
We profile the two main-table runs over all 100 training iterations on
one 8$\times$H100 node. The mean iteration takes
$1{,}730 \pm 541$ seconds for the 4B anchor and
$2{,}761 \pm 498$ seconds for the 8B anchor, where the variation is the
standard deviation across iterations. This corresponds to approximately
384 and 614 GPU-hours for the full 4B and 8B runs, respectively.
\Cref{tab:wall_clock_breakdown} reports the measured phase-level
breakdown.

\begin{table}[H]
\centering
\caption{Measured wall-clock breakdown of one INFUSER iteration for the
Qwen3-4B-Base and Qwen3-8B-Base main-table runs. Both runs use one
8$\times$H100 node. Times are means over 100 iterations; displayed
rows may not sum exactly to the totals because of rounding.}
\label{tab:wall_clock_breakdown}
\resizebox{0.98\textwidth}{!}{%
\begin{tabular}{@{}llrrrr@{}}
\toprule
Phase & Work type & 4B s/iter & 4B \% & 8B s/iter & 8B \% \\
\midrule
Dev reference rollout & Rollout & 193 & 11.1 & 341 & 12.4 \\
Generator rollout & Rollout & 193 & 11.2 & 138 & 5.0 \\
Solver rollout on generated questions & Rollout & 131 & 7.6 & 193 & 7.0 \\
Dev reference gradient & Reference gradient & 415 & 24.0 & 668 & 24.2 \\
Influence scoring & Scoring & 303 & 17.5 & 547 & 19.8 \\
Generator update & Optimizer update & 151 & 8.8 & 245 & 8.9 \\
Solver update & Optimizer update & 334 & 19.3 & 619 & 22.4 \\
Other (glue, logging, and I/O) & Overhead & 10 & 0.6 & 11 & 0.4 \\
\midrule
\textbf{Full iteration} & & \textbf{1,730} & \textbf{100} &
\textbf{2,761} & \textbf{100} \\
\bottomrule
\end{tabular}%
}
\end{table}

Influence scoring proper accounts for 17.5\% and 19.8\% of an iteration
at 4B and 8B, respectively. Its wall-clock cost is $0.91\times$ and
$0.88\times$ that of the corresponding solver update, consistent with
the one-solver-update-epoch estimate above. The dev reference gradient
is the larger INFUSER-specific cost at approximately 24\% for both
anchors; because it requires one backward pass over
$\mathcal{D}_{\mathrm{dev}}$, this component scales linearly with the
development-set size.

\section{Comparison with Alternative Formulations}
\label{app:alternatives}

\paragraph{Black-box outer-loop search.}
One natural approach to \eqref{eq:lower} is to perturb the generator, rerun the solver update, and keep the generator change only if the held-out objective improves. Systems such as \texttt{autoresearch}~\citep{karpathy2026autoresearch} instantiate this pattern for code- and hyperparameter-level experimentation. In our setting, however, such keep-or-discard retraining is too expensive for online curriculum adaptation, because each outer-loop proposal would require a separate inner-loop LLM RL run to estimate its effect on the held-out development objective.

\paragraph{Exact meta-gradient optimization.}
A second approach is exact bilevel differentiation. The objective in \eqref{eq:lower} can be mapped to MAML-style meta-learning~\citep{finn2017model}: both update model parameters on support or training data, then evaluate the post-update model on held-out data. In MAML, tasks $\mathcal{T}_i \sim p(\mathcal{T})$ are sampled from a task distribution, and the meta-learner optimizes an initialization $\vartheta$ through
\[
\min_{\vartheta}
\sum_{\mathcal{T}_i \sim p(\mathcal{T})}
\mathcal{L}^{\mathrm{qry}}_{\mathcal{T}_i}(f_{\vartheta_i'})
\quad
\text{s.t.}\quad
\vartheta_i'
=
\vartheta - \alpha \nabla_{\vartheta}
\mathcal{L}^{\mathrm{sup}}_{\mathcal{T}_i}(f_{\vartheta}).
\]
Our formulation differs in the object being optimized: MAML optimizes the shared initialization $\vartheta$, whereas INFUSER optimizes generator parameters $\phi$ that control the training-data distribution. This also relates to dataset distillation~\citep{wang2018dataset}, which optimizes training data through the learner's update, but does so by directly optimizing a small synthetic dataset rather than a generator policy over curricula. In principle, we could differentiate $J(\theta^+(\phi))$ through \eqref{eq:solver_step}; in practice, exact bilevel differentiation through LLM-scale RL updates is prohibitively expensive, motivating the first-order influence approximation in \S\ref{sec:influence}.

\begin{center}
\small
\captionof{table}{Comparison with alternative formulations. }
\renewcommand{\arraystretch}{1.25}
\begin{tabular}{@{}p{4cm}p{5cm}p{5cm}@{}}
\toprule
\textbf{Formulation} & \textbf{Fits because} & \textbf{Limitation} \\
\midrule
MAML-style meta-learning~\citep{finn2017model}
  & Identical bilevel structure: inner-loop update, outer-loop
    evaluation on held-out data.
  & MAML optimises a shared \emph{initialisation}; we optimise a
    data-generating \emph{policy}.  MAML requires backprop through
    the inner step (Hessian); we use a first-order approximation. \\
\addlinespace
Dataset distillation~\citep{wang2018dataset}
  & Outer loop optimises synthetic training data to maximise
    post-update performance on real data.
  & Distillation optimises fixed data vectors; we optimise a
    \emph{generative model} that produces an unbounded curriculum.
    Distillation typically assumes SGD; we handle AdamW with
    momentum preconditioning. \\
\addlinespace
Zero-sum / adversarial (GAN-like)
  & Generator ``challenges'' the solver.
  & Objectives are aligned, not opposed. The generator is rewarded for
    \emph{helping}, not fooling.  No minimax structure. \\
\addlinespace
\textbf{Bilevel meta-learning} (ours)
  & Both levels aligned toward objective $J$; generator shapes inner-loop
    update dynamics via influence-scored curriculum.  Influence score
    $\approx$ first-order meta-gradient.
  & Requires first-order approximation; exact meta-gradient
    intractable for large LMs. \\
\bottomrule
\end{tabular}
\end{center}

\section{Training and Evaluation Protocol}
\label{app:eval_protocol}

All held-out benchmark scores reported in \S\ref{sec:exp:main_benchmarks}
are produced by the same evaluation pipeline, which evaluates the
trained solver in vLLM with a fixed set of sampling
hyperparameters and a fixed pair of prompt templates (one for
multiple-choice questions and one for free-form answers).  This
appendix specifies this configuration in detail.

\subsection{Benchmarks}
\label{app:benchmarks}

We evaluate INFUSER on four benchmark families: mathematical
reasoning, general reasoning, medical reasoning, and coding.  For
clarity, we enumerate the exact benchmark sources used in
\Cref{tab:main_benchmarks}.

\paragraph{Mathematical reasoning.}
\begin{itemize}
  \item \textbf{MATH500.}  We use the standard 500-problem evaluation
    subset of the MATH benchmark introduced by
    \citet{hendrycks2021math}.
  \item \textbf{AIME2024.}  We evaluate on the 2024 edition of the
    American Invitational Mathematics Examination (AIME), following
    recent zero-data reasoning work that treats the official MAA exam
    problems as a held-out benchmark~\citep{maa2024aime}.
  \item \textbf{AIME2025.}  We likewise evaluate on the 2025 edition of
    the American Invitational Mathematics Examination from the
    Mathematical Association of America~\citep{maa2025aime}.
  \item \textbf{HMMT.}  We use a held-out benchmark assembled from
    official Harvard--MIT Mathematics Tournament problem archives.  In
    our evaluated benchmark, the 93 questions are drawn from the
    February~2025, November~2025, and February~2026 HMMT tournaments,
    which are olympiad-style high-school mathematics contests with
    algebra, geometry, combinatorics, and team-style problem-solving
    rounds~\citep{hmmtarchive,hmmttesting}.
  \item \textbf{OlympiadBench (Math).}  We use the mathematics subset
    of OlympiadBench, an olympiad-level bilingual benchmark spanning
    advanced mathematics and physics~\citep{he2024olympiadbench}.
  \item \textbf{OlympiadBench (Phys).}  We use the physics subset of
    the same OlympiadBench benchmark~\citep{he2024olympiadbench}.
\end{itemize}

\paragraph{General reasoning.}
\begin{itemize}
  \item \textbf{MMLU-Pro.}  We use MMLU-Pro, a more robust and
    reasoning-focused successor to MMLU with harder questions and more
    answer choices~\citep{wang2024mmlupro}.
  \item \textbf{GPQA-Diamond.}  We use the Diamond split of GPQA, a
    graduate-level Google-proof question answering benchmark designed
    to resist superficial pattern matching~\citep{rein2023gpqa}.
  \item \textbf{SuperGPQA.}  We use SuperGPQA, a graduate-level
    reasoning benchmark spanning 285 disciplines~\citep{superGPQA2025}.
  \item \textbf{BBEH.}  We use BIG-Bench Extra Hard (BBEH), a general
    reasoning benchmark designed to replace each BBH task with a
    substantially harder counterpart probing a similar reasoning
    skill~\citep{kazemi2025bbeh}.
\end{itemize}

\paragraph{Medical reasoning.}
\begin{itemize}
  \item \textbf{MedQA.}  We use MedQA, a medical multiple-choice QA
    benchmark collected from professional medical exams, including the
    USMLE setting commonly used in LLM evaluation~\citep{jin2020medqa}.
  \item \textbf{MedXpertQA.}  We use the text-evaluation subset of
    MedXpertQA, an expert-level medical reasoning benchmark spanning
    specialties and body systems; the local benchmark file contains
    2,450 text questions, matching the Text subset described in the
    benchmark paper~\citep{zuo2025medxpertqa}.
\end{itemize}

\paragraph{Coding.}
\begin{itemize}
  \item \textbf{HumanEval+.}  We use HumanEval+, the EvalPlus extension
    of HumanEval with substantially expanded unit tests for more
    rigorous code evaluation~\citep{liu2023evalplus,chen2021humaneval}.
  \item \textbf{LiveCodeBench.}  We use LiveCodeBench, a
    contamination-resistant coding benchmark built from temporally
    fresh competitive-programming problems~\citep{jain2024livecodebench}.
    In our implementation, we choose problems released
    between May~2023 and January~2025 according to the official dataset release notes.
\end{itemize}

\subsection{Checkpoint Selection}
\label{app:eval_protocol:ckpt_selection}

Whenever a result is reported as the ``best checkpoint'' of a training
run, the selection follows a fixed protocol.  We save a checkpoint
every $5$ training iterations and score each saved checkpoint on a
$2000$-question validation set $\mathcal{D}_{\mathrm{val}}$ that is
held fixed across all methods, anchors, and ablations.
$\mathcal{D}_{\mathrm{val}}$ is distinct from the dev set
$\mathcal{D}_{\mathrm{dev}}$ that supplies the influence anchor and is
not used in the per-iteration generator update.  We then keep the
checkpoint with the highest validation accuracy and evaluate
\emph{only that checkpoint} on the held-out benchmarks in
\S\ref{sec:exp:main_benchmarks} and \S\ref{sec:comparison}.  The same
every-$5$-iterations schedule and the same $\mathcal{D}_{\mathrm{val}}$
are used for the rerun R-Zero and AZR baselines.

\paragraph{Composition of $\mathcal{D}_{\mathrm{val}}$.}
$\mathcal{D}_{\mathrm{val}}$ is a stratified sample (random seed $42$)
drawn from a broader benchmark pool, with the per-source quotas in
\Cref{tab:val_set_composition}.  The quotas are chosen to give roughly
balanced signal across the math, general-reasoning, and medical
benchmark families that we report on in \Cref{tab:main_benchmarks}.
HumanEval+, LiveCodeBench, and MedQA are deliberately excluded from
$\mathcal{D}_{\mathrm{val}}$, so the entire coding category and one of
the two medical benchmarks remain fully out of sample for checkpoint
selection.  Because the quotas are sampled from the benchmark sources
themselves, individual $\mathcal{D}_{\mathrm{val}}$ questions can
overlap with the corresponding held-out evaluation set; we treat this
as a known limitation, partially mitigated by the small size of
$\mathcal{D}_{\mathrm{val}}$ relative to the full evaluation suite and
by holding the same $\mathcal{D}_{\mathrm{val}}$ fixed across all
methods, anchors, and ablations so any selection bias applies
uniformly.

\begin{table}[!t]
  \centering
  \caption{Per-source quotas of the $2000$-question validation set
    $\mathcal{D}_{\mathrm{val}}$, sampled with seed $42$ from the
    listed benchmark sources.}
  \label{tab:val_set_composition}
  \small
  \begin{tabular}{l r l r}
    \toprule
    Source & Count & Source & Count \\
    \midrule
    AIME (2024, 2025) & 60  & MedXpertQA (text) & 275 \\
    GPQA-Diamond      & 198 & OlympiadBench (Math, Phys) & 275 \\
    HMMT              & 93  & BBEH (MCQ)        & 70  \\
    MMLU-Pro (test)   & 275 & BBEH (open)       & 205 \\
    MATH-500          & 275 & SuperGPQA (all)   & 274 \\
    \midrule
    \multicolumn{3}{r}{\textit{Total}} & $2{,}000$ \\
    \bottomrule
  \end{tabular}
\end{table}

\subsection{Sampling Hyperparameters}
\label{app:eval_protocol:sampling}

For every benchmark and every trained model we sample from the solver with temperature $0.7$,
top-$p$ $0.8$, top-$k$ $20$, prompt length $4096$, and response length
$8192$. These settings follow the official Qwen3 non-thinking-mode
recommendations~\citep{qwen2025technicalreport} for the sampling parameters supported by our evaluation pipeline.
The only exceptions are the two coding benchmarks (HumanEval+ and LiveCodeBench),
where we extend the response length to $16384$ to accommodate longer code generations.

The number of samples drawn per question, $n$, depends on the
benchmark. For most benchmarks we decode $n = 1$ response per
question; for benchmarks with smaller question sets or higher variance we decode
multiple responses and report the average accuracy:

\begin{center}
\begin{tabular}{ll}
  \toprule
  Benchmark & $n$ samples/question \\
  \midrule
  AIME2024, AIME2025, HMMT & 32 \\
  HumanEval+ & 8 \\
  GPQA-Diamond & 5 \\
  MATH500 & 4 \\
  LiveCodeBench & 2 \\
  all other benchmarks & 1 \\
  \bottomrule
\end{tabular}
\end{center}

\subsection{Prompt Templates}
\label{app:eval_protocol:prompts}

Each question is rendered into a chat conversation with a system turn
and a user turn, and the resulting messages are tokenized via the
model's chat template before being sent to vLLM.  We use one of two
templates, chosen based on the benchmark's answer type.

\newtcolorbox{prompttemplatebox}[1]{
  enhanced,
  breakable,
  colback=sectionblue!3!white,
  colframe=sectionblue,
  colbacktitle=sectionblue,
  coltitle=white,
  boxrule=0.7pt,
  arc=10pt,
  left=8pt,
  right=8pt,
  top=8pt,
  bottom=8pt,
  title={#1},
  fonttitle=\bfseries,
}

\newtcblisting{promptmessage}[2][]{
  enhanced,
  breakable,
  listing only,
  colback=white,
  colframe=gray!55!black,
  colbacktitle=gray!55!black,
  coltitle=white,
  boxrule=0.45pt,
  arc=6pt,
  left=5pt,
  right=5pt,
  top=5pt,
  bottom=5pt,
  title={#2},
  fonttitle=\bfseries\small,
  listing options={
    basicstyle=\ttfamily\small,
    breaklines=true,
    columns=fullflexible,
    keepspaces=true,
  },
  #1
}

\paragraph{Multiple-choice questions (MCQ).}
This template is used for INFUSER training and for MCQ-type benchmarks such as MMLU-Pro, GPQA-Diamond, SuperGPQA, BBEH, MedQA, and
MedXpertQA.  The system turn fixes the output contract, and the user
turn wraps the question with step-by-step instructions and requires the final letter to be enclosed in
\texttt{\textbackslash boxed\{\}}:

\begin{prompttemplatebox}{MCQ Prompt Template}
\begin{promptmessage}[colback=sectionblue!5!white,
colframe=sectionblue!55!black,
colbacktitle=sectionblue!65!black]{System}
You are a knowledgeable assistant that solves multiple choice
questions step by step. Always show your reasoning and put your
final answer letter in \boxed{}.
\end{promptmessage}

\medskip

\begin{promptmessage}[colback=findingaccent!8!white,
colframe=findingaccent!70!black,
colbacktitle=findingaccent!70!black]{User}
Solve the following multiple choice question step by step.

{question}

Think through this problem carefully. Show your reasoning
process, then provide your final answer.

IMPORTANT: Your final answer MUST be enclosed in \boxed{}
using ONLY the letter of the correct choice (A, B, C, D, etc.).

Example format for your final answer:
\boxed{<correct choice letter>}

Now solve the problem:
\end{promptmessage}
\end{prompttemplatebox}

At scoring time the first \texttt{\textbackslash boxed\{...\}} span
in the response is parsed and compared against the gold letter; all
MCQ benchmarks in \Cref{tab:main_benchmarks} are graded by exact
match on the extracted letter.

\paragraph{Free-form questions.}
This template is used for MATH500, AIME2024, AIME2025, HMMT, OlympiadBench (Math and
Phys), HumanEval, and LiveCodeBench.  The template mirrors the MCQ
variant but asks for the final answer itself (number, expression, or
code) inside \texttt{\textbackslash boxed\{...\}}:

\begin{prompttemplatebox}{Free-Form Prompt Template}
\begin{promptmessage}[colback=sectionblue!5!white,
colframe=sectionblue!55!black,
colbacktitle=sectionblue!65!black]{System}
You are a knowledgeable assistant that solves questions step
by step. Always show your reasoning and put your final answer
in \boxed{}.
\end{promptmessage}

\medskip

\begin{promptmessage}[colback=findingaccent!8!white,
colframe=findingaccent!70!black,
colbacktitle=findingaccent!70!black]{User}
Solve the following question step by step.

{question}

Think through this problem carefully. Show your reasoning
process, then provide your final answer.

IMPORTANT: Your final answer MUST be enclosed in
\boxed{<answer>}

Now solve the problem:
\end{promptmessage}
\end{prompttemplatebox}

For math benchmarks, the extracted \texttt{\textbackslash boxed\{...\}}
span is passed through a programmatic equivalence checker, with an optional GPT-4o-class LLM judge as a tie-breaker for MATH500. For HumanEval and LiveCodeBench, the extracted span is treated as the
candidate program and executed against the benchmark's unit tests in a
sandboxed subprocess.

\paragraph{Generator prompt used during training.}
The main training runs use document-conditioned question generation
(\texttt{question\_source\_mode=document}, the default setting in the
training config).  For each sampled document, the pipeline checks
an explicit per-document \texttt{prompt\_type} tag; untagged documents use the
default MCQ prompt.  Thus science documents typically use the MCQ prompt
below, while math documents marked as \texttt{free\_form} use
the analogous free-form prompt whose JSON schema has no \texttt{choices}
field and instead sets \texttt{benchmark\_type=qa\_open} and
\texttt{data\_source=math}.  In both cases, the full document text is
inserted into the \texttt{\{text\}} slot and the mixed parser dispatches
the generated question to the corresponding solver/verifier path.

\begin{tcolorbox}[
  enhanced,
  colback=sectionblue!3!white,
  colframe=sectionblue,
  colbacktitle=sectionblue,
  coltitle=white,
  boxrule=0.7pt,
  arc=10pt,
  left=8pt,
  right=8pt,
  top=8pt,
  bottom=8pt,
  title={Generator Prompt Template (Document-Conditioned)},
  fonttitle=\bfseries,
]
The generator prompt is shown as the following two chat turns.
\end{tcolorbox}

\begin{promptmessage}[colback=sectionblue!5!white,
colframe=sectionblue!55!black,
colbacktitle=sectionblue!65!black]{System}
You are an expert educator creating challenging multiple-choice questions. Always output valid JSON with the exact structure requested.
\end{promptmessage}

\medskip

\begin{promptmessage}[colback=findingaccent!8!white,
colframe=findingaccent!70!black,
colbacktitle=findingaccent!70!black]{User}
Your task is to create CHALLENGING exam questions from a document by identifying complex relationships and multi-step reasoning paths.

## Document
[BEGINNING OF THE DOCUMENT]
{text}
[END OF THE DOCUMENT]

## Instructions

### Step 1: Complex Information Extraction for MCQ Design

**PRIORITY: Focus on information that enables multiple plausible interpretations and requires synthesis.**

Scan the text and identify information that naturally creates opportunities for sophisticated multiple-choice questions:

**Ideal MCQ content requires:**
* **Synthesis opportunities**: Relationships between 3+ concepts spanning different sections, implicit conclusions requiring combination of multiple facts, systems where changing one parameter affects others
* **Multi-step reasoning paths**: Processes with intermediate steps (each step = potential distractor), calculations with sequential dependencies, procedures requiring decision points about when/how to apply methods
* **Rich comparison spaces**: Comparative analyses revealing subtle distinctions ("however," "but," "except," "unlike"), trade-offs between approaches, overlapping categories or edge cases, prerequisites or conditional relationships
* **Application complexity**: Principles applied to novel scenarios, cause-and-effect chains with intermediate stages, mechanisms where partial understanding yields plausible-but-incomplete explanations
* **Domain-specific depth**: Multi-variable calculations (unit conversions, stoichiometry, equilibrium perturbations), classification problems with boundary conditions, experimental design with multiple controlling factors, predictions integrating multiple scientific laws

**AVOID** (these create poor MCQ questions):
* Single, directly stated facts that allow simple lookup
* Simple definitions that stands alone
* Values or numbers mentioned in isolation
* Information that requires no synthesis
* Lists without relationships between items
* Trivial categorizations
* Information where all wrong answers would be obviously implausible

### Step 2: Difficulty Enhancement Process

**EXPLICITLY STATE YOUR HARDENING PROCESS** Before generating the question, describe your strategy to make it harder:
1. What simple version would you avoid?
2. What complexity layers will you add?
3. Are there any seamless traps for common misconceptions to exploit for distractors?
4. How can you leverage subtle, non-obvious interactions between different content elements to create more engaging and intellectually demanding questions?
5. What common shortcuts will you block?
6. How will you ensure multi-step reasoning is required?

### Step 3: Advanced Question Generation

Generate ONE high-quality MCQ question that:
* Requires applying multiple concepts from different parts of the document
* Tests understanding of relationships, not just recall of facts
* Forces reasoning through multiple steps to reach the answer
* May require comparing or contrasting different scenarios
* Could involve "what if" scenarios based on principles in the text
* Tests ability to apply concepts to slightly modified situations

**CRITICAL - Self-Contained Requirements**:
* Questions must be 100% self-contained and standalone
* NEVER use: "according to the document", "in the document", "as mentioned", "the passage states", "based on the analysis", etc.
* Write as if for a formal exam with no reference material
* Include all necessary context within the question itself, but don't reveal any intermediate reasoning steps or key insights that would make the question easy
* Define any specialized terms if needed for clarity

### Step 4: Difficulty-Driven Design

**TARGET: Generate HARD/EXTRA HARD questions by design**
* HARD: Synthesize 4+ concepts; multi-step problem solving; pattern recognition
* EXTRA HARD: Complex system analysis; counter-intuitive applications; edge cases

Design questions that CANNOT be answered by:
* Looking up a single fact
* Finding one sentence with the answer
* Simple keyword matching

### Step 5: Knowledge Integration Requirements

Document the reasoning path that shows why this is a difficult question:
* List 3+ distinct pieces of information needed from different parts
* Show the logical connections required between these pieces
* Explain why simple lookup won't work
* Include intermediate reasoning steps

### Step 6: Multiple Choice Design Guidelines

Create between 4 and 8 answer choices following these STRICT rules:

**Length Balance**: All options must be approximately equal length (+/-20%)
**Unit Consistency**: All numerical answers must use identical units and formatting
**Tone Neutrality**: Avoid overly certain language ("definitely", "always", "never") unless justified
**Plausibility**: All distractors must be genuinely plausible based on partial understanding
**More choices increase difficulty**: Use 6-8 choices for complex questions

**Distractor Design**:
* Common calculation errors from the multi-step process
* Results from applying only partial reasoning
* Mixing up related concepts from the document
* Reasonable approximations that miss key factors

### Step 7: Self-Testing Filter (AFTER MCQ Creation)

**SOLVE YOUR OWN MCQ AS A STUDENT WOULD** Now test the complete multiple choice question:
1. What's the quickest path a student might try with these options?
2. Can you eliminate 2+ options without full understanding? If yes, redesign distractors
3. Does seeing the options make the answer obvious? If yes, improve distractors
4. Count the reasoning steps required even with options visible - if less than 3, REJECT
5. Time estimate: Would this MCQ take <30 seconds? If yes, make it harder
6. Could a student guess correctly by pattern matching the options? If yes, rebalance

### Step 8: Final Complexity Verification

Before finalizing, verify your question is NOT Easy by checking:
* Can it be answered by finding one sentence? If yes, redesign
* Does it require connecting multiple document sections? If no, add complexity
* Would someone need to understand relationships, not just facts? If no, refocus
* Are all MCQ options balanced and using consistent formatting? If no, revise
* Did your self-test of the MCQ take more than 1 minute? If no, increase difficulty

## Output Format

You MUST output ONLY a valid JSON object with this exact structure:

{{
    "question_text": "Your complete, self-contained question here?",
    "choices": [
        "First choice text (without letter prefix)",
        "Second choice text (without letter prefix)",
        "Third choice text (without letter prefix)",
        "Fourth choice text (without letter prefix)",
        "Fifth choice text (optional)",
        "Sixth choice text (optional)",
        "Seventh choice text (optional)",
        "Eighth choice text (optional)"
    ],
    "ground_truth": "The exact text of the correct choice (must match one of the choices exactly)",
    "difficulty": "hard",
    "answer_quote": [
        "Relevant quote 1 from the document showing key information",
        "Relevant quote 2 from the document showing different piece needed",
        "Relevant quote 3 (include multiple quotes showing different pieces needed)"
    ],
    "hardening_process": "Your explicit strategy for making this question difficult (from Step 2)",
    "knowledge_and_reasoning_steps": "Detailed reasoning path showing why this is Hard/Extra Hard difficulty",
    "self_test_solution": "Your step-by-step solution of the MCQ showing the difficulty (from Step 7)"
}}

Field descriptions:
- "question_text": A challenging, self-contained question requiring synthesis. Return empty string if document lacks sufficient complexity.
- "choices": Array of 4-8 answer options without letter prefixes
- "ground_truth": The exact text of the correct choice (MUST match one of the choices exactly)
- "difficulty": Target difficulty level (hard or extra_hard)
- "answer_quote": Multiple verbatim quotes from the document showing the different pieces needed (not just one quote)
- "hardening_process": Your explicit strategy for making this question difficult (from Step 2)
- "knowledge_and_reasoning_steps": Detailed reasoning path showing why this is Hard/Extra Hard difficulty
- "self_test_solution": Your step-by-step solution of the MCQ showing the difficulty (from Step 7)

CRITICAL RULES:
1. Your final answer must contain exactly one JSON object matching the requested schema. Put this JSON object last.
2. The "choices" array must have at least 4 items and at most 8 items
3. Do NOT include letter prefixes (A), B), etc.) in the choices
4. "ground_truth" must be the exact text of one of the choices
5. The question must be answerable, with the key component or solution step within the document content
6. If the document lacks sufficient complexity, return empty strings for all fields

Schema illustration only (do not copy this content):
{{"question_text": "<new self-contained question derived from the document>", "choices": ["<choice 1>", "<choice 2>", "<choice 3>", "<choice 4>"], "ground_truth": "<exact text of one choice>", "difficulty": "hard", "answer_quote": ["<document quote 1>", "<document quote 2>", "<document quote 3>"], "hardening_process": "<difficulty strategy>", "knowledge_and_reasoning_steps": "<reasoning path>", "self_test_solution": "<step-by-step solution>"}}.
\end{promptmessage}

\paragraph{Free-form math route.}
This is the case for the math RLVR and INFUSER hybrid runs in \S\ref{sec:rlvr_hybrid}, which use the same training pipeline but with \texttt{prompt\_type=free\_form} for all math RLVR questions.
For \texttt{prompt\_type=free\_form} math documents, the user prompt keeps the same
document-conditioned structure but asks for one machine-verifiable math
question.  Its JSON object replaces the MCQ \texttt{choices} field with
an answer type, marks the example as open-ended math QA, requires
\texttt{ground\_truth} to be a single concise mathematical answer without
units, prose, lists, or surrounding \texttt{\textbackslash boxed\{\}}, and
routes the resulting question through the math verifier path.

\section{Training configurations for compared methods}
\label{app:method_configurations}
\label{app:training_config}

We compare \textbf{INFUSER} with the base model and four contemporaneous self-evolution methods:
R-Zero~\citep{huang2025rzero}, AZR~\citep{zhao2025azr}, R-Few~\citep{yu2025rfew}, and SPICE~\citep{fang2025spice} on Qwen3-4B-Base and Qwen3-8B-Base as anchors.
We report
INFUSER as the mean over three random seeds, selecting the best
checkpoint within each run by accuracy on a small validation set
evaluated every $5$ training iterations (\S\ref{app:eval_protocol:ckpt_selection}).
For R-Zero and AZR, we rerun their released training
code under the original settings: $5$ R-Zero iterations and $500$
AZR training steps.
R-Few and SPICE are self-reported because public training code is unavailable, and should be read with caution.
Results
are summarized in Table~\ref{tab:main_benchmarks}.

\Cref{tab:method_configurations} compares the training configuration of
\textbf{INFUSER} side-by-side with the three open-source baselines we
rerun in \S\ref{sec:exp:main_benchmarks}: the two self-evolution
methods R-Zero \citep{huang2025rzero} and Absolute Zero Reasoner (AZR)
\citep{zhao2025azr}, together with General-Reasoner (GR)
\citep{ma2025generalreasoner}, a Zero-style RLVR baseline that we
include for completeness even though it relies on frontier-model
curation rather than self-evolution. We follow the row layout of
\citet[Table~5]{fang2025spice} so the configuration contrast is
explicit. The mapping between role names is INFUSER's
\emph{generator} $\leftrightarrow$ R-Zero \emph{challenger}
$\leftrightarrow$ AZR \emph{proposer}, and INFUSER's \emph{solver}
$\leftrightarrow$ R-Zero \emph{reasoner} $\leftrightarrow$ AZR
\emph{solver} $\leftrightarrow$ GR \emph{actor}; GR trains a single
solver-only actor on a fixed curated question pool and has no
generator role.

INFUSER values are taken from our Qwen3 training runs (LR rows list
the Qwen3-4B-Base and Qwen3-8B-Base settings); R-Zero and AZR values
are reproduced from the configurations released with their training
code, cross-checked against \citet[Table~5]{fang2025spice} where they
re-ran both baselines; GR values are the settings of our Qwen3-8B-Base
rerun, which mirrors the Qwen3-14B-Base column of
\citet[Table~9]{ma2025generalreasoner} on a single 8$\times$H100 node.

\begin{table}[!t]
\centering
\caption{Training configurations for INFUSER and the three open-source
baselines we rerun. The \emph{Generator} role corresponds to the
Challenger in R-Zero and to the Proposer in AZR; the \emph{Solver}
role corresponds to the Reasoner in R-Zero, to the Solver in AZR, and
to the single-actor in GR (which has no generator). Slash-separated
entries are role-specific or model-scale-specific as noted below.}
\label{tab:method_configurations}
\label{tab:training_config}
\small
\renewcommand{\arraystretch}{1.15}
\resizebox{0.98\textwidth}{!}{%
\begin{tabular}{@{}lcccc@{}}
\toprule
\textbf{Configuration} & \textbf{INFUSER} & \textbf{R-Zero} & \textbf{AZR} & \textbf{GR} \\
\midrule
\multicolumn{5}{@{}l}{\textit{Data Source}} \\
Corpus documents
  & 12{,}260
  & --
  & --
  & -- \\
Dev set size ($|\mathcal{D}_{\mathrm{dev}}|$)
  & 800
  & --
  & --
  & 100\textsuperscript{g} \\
Question source
  & Doc-grounded
  & Self-generated
  & Self-generated
  & WebInstruct-verified\textsuperscript{g} \\
External grounding
  & \checkmark
  & $\times$
  & Python executor
  & 1.5B verifier model\textsuperscript{g} \\
\midrule
\multicolumn{5}{@{}l}{\textit{Training Details}} \\
Generator training
  & \checkmark
  & \checkmark
  & \checkmark
  & $\times$ \\
Generator sampling ($n$)
  & 8
  & 4
  & 1
  & -- \\
Solver training
  & \checkmark
  & \checkmark
  & \checkmark
  & \checkmark \\
Solver sampling ($n$)
  & 8
  & 5
  & 1\textsuperscript{a}
  & 8 \\
Temperature
  & 0.7
  & 1.0
  & 1.0
  & 0.7 \\
Algorithm
  & Dr.GRPO / DuGRPO
  & GRPO
  & REINFORCE++
  & GRPO \\
Optimizer
  & AdamW
  & AdamW
  & AdamW
  & AdamW \\
Generator learning rate
  & $6/4 \times 10^{-6}$\textsuperscript{b}
  & $1 \times 10^{-6}$
  & $1 \times 10^{-6}$\textsuperscript{c}
  & -- \\
Solver learning rate
  & $2 \times 10^{-6}$
  & $1 \times 10^{-6}$
  & $1 \times 10^{-6}$\textsuperscript{c}
  & $5 \times 10^{-7}$ \\
LR schedule
  & Constant (no warmup)
  & Constant (no warmup)
  & Constant (no warmup)
  & Constant (no warmup) \\
Mini-batch size
  & 32
  & 16 / 128\textsuperscript{d}
  & 384\textsuperscript{c}
  & 256 \\
Rollout correction
  & Token-level TIS ($\rho_{\max}{=}2.0$)
  & None
  & None
  & None \\
\midrule
\multicolumn{5}{@{}l}{\textit{Reward Design}} \\
Generator reward
  & Influence (precond.\ cosine)
  & $1{-}2\lvert p{-}0.5\rvert$
  & $1{-}p$ if $0{<}p{<}1$, else $0$
  & -- \\
Solver reward
  & Binary correctness
  & Binary (vs.\ pseudo-label)
  & Binary (vs.\ executor)
  & Binary (vs.\ verifier model)\textsuperscript{g} \\
Invalid penalty
  & 0.0
  & $-1$ (Challenger)
  & $-0.5$ / $-1^{\ast}$
  & $0$ \\
\midrule
\multicolumn{5}{@{}l}{\textit{Performance}} \\
Training iterations
  & 100
  & 5\textsuperscript{$\dagger$}
  & 500 (steps)
  & 3 epochs ($=$669 steps) \\
Batch size
  & 128 docs
  & 8{,}000 generated questions\textsuperscript{e}
  & 64 tasks
  & 1{,}024 questions \\
GPUs
  & $8\times$ H100 80\,GB
  & 4 / 8 GPUs\textsuperscript{d}
  & $2/4\times$ 80\,GB GPUs\textsuperscript{f}
  & $8\times$ H100 80\,GB \\
\bottomrule
\end{tabular}%
}
\smallskip
\par
\footnotesize
\textsuperscript{a}~AZR uses a single training rollout per task but
estimates learnability from $\sim 8$ Monte-Carlo Solver attempts
\citep[\S 4]{zhao2025azr}.\quad
\textsuperscript{b}~Generator learning rate values are
$6\times 10^{-6}$ for Qwen3-4B-Base and $4\times 10^{-6}$ for
Qwen3-8B-Base.\quad
\textsuperscript{c}~AZR trains a single shared actor policy for
proposal and solution; its training script sets
actor learning rate $1\times 10^{-6}$ and its runtime sets
PPO mini-batch size to $64\times3\times2=384$.\quad
\textsuperscript{d}~R-Zero slash-separated values are
Challenger / Reasoner values; its Challenger PPO uses batch size 16 on
4 GPUs, while Reasoner PPO uses batch size 128 on 8 GPUs.\quad
\textsuperscript{e}~R-Zero generates $1{,}000$ candidate questions on
each of 8 parallel generator workers before self-consistency filtering
and Reasoner training.\quad
\textsuperscript{f}~AZR GPU counts are for Qwen3-4B-Base /
Qwen3-8B-Base runs, respectively.\quad
\textsuperscript{$\dagger$}~R-Zero is reported to degrade after
$\sim 5$ iterations \citep{huang2025rzero}; we report its best
checkpoint within those 5 iterations.\quad
\textsuperscript{$\ast$}~AZR applies $-0.5$ to an incorrect but
well-formatted solver response and $-1$ to a malformed response.\quad
\textsuperscript{g}~GR trains on TIGER-Lab/WebInstruct-verified
\citep{ma2025generalreasoner}: $\sim$230K questions filtered from
$\sim$5M web instructions by a frontier model, with a separately
trained 1.5B generative verifier (TIGER-Lab/general-verifier) that
supplies the binary solver reward; the validation slice is the first
100 rows of the WebInstruct-verified test split.
\end{table}

\section{Pilot Extension: Combining INFUSER with Rule-Verifiable Math RLVR}
\label{app:self_evolve_rlvr_hybrid}

\begin{figure}[!t]
  \centering
  \includegraphics[width=0.7\textwidth]{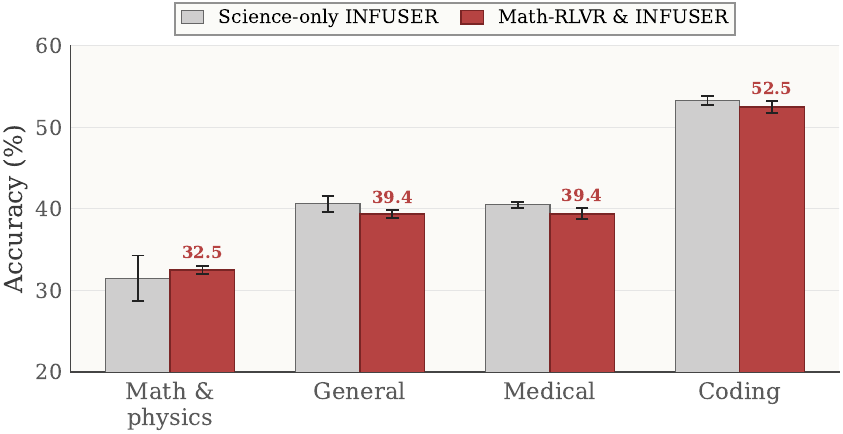}
  \caption{%
    Category-average profile for the pilot mixed science+RLVR runs on
    Qwen3-8B-Base, visualizing the per-seed averages reported in
    \Cref{fig:hybrid_overview}.  Bars show the mean over three
    seeds for each setting; error bars show one cross-seed standard
    deviation.  Verifiable math RLVR tightens the math-and-physics error
    bar (the channel it targets) while the non-math categories dip
    slightly under the reduced science document budget.
  }
  \label{fig:self_evolve_rlvr_hybrid}
\end{figure}

This appendix gives the construction and full per-seed results for the
hybrid science+RLVR pilot summarized in \S\ref{sec:rlvr_hybrid}.  The
pilot starts from the science-document INFUSER setting used in the main
experiments and adds a verifiable mathematics component to both the
influence anchor and the solver-training curriculum, asking whether one
training loop can combine document-grounded science self-evolution with
direct verifiable math RLVR.  The motivating seed instability, the
test-time-compute mechanism ($r = 0.997$ between evaluation-time response
length and math accuracy), and the headline results are presented in
\S\ref{sec:rlvr_hybrid} together with
\Cref{fig:hybrid_overview,fig:hybrid_response_length_qw8bb,fig:self_evolve_rlvr_hybrid}.

\paragraph{Data mixture.}
The mixed run follows the Qwen3-8B-Base INFUSER recipe: solver
learning rate $2\times 10^{-6}$, generator learning rate
$4\times 10^{-6}$, Dr.GRPO solver updates, DuGRPO generator updates,
and preconditioned-cosine influence scoring.  The dev anchor
$\mathcal{D}_{\mathrm{dev}}$ has $800$ questions, split evenly between
sampled SuperGPQA Science MCQs and AIME-history free-form questions.
Science rows use the existing MCQ scoring path.  The AIME dev-anchor rows
have empty choice lists and \texttt{data\_source=aime}, which sends them
to the AIME integer verifier.  This dev anchor should be distinguished
from the training pool: Putnam enters through the training-side math
pool, not through the $800$-row dev anchor.

The training pool combines the original $12{,}260$ science textbook
chunks with $10{,}000$ math rows drawn from Putnam and AIME-history.
The math pool is constructed from $121$ unique Putnam problems and
$918$ unique AIME-history problems, then filled to $10{,}000$ rows by
round-robin repetition and shuffled with seed $42$.  The realized
training mixture contains $1{,}210$ Putnam rows and $8{,}790$
AIME-history rows.  At the data level, the resulting curriculum
juxtaposes document-grounded science sources with verifiable
mathematics: the science side still requires the generator to synthesize
training questions from documents, while the Putnam/AIME-history side
provides externally answered problems that can directly support RLVR.

\paragraph{Training recipe.}
The run uses the same five-phase INFUSER loop as
\Cref{alg:infuser}: compute a dev-set reference gradient on the mixed
science/AIME anchor, build a training batch from the science and math
pool, estimate influence scores for the resulting solver updates,
update the generator with DuGRPO, and update the solver with Dr.GRPO.
Unlabeled science chunks use the default document-conditioned MCQ
generation path, so science supervision still depends on the
generator's ability to convert documents into useful QA pairs.  The
math side supplies externally answered Putnam/AIME-history problems for
the RLVR component: AIME rows use integer answer checking, while
Putnam rows use the math-verification path for free-form mathematical
answers.  The mixed dev anchor supplies the influence-scoring
direction, with science MCQ signal and AIME free-form signal both
present in the dev gradients.  We also add a
small mid-EOS shaping penalty of $-0.5$ to discourage responses that emit
\texttt{<|endoftext|>} before a boxed answer.  This penalty is
additive; it is not a length cap.

\FloatBarrier

\paragraph{Full results.}
The per-seed category averages tabulated in the left panel of the
main-text \Cref{fig:hybrid_overview}, and re-plotted as a profile in
\Cref{fig:self_evolve_rlvr_hybrid}, use the same six-benchmark math
grouping as \Cref{tab:main_benchmarks} (MATH500, AIME2024, AIME2025, HMMT,
OlympiadBench Math, and OlympiadBench Phys).  The three Science-only
INFUSER seeds correspond to the preconditioned-cosine checkpoints
seed456/ckpt95, seed123/ckpt55, and seed42/ckpt95 that underlie the
Qwen3-8B anchor in the main comparison; the three Math-RLVR \& INFUSER
seeds reuse the same seeds.  Averaged over the three
seeds, the mixed setting raises AIME2024 from $18.58\%$ to $21.25\%$ and
the math-and-physics category average from $31.49\%$ to $32.52\%$.  More
importantly, the cross-seed sample standard deviation of the
math-and-physics average drops from $2.80$ to $0.48$ percentage points
(and on AIME2024 from $2.94$ to $0.27$), confirming that verifiable math
RLVR resolves the seed-dependent equilibrium ambiguity diagnosed in
\S\ref{sec:rlvr_hybrid}.  The other categories decline modestly: general
reasoning falls from $40.63\%$ to $39.37\%$, medical from $40.52\%$ to
$39.39\%$, and coding from $53.30\%$ to $52.49\%$.  This tradeoff has a
direct explanation rooted in the fixed total training budget.  The math
pool contributes $10{,}000$ of $22{,}260$ total training rows, roughly
$45\%$ of the curriculum.  Under the same total number of solver training
steps, the solver therefore sees approximately half as many science
documents per training loop compared to the science-only setting.
Reduced exposure to science documents weakens the curriculum signal that
drives general reasoning and out-of-domain transfer, exactly the gains
that science self-evolution delivers in the main experiments.  The
decline is therefore not a sign of interference between the two
objectives, but a predictable consequence of the current budget
allocation.

\paragraph{Response length.}
The bottom row of \Cref{fig:hybrid_response_length_qw8bb} confirms the
mechanism.  Across all three hybrid seeds, response length on AIME,
HMMT, and MATH500 collapses to a tightly clustered trajectory from
early in training.  The verifiable math RLVR signal imposes a
hard constraint on reasoning depth: the solver must produce correct
mathematical answers to earn reward, which prevents the collapse to
short thinking that destabilizes science-only seeds.  The alignment
between length stabilization and accuracy stabilization supports the
interpretation that reasoning-depth equilibrium is the primary
mechanism through which seed variance manifests in science-only math
performance.

\paragraph{Takeaway.}
\Cref{fig:hybrid_overview,fig:self_evolve_rlvr_hybrid}
confirm the feasibility of running INFUSER with a mixed anchor: one
training loop can jointly handle document-grounded science
self-evolution and verifiable math RLVR.  The current ${\approx}45\%$ math
budget allocation stabilizes math performance at the cost of weakened
science-document signal.  This is a budget allocation problem rather
than a fundamental incompatibility; tuning the ratio between math and
science rows is the natural lever for future work aiming to obtain
uniform gains across both dimensions.

\FloatBarrier

\section{Document Pool Construction Pipeline}
\label{app:document_pipeline}

We describe the pipeline that builds the document pool
$\mathcal{D}_{\mathrm{doc}}$ used by \textbf{INFUSER}. The
pipeline is fully automated and consists of five stages:

\begin{enumerate}[label=(\roman*),leftmargin=* ,topsep=2pt,itemsep=2pt]
  \item The development set is parsed into a finite set of
    subdomains that the document pool must cover.
  \item For each subdomain, an external LLM searches the open web and
    downloads open-access textbooks that target it.
  \item Each downloaded PDF is converted to Markdown with a
    layout-aware tool.
  \item The Markdown is split into token-bounded chunks with a
    structural-aware splitter. 
  \item An LLM judge filters out non-essential content.  
\end{enumerate}
The corpus statistics that result from this pipeline (final size
$|\mathcal{D}_{\mathrm{doc}}| = 12{,}260$ chunks, broken down by discipline) are reported in \S\ref{app:dedup_check}.

\paragraph{Taxonomy extraction from the development set.}
Source selection is conditioned on the domain and the subdomain of
each entry in the development set $\mathcal{D}_{\mathrm{dev}}$. Here,
the domain refers to the broad area of study (e.g.\ Physics), and
the subdomain refers to a finer specialization within it
(e.g.\ Quantum Mechanics under Physics), so that every
$(\text{domain}, \text{subdomain})$ pair probed by the dev set
receives dedicated textbook coverage in the pool. We distinguish two
cases according to whether $\mathcal{D}_{\mathrm{dev}}$ already
provides a multi-level taxonomy.
\begin{itemize}[leftmargin=1.5em,topsep=2pt,itemsep=2pt]
  \item \textbf{Built-in taxonomy.} For development sets that already
    carry a multi-level taxonomy, we use it directly. SuperGPQA
    \citep{superGPQA2025}, our running example, annotates every
    question with three nested labels: a top-level
    \texttt{discipline} (e.g.\ Science), a \texttt{field}
    (e.g.\ Physics, Mathematics), and a fine-grained \texttt{subfield}
    (e.g.\ Quantum Mechanics, Ordinary Differential Equations) drawn
    from $285$ subfields in total. We map \texttt{field} to the
    domain and \texttt{subfield} to the subdomain in our
    $(\text{domain}, \text{subdomain})$ representation, and pass both
    levels to the textbook search stage so that the search is guided
    by the broad area and refined by the specialization.
  \item \textbf{LLM-assigned taxonomy.} For general-purpose
    development sets that lack a built-in taxonomy (e.g.\ MedQA), we
    use an external LLM to assign each dev question both a domain and
    a subdomain label, and merge the resulting labels into a finite
    set of $(\text{domain}, \text{subdomain})$ pairs. Because
    labelling occurs entirely on the development side of the pipeline, the
    same external LLM can be reused for the downstream textbook
    search step described below.
\end{itemize}
Both cases produce the same intermediate object: a finite set
$\mathcal{C} = \{(d_1, c_1), \dots, (d_K, c_K)\}$ of
$(\text{domain}, \text{subdomain})$ pairs that the document pool
$\mathcal{D}_{\mathrm{doc}}$ must cover. Individual dev questions are
not used downstream of this step: only the pair set $\mathcal{C}$ is
passed to the textbook search stage.

\paragraph{Open-access textbook search per (domain, subdomain) pair.}
For each $(d, c) \in \mathcal{C}$, the same external LLM acts as a
web-research agent: it issues queries conditioned only on the pair
$(d, c)$ to find open-access textbooks targeting that
specialization within the broad domain, validates the returned URLs
by attempting a download, and stores the downloaded PDFs in a local
source directory. The agent is equipped with web search, URL
crawling, link extraction, and file-download tools. Crucially, the
agent never sees individual dev questions; the search prompt
receives only the $(\text{domain}, \text{subdomain})$ pair. This
keeps any question-level signal out of the source-selection step and
is sufficient to retrieve textbooks that cover the specialization.
In our runs, the external LLM is the latest version of ChatGPT
served through its web interface; any sufficiently capable
conversational LLM with browsing tools is a drop-in replacement.
Two operational constraints govern the resulting source set.
\emph{(a)~Open access.} Only open-access resources are admitted;
materials behind paywalls or other access restrictions are excluded. \emph{(b)~Source-level deduplication.}
Before any download, the agent consults a curated registry that
records every previously downloaded resource and admits a candidate
only if its canonicalised identifier is not already present. As a
result, the same textbook is never ingested twice across runs or
across pairs.

\paragraph{Source registry.}
The agent maintains a curated registry that stores, for each admitted
resource, a stable identifier, title, author, discipline, category
(e.g.\ \texttt{textbook}, \texttt{reference}, \texttt{tutorial}),
source and download URLs, the original filename, a short description,
and the download date. This registry is the single source of truth for the
raw-source side of the pipeline; downstream stages operate exclusively
on the files it points to, so the chunker and the LLM judge are
deterministic functions of the registry contents.

\paragraph{PDF to Markdown conversion.}
Every PDF in the registry is converted to Markdown with
\texttt{marker-pdf}\footnote{%
\url{https://github.com/VikParuchuri/marker}, accessed via the
\texttt{marker\_single} CLI.}, a layout-aware converter that handles
multi-column layouts, equations, tables, and figure captions, and
performs OCR-style cleaning when the underlying PDF lacks an
extractable text layer. The output is per-source Markdown that
preserves the document's heading hierarchy. Preserving headings is
critical: the chunker described next uses them as primary split
points.

\paragraph{Header-aware chunking with multi-level fallback.}
We split each Markdown document into token-bounded chunks using a
header-aware splitter,\footnote{We use
\texttt{MarkdownHeaderTextSplitter} from the
\texttt{langchain-text-splitters} library.} with a guaranteed token
budget per output chunk. Token counts are estimated with the Qwen3-32B
tokenizer~\citep{qwen2025technicalreport}; the chunker enforces a
maximum of $T_{\max}$ tokens per chunk and discards stray fragments
below $T_{\min}$. We set $T_{\min}=200$ and $T_{\max}=2048$.
When a header section already fits within $T_{\max}$, it is emitted
as a single chunk. When a section is oversized, the chunker applies
the following four-level fallback in order, stopping as soon as every
resulting fragment fits within $T_{\max}$: (1)~paragraph split,
(2)~sentence split, (3)~comma~/ semicolon split, (4)~adaptive
character split (chunk-size annealed until the fragments fit). This
ordering preserves the natural prose structure as long as possible
and falls through to lower-level splits only when the higher levels
still produce oversized fragments. Every emitted chunk is therefore
guaranteed to satisfy
$T_{\min} \leq \text{tokens} \leq T_{\max}$.

\paragraph{LLM-driven content filtering.}
We apply LLM-driven filtering at two granularities. First,
\emph{before} chunking, an optional structure pass runs over the
converted Markdown to identify the ``first chapter'' and
``end marker'' boundaries, so that front matter (table of contents,
prefaces, lists of contributors) and back matter (bibliography,
indexes, appendix exercise keys) are excluded from the chunk pool
wholesale. Second, \emph{after} chunking, an LLM judge scores each
remaining chunk for teaching value and drops chunks that fail the
bar. The judge is a Qwen3-8B
model~\citep{qwen2025technicalreport} served via vLLM and queried
with a fixed rubric that asks the judge to return a binary
\texttt{keep} decision plus a one-line reason. A chunk is kept only
when it has at least a few complete sentences that explain a
concept, method, result, or definition, and a reader could learn
something non-trivial from it without seeing the surrounding pages.
A chunk is discarded if any of the following hold: (i)~it is mostly
index-like or glossary-like (terms followed by page numbers,
cross-references, or markdown page-anchor links); (ii)~it is mostly
tables, character tables, or matrices of symbols and numbers without
surrounding explanation; (iii)~it lacks professional-level content
or is subjective navigation/structural text such as a preface,
foreword, acknowledgements, bare structural headings, or labels like
``Index'', ``References'', ``Table of Contents''; (iv)~it is
garbled OCR or broken fragments; (v)~it consists mainly of pointers
to other material (``see Figure~2'', ``see Chapter~5'') without
explaining the underlying ideas; or (vi)~it is an answer key,
solution manual, or list of short answers to review/practice
questions, even if some entries carry brief explanations. A
borderline rule biases the judge toward
\texttt{keep=false} when a chunk mixes noise with only a tiny amount
of real content, so the resulting pool is conservatively filtered
toward self-contained teaching material. The judge is run with high
parallelism per source document. The exact prompt template, including
both keep/discard criteria and the two calibration examples used in
production, is shown below; the chunk text is inserted into the
\texttt{\$\{content\}} slot.

\begin{tcolorbox}[
  enhanced,
  colback=sectionblue!3!white,
  colframe=sectionblue,
  colbacktitle=sectionblue,
  coltitle=white,
  boxrule=0.7pt,
  arc=10pt,
  left=8pt,
  right=8pt,
  top=8pt,
  bottom=8pt,
  title={LLM Judge Prompt for Chunk-Quality Filtering},
  fonttitle=\bfseries,
]
The LLM judge is queried with a single user turn containing the rubric, two calibration examples, and the chunk to evaluate.
\end{tcolorbox}

\begin{promptmessage}[colback=findingaccent!8!white,
colframe=findingaccent!70!black,
colbacktitle=findingaccent!70!black]{User}
You are evaluating a text chunk to decide if it should be kept as training material for knowledge distillation.

Goal: keep only chunks that contain self-contained, explanatory, professional-level content. If you are unsure, choose keep=false.

KEEP (keep=true) only if:
- The chunk has at least a few complete sentences that explain a concept, method, result, or definition.
- A reader could learn something non-trivial from this chunk without seeing surrounding pages.

DISCARD (keep=false) if ANY of the following are true:
1) The chunk is mostly index-like or glossary-like: terms followed by page numbers, cross-references, or markdown links like [282](#page-289-13). Example patterns:
   - 'molecular vibrations, 101-103; wave functions, 14-26; ...'
   - 'radial artery, **[890](#page-897-3)**, **[918](#page-925-17)** radial collateral ligament, ...'
   Even if the chunk is very long, if it is mostly terms + page numbers/links, discard it.
2) The chunk is mostly a table, character table, or matrix of symbols/numbers (including markdown tables with many '|' characters) and there is no surrounding explanation of what the table means.
3) The chunk lacks professional-level knowledge or contains subjective content (e.g. preface, foreword, acknowledgements, structural headings like 'Chapter 3 - Results', or navigation text like 'Index', 'References', 'Table of Contents').
4) The chunk is obviously garbled OCR, or mainly broken fragments that are hard to interpret.
5) The chunk is mainly pointers to other material (e.g. 'see Figure 2', 'see Chapter 5') without explaining the underlying ideas.
6) The chunk is an answer key, solution manual, or list of short answers to review/practice questions (e.g. '[1] B [2] D [3] C ...' or '[1] The kidneys. [2] X-rays.'). Even if some answers contain brief explanations, answer keys are not self-contained teaching material.

Borderline rule: if the chunk is a mix of noise and a tiny amount of real content, choose keep=false.

Examples (for calibration only):
Example KEEP:
Chunk: 'Gradient descent updates parameters by moving opposite to the gradient. Given learning rate eta, the update is theta_{t+1} = theta_t - eta * grad L(theta_t). This iteratively reduces the loss under mild smoothness assumptions.'
Output: {"keep": true, "reason": "Self-contained explanation of gradient descent and its update rule."}

Example DISCARD (index-like):
Chunk: 'Wave equation, 16; wave functions, 14-26; molecular orbitals, 117; selection rules, 414-415; semiconductors, 231-234; silicon, 231, 241, 250.'
Output: {"keep": false, "reason": "Index-style terms with page numbers, no explanatory sentences."}

Now evaluate the following chunk.

Chunk:
${content}

MUST TO RETURN ONLY JSON WITH FORMAT:
{"keep": true|false, "reason": "short explanation (max 20 words)"}
\end{promptmessage}

\paragraph{Output format and final pool composition.}
Each source produces a single JSON file containing the surviving
chunks, with one record per chunk and metadata indicating the source
identifier and the header path of the chunk inside the original
document. Concatenating across all sources in
the registry yields the document pool $\mathcal{D}_{\mathrm{doc}}$
with $|\mathcal{D}_{\mathrm{doc}}| = 12{,}260$ chunks. The
discipline-level composition (Biochemistry, Physics, Astronomy,
Geography) and the average chunk length in characters and tokens are
reported alongside the deduplication analysis in
\S\ref{app:dedup_check}; we do not duplicate those numbers here.

\paragraph{Reproducibility.}
The pipeline is deterministic up to the LLM agents' sampling
randomness and is reproducible from the registry: re-running the
chunker and the LLM judge on the registered sources reproduces the
same chunk pool up to floating-point and sampling variation. The
registry, chunker, and judge configuration are released alongside the
codebase.

\section{Deduplication check against the OLMo-3 SFT corpus}
\label{app:dedup_check}

This appendix documents the near-duplicate check summarized in
\S\ref{sec:instruction_finetune}.

\paragraph{Motivation.}
Since the IF anchor in \S\ref{sec:instruction_finetune} is
OLMo-3-7B-Instruct-SFT, a natural concern is that the benchmark signal
on INFUSER could be inflated by data contamination: if our document
pool $\mathcal{D}_{\mathrm{doc}}$ or our development set
$\mathcal{D}_{\mathrm{dev}}$ overlaps with the OLMo-3
supervised-finetune mixture, the anchor has already seen the raw
content from which INFUSER curates its curricula, and any observed
lift could reflect memorization rather than influence-guided
self-improvement. We therefore run a lexical near-duplicate check of
both $\mathcal{D}_{\mathrm{doc}}$ and $\mathcal{D}_{\mathrm{dev}}$
against the publicly released OLMo-3 SFT data.

\paragraph{Corpora.}
The reference side comprises two components.
$\mathcal{D}_{\mathrm{doc}}$ is the document pool used in our training
runs: a collection of $12{,}260$ PDF-derived chunks
(Biochemistry $58\%$, Physics $24\%$, Astronomy $14\%$,
Geography $5\%$; average length $\sim 5\text{k}$ characters,
$\sim 1.3\text{k}$ tokens per chunk). $\mathcal{D}_{\mathrm{dev}}$ is
the $800$-question SuperGPQA Science development subset used by
INFUSER to drive the influence score. For indexing, we serialize each
question by concatenating the question text, choice list, and
reference answer into a single string.

The instruction-tuning side is the $2{,}152{,}112$-sample mixture
\texttt{allenai/Dolci-Instruct-SFT}~\citep{allenai2025dolci}, released
alongside OLMo-3-7B-Instruct-SFT. Each Dolci record carries an ordered
\texttt{messages} list with four possible roles (\texttt{user},
\texttt{assistant}, \texttt{system}, \texttt{environment}). We
concatenate the \texttt{user} and \texttt{assistant} contents per
record, since these are the two roles in which textbook-derived
content would plausibly appear (quoted in a prompt or reproduced in an
answer). The \texttt{system} role is dominated by function-calling
boilerplate and the \texttt{environment} role by tool-call JSON
output; neither carries prose excerpted from a science textbook or a
science exam.

\paragraph{Standard we follow.}
We follow the reference intra-corpus deduplication pipeline documented
by \citet{cerebras_dedup}, which in turn reproduces the word-$n$-gram
\emph{MinHashLSH} approach of \citet{lee2022dedup} built on the
probabilistic resemblance framework of
\citet{broder1997resemblance}. In this regime, each text blob is
normalized under Unicode NFC, casefolded, stripped of ASCII
punctuation, and collapsed to single whitespace; the resulting token
stream is shingled into word $13$-grams. A $128$-permutation MinHash
signature is computed per blob and inserted into a MinHashLSH index
with Jaccard threshold $0.8$. Following \citet{cerebras_dedup},
records shorter than $200$ characters after normalization are excluded
from indexing.

\paragraph{Containment pass.}
Because the document chunks in $\mathcal{D}_{\mathrm{doc}}$ are
typically much longer than a single SFT sample, symmetric Jaccard is
not the only regime that matters: an SFT sample could copy a short
passage fully contained in a much longer chunk, in which case the
symmetric similarity would fall well below the $0.8$ floor even when
every shingle of the sample appears in the chunk. To handle this
length-asymmetric case, we additionally build a \emph{MinHashLSH
Ensemble}~\citep{zhu2016lshensemble} over the same $128$-permutation
signatures, with a containment threshold of $0.8$ and $32$ partitions.
The same construction is applied to $\mathcal{D}_{\mathrm{dev}}$, so
both the symmetric and asymmetric regimes are probed on each side.
For every candidate pair returned by either index we verify the
estimated Jaccard (and, for ensemble candidates, the MinHash-derived
containment $\widehat{|A \cap B|}/|A|$ with
$\widehat{|A \cap B|} = \widehat{J}\cdot(|A|+|B|)/(1+\widehat{J})$)
against the thresholds before recording a match. All signatures and
indexes are computed with the \texttt{datasketch} Python
library~\citep{zhu_datasketch}, and scanning is parallelized across
the $15$ parquet shards of the Dolci train split.

\paragraph{Results.}
Across the full $2{,}152{,}111$ non-trivial SFT samples (after
dropping $76{,}337$ below the length floor, $3.5\%$), the combined
Jaccard-LSH + LSH-Ensemble scan returns \textbf{zero matches} against
$\mathcal{D}_{\mathrm{doc}}$ and \textbf{zero matches} against
$\mathcal{D}_{\mathrm{dev}}$ at the standard thresholds. No SFT sample
has estimated Jaccard similarity $\geq 0.8$ with any chunk or
dev-question serialization, and no SFT sample has estimated
containment $\geq 0.8$ in any chunk or dev-question serialization. We
therefore conclude that neither the document pool
$\mathcal{D}_{\mathrm{doc}}$ nor the dev set
$\mathcal{D}_{\mathrm{dev}}$ leaks into the OLMo-3 SFT mixture at the
near-duplicate thresholds recommended by \citet{cerebras_dedup} and
\citet{lee2022dedup}, so the benchmark lifts attributed to INFUSER in
\S\ref{sec:instruction_finetune} cannot be explained by anchor-level
memorization of our training corpus.

\paragraph{Caveats.}
Word $13$-grams with a Jaccard threshold of $0.8$ catch verbatim
near-copies robustly but are deliberately insensitive to paraphrases,
summaries, and fact-level reformulations. A secondary scan with word
$5$-grams at Jaccard threshold $0.5$ and containment threshold $0.8$
surfaces only two $\mathcal{D}_{\mathrm{doc}}$ chunks (both from the
same analytical-chemistry textbook) that share $\sim 20\text{--}30\%$
of their shingles with two \texttt{OpenThoughts3+ Science} prompts.
Manual inspection shows that the matched shingles are stock phrasing
shared by exercises from the same textbook family (standard
voltammetry setup, calibration language), not direct reuse of our
chunk text; we therefore do not treat them as contamination. The
looser scan finds no such near-misses for $\mathcal{D}_{\mathrm{dev}}$.
Finally, $\mathcal{D}_{\mathrm{doc}}$ and $\mathcal{D}_{\mathrm{dev}}$
cover the sciences only, while the Dolci mixture spans many
non-science domains (coding, general instruction-following, tool use,
multilingual). The \emph{a priori} overlap probability with those
domains is low, which is consistent with the null result.

\section{Raw Data Behind Main-Text Figures}\label{app:raw_data_tables}

The main text presents several results graphically.  This appendix
collects the raw per-benchmark accuracies behind those figures so that
readers can audit individual numbers, recompute deltas, or quote
specific benchmark scores.  

\paragraph{Headline-benchmark scores.}
\Cref{tab:best_score_bar_qw8bb} reports the per-method accuracy on the
four headline benchmarks plotted in \Cref{fig:main_qw8bb}~(left).  The
values are pulled from the same report database as
\Cref{tab:main_benchmarks}, so this table is the Qwen3-8B-Base subset
of the main table on those four benchmarks, listed here for
convenience.

\begin{table}[!t]
  \centering
  \caption{%
    Per-benchmark accuracy (\%) for the four headline benchmarks
    plotted in \Cref{fig:main_qw8bb} (left) on the Qwen3-8B-Base
    anchor.  Values are pulled from the same report database as
    \Cref{tab:main_benchmarks}, so this table is the Qwen3-8B-Base
    subset of the main table on these four benchmarks.
    R-Few\textsuperscript{$\dagger$} and SPICE\textsuperscript{$\dagger$}
    columns are self-reported (see \S\ref{app:eval_protocol}).  Bold
    entries mark the best score per row.
  }
  \label{tab:best_score_bar_qw8bb}
  \begin{tabular}{l rrrrrr}
    \toprule
    Benchmark & Base & R-Zero & AZR & R-Few\textsuperscript{$\dagger$} & SPICE\textsuperscript{$\dagger$} & INFUSER \\
    \midrule
    MATH500 & 76.05 & 80.55 & 80.95 & 82.60 & 79.40 & \textbf{82.77} \\
    OlympiadBench (Math) & 40.36 & 45.10 & 47.92 & 46.40 & 42.50 & \textbf{50.24} \\
    MMLU-Pro & 59.91 & 61.82 & 62.32 & 63.20 & 65.00 & \textbf{66.20} \\
    SuperGPQA & 30.62 & 32.06 & 32.63 & 33.50 & 35.70 & \textbf{37.77} \\
    \bottomrule
  \end{tabular}
\end{table}

\paragraph{Dev-set leakage test.}
\Cref{tab:dev_leakage} reports the dev-subset and held-out-complement
accuracies underlying \Cref{fig:dev_leakage}.  The
held-out complement row is derived by subtracting the
$800$-question dev counts from the full SuperGPQA Science pool.

\begin{table}[!t]
  \centering
  \caption{%
    Dev-dataset leakage test on Qwen3-8B-Base.  We score the base
    model, INFUSER, and the Dev-only baseline on the 800-question
    training dev subset ($\mathcal{D}_{\mathrm{dev}}$) and on the
    $9{,}038$-question SuperGPQA Science held-out complement
    (SuperGPQA Science with $\mathcal{D}_{\mathrm{dev}}$ removed).
    $\Delta$ is the absolute improvement over the base model.  The
    held-out complement row is derived automatically from the report's
    dev-subset and full-pool scores.
  }
  \label{tab:dev_leakage}
  \resizebox{0.97\textwidth}{!}{%
  \begin{tabular}{l r r r r r}
    \toprule
    & & \multicolumn{2}{c}{INFUSER} & \multicolumn{2}{c}{Dev-only} \\
    \cmidrule(lr){3-4} \cmidrule(lr){5-6}
    Dataset & Base & Acc. & $\Delta$ & Acc. & $\Delta$ \\
    \midrule
    $\mathcal{D}_{\mathrm{dev}}$ (800-question training subset) & 32.37 & 41.85 & $+9.48$ & 86.50 & \textcolor{red!70!black}{$\mathbf{+54.13}$} \\
    SuperGPQA Science held-out ($9{,}038$ questions, $\mathcal{D}_{\mathrm{dev}}$ removed) & 30.06 & 39.22 & $+9.16$ & 38.35 & $+8.29$ \\
    \bottomrule
  \end{tabular}%
  }
\end{table}

\paragraph{Generator ablations (source and update).}
\Cref{tab:qw8_ablation} consolidates the per-benchmark accuracies
behind the two Qwen3-8B-Base ablations in
\S\ref{sec:comparison}.  The first four trained columns, \emph{INFUSER},
\emph{Fix-gen}, \emph{Strong-gen}, and \emph{Dev-only}, correspond to
the generator-source ablation in \Cref{fig:source_ablation}.  The
remaining three columns, \texttt{group\_std}, \texttt{batch\_std}, and
\texttt{sgd\_cosine}, correspond to the generator-update ablation in
\Cref{fig:norm_ablation} and isolate the within-group normalizer, the
batch normalizer, and the SGD-style (non-preconditioned) similarity
variant; INFUSER itself is the DuGRPO anchor for that ablation, so it
appears once in the shared INFUSER column.

\begin{table}[!t]
  \centering
  \caption{%
    Per-benchmark solver accuracy (\%) on Qwen3-8B-Base for the
    consolidated ablation behind \Cref{fig:source_ablation}
    (generator source: INFUSER, Fix-gen, Strong-gen, Dev-only)
    and \Cref{fig:norm_ablation} (generator update: DuGRPO and
    its three normalization / similarity variants
    \texttt{group\_std}, \texttt{batch\_std},
    \texttt{sgd\_cosine}).  INFUSER is the DuGRPO anchor and
    appears in both ablations.  The INFUSER column is the seeded
    reference run that feeds \Cref{tab:main_benchmarks}; the
    corresponding cell in \Cref{tab:glr_sweep} (Qwen3-8B-Base,
    $G_\text{lr}{=}4{\times}10^{-6}$) uses a separate seedless sweep
    run with its own best checkpoint, so the two cells need not
    match.  Bold entries mark the best score per row across the
    seven trained columns (i.e.\ excluding Base).
  }
  \label{tab:qw8_ablation}
  \resizebox{\textwidth}{!}{%
  \begin{tabular}{l rrrrrrrr}
    \toprule
    Benchmark & Base & INFUSER & Fix-gen & Strong-gen & Dev-only & \texttt{group\_std} & \texttt{batch\_std} & \texttt{sgd\_cosine} \\
    \midrule
    \multicolumn{9}{l}{\textit{General reasoning}} \\
    MMLU-Pro & 59.91 & 67.81 & 65.48 & \textbf{68.46} & 62.55 & 66.31 & 65.69 & 66.01 \\
    GPQA-Diamond & 36.87 & \textbf{47.47} & 45.56 & 45.86 & 44.55 & 43.43 & 42.83 & 43.84 \\
    SuperGPQA & 30.62 & 38.86 & 37.87 & \textbf{41.01} & 40.36 & 38.04 & 36.97 & 37.80 \\
    BBEH & 10.30 & 12.51 & 12.79 & 12.46 & \textbf{13.57} & 12.91 & 12.66 & 12.15 \\
    \rowcolor{gray!15}\hspace{1em}\emph{Category average} & 34.43 & 41.66 & 40.43 & \textbf{41.95} & 40.26 & 40.17 & 39.54 & 39.95 \\
    \midrule
    \multicolumn{9}{l}{\textit{Math \& physics reasoning}} \\
    MATH500 & 76.05 & \textbf{84.25} & 78.70 & 82.35 & 83.05 & 80.10 & 80.05 & 80.00 \\
    AIME2024 & 12.92 & 19.06 & 15.31 & 14.90 & \textbf{21.77} & 15.00 & 15.21 & 12.60 \\
    AIME2025 & 11.87 & \textbf{18.02} & 14.06 & 13.33 & 17.60 & 12.40 & 13.13 & 11.98 \\
    HMMT & 2.96 & \textbf{9.64} & 3.93 & 5.68 & 7.90 & 3.97 & 4.50 & 4.03 \\
    OlympiadBench (Math) & 40.36 & \textbf{54.45} & 44.96 & 46.74 & 48.96 & 43.32 & 46.59 & 45.10 \\
    OlympiadBench (Phys) & 12.29 & 14.41 & \textbf{14.83} & 13.14 & 13.98 & 13.98 & 13.56 & 13.98 \\
    \rowcolor{gray!15}\hspace{1em}\emph{Category average} & 26.08 & \textbf{33.31} & 28.63 & 29.36 & 32.21 & 28.13 & 28.84 & 27.95 \\
    \midrule
    \multicolumn{9}{l}{\textit{Medical}} \\
    MedQA & 64.18 & 66.46 & 65.04 & 67.40 & \textbf{67.95} & 65.99 & 65.67 & 66.06 \\
    MedXpertQA & 14.49 & 14.57 & 15.22 & \textbf{17.47} & 15.31 & 14.94 & 16.33 & 16.00 \\
    \rowcolor{gray!15}\hspace{1em}\emph{Category average} & 39.34 & 40.52 & 40.13 & \textbf{42.44} & 41.63 & 40.46 & 41.00 & 41.03 \\
    \midrule
    \multicolumn{9}{l}{\textit{Coding}} \\
    HumanEval+ & 75.94 & 78.86 & 77.52 & 76.68 & 75.61 & 78.12 & 78.89 & \textbf{79.65} \\
    LiveCodeBench v1-5 & 25.23 & 28.47 & 27.73 & \textbf{28.75} & 26.59 & 27.67 & 28.01 & 28.35 \\
    \rowcolor{gray!15}\hspace{1em}\emph{Category average} & 50.59 & 53.67 & 52.63 & 52.72 & 51.10 & 52.90 & 53.45 & \textbf{54.00} \\
    \bottomrule
  \end{tabular}%
  }
\end{table}

\paragraph{Generator learning-rate sweep.}
\Cref{tab:glr_sweep} lists per-benchmark accuracies for the four
generator learning rates plotted in \Cref{fig:glr_sweep}
on both the Qwen3-4B-Base and Qwen3-8B-Base anchors.  $G_\text{lr}{=}0$
is the Fix-gen baseline.

\begin{table}[!t]
  \centering
  \caption{%
    Per-benchmark solver accuracy (\%) for the generator
    learning-rate sweep on Qwen3-4B-Base and Qwen3-8B-Base (raw
    numbers behind \Cref{fig:glr_sweep}).
    The $G_\text{lr}{=}0$ column is the Fix-gen baseline that
    freezes the generator at its initial checkpoint; nonzero
    columns use the best checkpoint per run, selected by the
    validation protocol of \S\ref{app:eval_protocol:ckpt_selection}.
    Bold entries mark the best $G_\text{lr}$ setting per row
    within each anchor.
  }
  \label{tab:glr_sweep}
  \resizebox{0.8\textwidth}{!}{%
  \begin{tabular}{l rrrrr}
    \toprule
    Benchmark & Base & $G_\text{lr}{=}0$ & $G_\text{lr}{=}2{\times}10^{-6}$ & $G_\text{lr}{=}4{\times}10^{-6}$ & $G_\text{lr}{=}6{\times}10^{-6}$ \\
    \midrule
    \multicolumn{6}{l}{\textbf{Qwen3-4B-Base}} \\
    \multicolumn{6}{l}{\hspace{0.5em}\textit{General reasoning}} \\
    MMLU-Pro & 52.98 & 59.46 & 59.39 & 59.78 & \textbf{60.68} \\
    GPQA-Diamond & 31.41 & 38.59 & 37.07 & \textbf{39.39} & 35.35 \\
    SuperGPQA & 25.88 & 33.00 & 33.12 & 32.07 & \textbf{33.90} \\
    BBEH & 5.18 & 10.55 & 9.21 & 9.30 & \textbf{12.11} \\
    \rowcolor{gray!15}\hspace{1em}\emph{Category average} & 28.86 & 35.40 & 34.70 & 35.14 & \textbf{35.51} \\
    \multicolumn{6}{l}{\hspace{0.5em}\textit{Math \& physics reasoning}} \\
    MATH500 & 61.20 & 76.25 & 76.75 & 74.85 & \textbf{77.90} \\
    AIME2024 & 10.42 & 10.62 & \textbf{14.48} & 9.27 & 11.87 \\
    AIME2025 & 8.44 & 8.85 & 10.42 & 9.79 & \textbf{11.56} \\
    HMMT & 2.49 & 2.86 & \textbf{3.36} & 2.96 & 3.19 \\
    OlympiadBench (Math) & 35.31 & \textbf{42.43} & 41.54 & 37.98 & 42.14 \\
    OlympiadBench (Phys) & 10.17 & \textbf{12.71} & 11.86 & 11.86 & 8.90 \\
    \rowcolor{gray!15}\hspace{1em}\emph{Category average} & 21.34 & 25.62 & \textbf{26.40} & 24.45 & 25.93 \\
    \multicolumn{6}{l}{\hspace{0.5em}\textit{Medical}} \\
    MedQA & 55.46 & 58.37 & 56.95 & 58.68 & \textbf{59.47} \\
    MedXpertQA & 13.02 & \textbf{13.88} & 13.18 & 13.18 & 13.80 \\
    \rowcolor{gray!15}\hspace{1em}\emph{Category average} & 34.24 & 36.13 & 35.07 & 35.93 & \textbf{36.64} \\
    \multicolumn{6}{l}{\hspace{0.5em}\textit{Coding}} \\
    HumanEval+ & 70.27 & 74.54 & \textbf{76.22} & 74.47 & 75.23 \\
    LiveCodeBench v1-5 & 20.68 & 22.05 & 22.67 & 21.70 & \textbf{23.01} \\
    \rowcolor{gray!15}\hspace{1em}\emph{Category average} & 45.47 & 48.30 & \textbf{49.45} & 48.09 & 49.12 \\
    \rowcolor{gray!15}\textbf{14-benchmark mean} & 28.78 & 33.15 & 33.30 & 32.52 & \textbf{33.51} \\
    \midrule
    \multicolumn{6}{l}{\textbf{Qwen3-8B-Base}} \\
    \multicolumn{6}{l}{\hspace{0.5em}\textit{General reasoning}} \\
    MMLU-Pro & 59.91 & 65.48 & 65.03 & 64.54 & \textbf{66.00} \\
    GPQA-Diamond & 36.87 & 45.56 & 44.34 & \textbf{45.76} & 43.77 \\
    SuperGPQA & 30.62 & 37.87 & \textbf{37.92} & 36.33 & 36.69 \\
    BBEH & 10.30 & \textbf{12.79} & 11.97 & 12.35 & 12.14 \\
    \rowcolor{gray!15}\hspace{1em}\emph{Category average} & 34.43 & \textbf{40.43} & 39.82 & 39.75 & 39.65 \\
    \multicolumn{6}{l}{\hspace{0.5em}\textit{Math \& physics reasoning}} \\
    MATH500 & 76.05 & 78.70 & 80.55 & \textbf{85.25} & 81.03 \\
    AIME2024 & 12.92 & 15.31 & 12.92 & \textbf{21.25} & 17.05 \\
    AIME2025 & 11.87 & 14.06 & 11.25 & \textbf{17.19} & 14.27 \\
    HMMT & 2.96 & 3.93 & 4.07 & \textbf{8.17} & 5.49 \\
    OlympiadBench (Math) & 40.36 & 44.96 & 41.99 & \textbf{51.48} & 46.24 \\
    OlympiadBench (Phys) & 12.29 & \textbf{14.83} & 11.86 & 13.98 & 12.01 \\
    \rowcolor{gray!15}\hspace{1em}\emph{Category average} & 26.08 & 28.63 & 27.11 & \textbf{32.89} & 29.35 \\
    \multicolumn{6}{l}{\hspace{0.5em}\textit{Medical}} \\
    MedQA & 64.18 & 65.04 & 63.79 & 65.12 & \textbf{66.01} \\
    MedXpertQA & 14.49 & 15.22 & \textbf{16.37} & 15.18 & 15.33 \\
    \rowcolor{gray!15}\hspace{1em}\emph{Category average} & 39.34 & 40.13 & 40.08 & 40.15 & \textbf{40.67} \\
    \multicolumn{6}{l}{\hspace{0.5em}\textit{Coding}} \\
    HumanEval+ & 75.94 & 77.52 & \textbf{80.11} & 77.97 & 78.43 \\
    LiveCodeBench v1-5 & 25.23 & \textbf{27.73} & 26.59 & 27.44 & 26.97 \\
    \rowcolor{gray!15}\hspace{1em}\emph{Category average} & 50.59 & 52.63 & \textbf{53.35} & 52.71 & 52.70 \\
    \rowcolor{gray!15}\textbf{14-benchmark mean} & 33.86 & 37.07 & 36.34 & \textbf{38.71} & 37.25 \\
    \bottomrule
  \end{tabular}%
  }
\end{table}

\paragraph{Generator question quality.}
\Cref{tab:gen_quality_qw8bb} reports the per-checkpoint accuracies of
the four solvers tracked across the co-evolving generator's questions
in \Cref{fig:gen_quality_qw8bb}.

\begin{table}[!t]
  \centering
  \caption{%
    Per-checkpoint solver accuracy (\%) on the questions produced
    by INFUSER's co-evolving generator at training iterations
    $\{0, 30, 60, 90\}$ on the Qwen3-8B-Base anchor (raw numbers
    behind \Cref{fig:gen_quality_qw8bb}).  ``Qwen3-8B-Base'' is the
    fixed reference base solver, ``INFUSER Solver'' is the evolving
    co-trained solver's own training-time accuracy on its current
    questions, and ``GPT-5.4-mini'' / ``GPT-5.4'' are strong-solver
    references.
  }
  \label{tab:gen_quality_qw8bb}
  \begin{tabular}{l rrrr}
    \toprule
    Solver & Iter.\ 0 & Iter.\ 30 & Iter.\ 60 & Iter.\ 90 \\
    \midrule
    Qwen3-8B-Base & 59.50 & 47.90 & 53.80 & 56.70 \\
    INFUSER Solver & 59.40 & 55.22 & 61.82 & 64.31 \\
    GPT-5.4-mini & 64.80 & 57.00 & 61.90 & 66.80 \\
    GPT-5.4 & 64.50 & 59.80 & 66.00 & 70.30 \\
    \bottomrule
  \end{tabular}
\end{table}

\paragraph{Instruction-finetuned anchor extension.}
\Cref{tab:olmo_instruct_sft} reports the per-benchmark accuracy behind
the three radar plots in \S\ref{sec:instruction_finetune}.

\begin{table}[!htbp]
  \centering
  \caption{%
    Solver accuracy (\%) on held-out benchmarks for
    OLMo-3-7B-Instruct-SFT as an instruction-finetuned anchor.
    Bolded entries mark the best score among the three variants per
    row; all three columns are produced by the same evaluation
    pipeline as \Cref{tab:main_benchmarks}.  Fix-gen is the
    frozen-generator ablation from \S\ref{sec:comparison}; INFUSER
    uses preconditioned-cosine influence with solver lr
    $2\times 10^{-6}$ and generator lr $4\times 10^{-6}$.
  }
  \label{tab:olmo_instruct_sft}
  \resizebox{0.5\textwidth}{!}{%
  \begin{tabular}{l ccc}
    \toprule
    Benchmark & Base & Fix-gen & INFUSER \\
    \midrule
    \multicolumn{4}{l}{\textit{General reasoning}} \\
    MMLU-Pro & 49.0 & 51.5 & \textbf{54.1} \\
    GPQA-Diamond & 31.6 & \textbf{36.2} & 35.6 \\
    SuperGPQA & 22.8 & 26.7 & \textbf{28.4} \\
    BBEH & 8.1 & \textbf{10.4} & 10.2 \\
    \rowcolor{gray!15}\hspace{1em}\emph{Category average} & 27.9 & 31.2 & \textbf{32.1} \\
    \midrule
    \multicolumn{4}{l}{\textit{Math \& physics reasoning}} \\
    MATH500 & 68.9 & 68.4 & \textbf{70.6} \\
    AIME2024 & 5.8 & 6.3 & \textbf{6.8} \\
    AIME2025 & 7.1 & 6.8 & \textbf{9.8} \\
    HMMT & 3.2 & 4.5 & \textbf{4.8} \\
    OlympiadBench (Math+Phys) & 25.5 & 27.0 & \textbf{27.3} \\
    \rowcolor{gray!15}\hspace{1em}\emph{Category average} & 22.1 & 22.6 & \textbf{23.9} \\
    \midrule
    \multicolumn{4}{l}{\textit{Medical}} \\
    MedQA & 45.1 & 43.1 & \textbf{45.3} \\
    MedXpertQA & 12.7 & 13.4 & \textbf{14.2} \\
    \rowcolor{gray!15}\hspace{1em}\emph{Category average} & 28.9 & 28.3 & \textbf{29.8} \\
    \midrule
    \multicolumn{4}{l}{\textit{Coding}} \\
    HumanEval+ & 67.2 & \textbf{68.0} & 66.6 \\
    LiveCodeBench & 12.8 & 10.6 & \textbf{13.3} \\
    \rowcolor{gray!15}\hspace{1em}\emph{Category average} & \textbf{40.0} & 39.3 & 39.9 \\
    \midrule
    \rowcolor{gray!15}\textbf{Overall average} & 27.7 & 28.7 & \textbf{29.8} \\
    \bottomrule
  \end{tabular}%
  }
\end{table}

\paragraph{Dev-set-size sensitivity.}
\Cref{tab:dev_size_sensitivity} reports the completed Qwen3-4B-Base
sweep discussed in \S\ref{sec:dev_size_sensitivity}.

\begin{table}[!t]
  \centering
  \caption{%
    Dev-set-size sensitivity on Qwen3-4B-Base.  Category columns
    report strict accuracy (\%) averaged within each paper category;
    the overall column is the unweighted mean over all 14 benchmark
    rows.  Each INFUSER setting is one selected-checkpoint run, so
    these entries are point estimates rather than seed averages.
    $\Delta$ is the absolute gain over the untrained Base mean.
  }
  \label{tab:dev_size_sensitivity}
  \resizebox{0.92\textwidth}{!}{%
  \begin{tabular}{lrrrrrrr}
    \toprule
    Dev size & Checkpoint & General & Math \& physics & Medical & Coding & 14-benchmark mean & $\Delta$ \\
    \midrule
    --- & --- & 29.46 & 21.34 & 34.24 & 45.47 & 28.95 & --- \\
    100 & 90 & 34.33 & 23.92 & 35.59 & 48.99 & 32.14 & $+3.19$ \\
    200 & 80 & 34.49 & 24.15 & 36.70 & 48.75 & 32.41 & $+3.46$ \\
    400 & 60 & 34.56 & 26.11 & 36.76 & 48.86 & 33.30 & $+4.35$ \\
    800 & 75 & 35.63 & 25.93 & 36.63 & 49.12 & 33.54 & $+4.59$ \\
    \bottomrule
  \end{tabular}%
  }
\end{table}

\begin{figure}[!t]
  \centering
  \includegraphics[width=\linewidth]{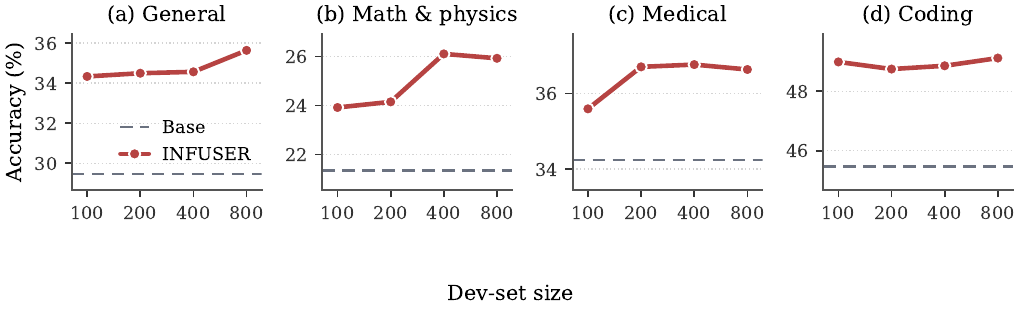}
  \caption{Qwen3-4B-Base dev-set-size sensitivity by evaluation domain.
  Each INFUSER point is one run selected under the common validation
  protocol (checkpoints 90, 80, 60, and 75 for 100, 200, 400, and 800
  examples); dashed lines show the corresponding untrained Base scores.}
  \label{fig:dev_size_sensitivity_domains}
\end{figure}

\paragraph{Influence-reward diagnostics.}
\Cref{tab:influence_diagnostics} reports the three-seed co-evolution
trajectories and the single-run frozen-solver control discussed in
\S\ref{sec:curriculum_analysis}.

\begin{table*}[!t]
  \centering
  \caption{%
    Influence-reward diagnostics.  Co-evolution rows report the mean
    and sample standard deviation across the three targeted seed groups
    at exact common logged steps; a parenthetical logged step is shown
    only when the nearest common step replaces a nominal 10-iteration
    mark.  Aggregate rows first average over each seed's logged steps
    in the stated range, then summarize the three seed averages.  The frozen-solver
    control reports its distinct cosine metric every five iterations,
    with the true centered-20-sample curve (truncated, not shifted, at endpoints).
    All values are in units of $10^{-3}$.
  }
  \label{tab:influence_diagnostics}
  \resizebox{\textwidth}{!}{%
  \begin{tabular}{llrrrr}
    \toprule
    Setting & Iteration / range & Logged iter. & Mean / raw & Seed sample SD & Centered-20 \\
    \midrule
    Qwen3-4B-Base & 0 & 0 & +1.98 & 0.55 & --- \\
    Qwen3-4B-Base & 10 & 10 & +0.16 & 0.11 & --- \\
    Qwen3-4B-Base & 20 & 20 & +0.07 & 0.27 & --- \\
    Qwen3-4B-Base & 30 & 30 & +0.10 & 0.11 & --- \\
    Qwen3-4B-Base & 40 & 40 & -0.05 & 0.05 & --- \\
    Qwen3-4B-Base & 50 & 50 & +0.00 & 0.31 & --- \\
    Qwen3-4B-Base & 60 & 60 & -0.08 & 0.30 & --- \\
    Qwen3-4B-Base & 70 & 70 & +0.14 & 0.46 & --- \\
    Qwen3-4B-Base & 80 & 80 & -0.15 & 0.11 & --- \\
    Qwen3-4B-Base & 90 & 90 & -0.07 & 0.30 & --- \\
    Qwen3-4B-Base & all common iters & 0--99 & \textbf{+0.10} & \textbf{0.01} & --- \\
    Qwen3-4B-Base & common iters $\geq 20$ & 20--99 & \textbf{+0.03} & \textbf{0.01} & --- \\
    \addlinespace
    Qwen3-8B-Base & 0 & 0 & +0.94 & 0.27 & --- \\
    Qwen3-8B-Base & 10 & 10 & +0.09 & 0.25 & --- \\
    Qwen3-8B-Base & 20 & 20 & -0.03 & 0.12 & --- \\
    Qwen3-8B-Base & 30 & 30 & +0.26 & 0.20 & --- \\
    Qwen3-8B-Base & 40 & 40 & -0.13 & 0.12 & --- \\
    Qwen3-8B-Base & 50 & 50 & +0.12 & 0.22 & --- \\
    Qwen3-8B-Base & 60 & 60 & +0.09 & 0.09 & --- \\
    Qwen3-8B-Base & 70 & 70 & +0.09 & 0.11 & --- \\
    Qwen3-8B-Base & 80 & 80 & +0.02 & 0.18 & --- \\
    Qwen3-8B-Base & 90 & 90 & -0.07 & 0.17 & --- \\
    Qwen3-8B-Base & all common iters & 0--99 & \textbf{+0.07} & \textbf{0.04} & --- \\
    Qwen3-8B-Base & common iters $\geq 20$ & 20--99 & \textbf{+0.04} & \textbf{0.03} & --- \\
    \addlinespace
    Frozen solver & 0 & 0 & +0.62 & --- & +0.45 \\
    Frozen solver & 5 & 5 & -1.06 & --- & +0.12 \\
    Frozen solver & 10 & 10 & +1.20 & --- & +0.30 \\
    Frozen solver & 15 & 15 & -0.62 & --- & +0.23 \\
    Frozen solver & 20 & 20 & -0.51 & --- & +0.37 \\
    Frozen solver & 25 & 25 & +0.90 & --- & +0.61 \\
    Frozen solver & 30 & 30 & +0.07 & --- & +0.60 \\
    Frozen solver & 35 & 35 & +0.87 & --- & +0.55 \\
    Frozen solver & 40 & 40 & +2.52 & --- & +0.76 \\
    Frozen solver & 45 & 45 & +0.67 & --- & +0.89 \\
    Frozen solver & 47 & 47 & +0.38 & --- & +0.82 \\
    \bottomrule
  \end{tabular}%
  }
  \vspace{2pt}
  \begin{minipage}{0.98\textwidth}\footnotesize
  Co-evolution metric: \texttt{influence\_sim/score/mean}. Frozen-control metric: \texttt{influence\_sim/cosine/mean}. Frozen run: \texttt{tnh5a27m} (FW-Glr\_4e-6-vanilla\_smtm-TIS\_token-repeat\_doc\_20260303\_125229). Centered window: 20 samples. Seed labels: \texttt{seed42/default}, \texttt{seed123}, \texttt{seed456}.
  \end{minipage}
\end{table*}

\end{document}